\documentclass[11pt]{article}

\usepackage{epsfig,amsmath,latexsym,amssymb}
\usepackage{graphicx}
\usepackage{lscape}
\usepackage{picture, eso-pic, tikz} 
\usepackage{dsfont}
\usepackage{listings}
\usepackage{changes}
\usepackage{hyperref}

\renewcommand{\baselinestretch}{1.1}
\def\R{{\mathbb R}}  
\def\N{{\mathbb N}}  
\def\E{{\mathbb E}}  %

\newcommand{\Remm}[1]{}
\newtheorem{theo}{Theorem}[section]

\newtheorem{model ass}[theo]{Model Assumptions}
\newtheorem{ass}[theo]{Assumptions}

\newtheorem{rems}[theo]{Remarks}

\numberwithin{equation}{section}

\definecolor{MyGray}{rgb}{0.92,0.92,0.92}


\definecolor{British racing}{rgb}{0.0, 0.5, 0.0}
\def\bx{\boldsymbol{x}}

\def\bz{\boldsymbol{z}}
\def\bw{\boldsymbol{w}}

\def\b0{\boldsymbol{0}}

\def\bTheta{\boldsymbol{\Theta}}

\def\bbeta{\boldsymbol{\beta}}

\def\b0{\boldsymbol{0}}

\begin{document}
\author{Ronald Richman\footnote{University of the Witwatersrand, ronaldrichman@gmail.com} 
\and Mario V.~W\"uthrich\footnote{RiskLab, Department of Mathematics, ETH Zurich,
mario.wuethrich@math.ethz.ch}}

\date{Version of \today}
\title{LocalGLMnet: interpretable deep learning for tabular data}
\maketitle

\begin{abstract}
\noindent  
Deep learning models have gained great popularity in statistical modeling because they lead to
very competitive regression models, often outperforming classical statistical models
such as generalized linear models. The disadvantage of deep learning models is that their
solutions are difficult to interpret and explain, and variable selection is not easily possible
because deep learning models solve feature engineering and variable selection internally in a nontransparent way.
Inspired by the appealing structure of generalized linear models, we propose a new
network architecture that shares similar features as generalized linear models, but
provides superior predictive power benefiting from the art of representation learning. 
This new architecture allows for variable selection of tabular data
and for interpretation of the calibrated deep learning model, in fact, our approach
provides an additive decomposition in the spirit of Shapley values and integrated gradients.

~

\end{abstract}

\section{Introduction}
Deep learning models celebrate great success in statistical modeling because they often provide
superior predictive power over classical regression models. This success is based on the fact that
deep learning models perform representation learning of features, which means that they bring features
into the right structure to be able to extract maximal information for the prediction task at hand. This feature
engineering is done internally in a nontransparent way by the deep learning model. For this reason
deep learning solutions are often criticized to be non-explainable and interpretable, in particular, because
this process of representation learning is performed in high-dimensional spaces analyzing bits and pieces
of the feature information. Recent research has been focusing on interpreting machine learning predictions
in retrospect, see, e.g., Friedman's partial dependence plot (PDP) \cite{friedman2001greedy},
the accumulated local effects (ALE) method of Apley--Zhu \cite{Apley}, the locally interpretable model-agnostic explanation (LIME) introduced by Ribeiro et al.~\cite{Ribeiro},
the SHapley Additive exPlanations (SHAP) of Lundberg--Lee \cite{LundbergLee}
 or the marginal attribution by conditioning on quantiles (MACQ)
method proposed by Merz et al.~\cite{MerzRichmanTsanakas}. PDP, ALE, LIME and SHAP can be used for any machine
learning method such as random forests, boosting or neural networks, whereas MACQ requires differentiability of
the regression function which is the case for neural networks under differentiable activation functions; for a review
of more gradient based methods we refer to Merz et al.~\cite{MerzRichmanTsanakas}.

We follow a different approach here, namely, we 
propose a new network architecture that has an internal structure that
directly allows  for interpreting and explaining. Moreover, this internal structure also allows
for variable selection of tabular feature data and to extract interactions between feature components.
The starting point of our proposal is the framework of generalized linear models (GLMs) introduced
by Nelder--Wedderburn \cite{Nelder} and McCullagh--Nelder \cite{MN}.
GLMs are characterized by the choice of a link function that maps the regression function to a linear predictor, 
and, thus, leading to a linear functional form that directly describes the influence of each predictor variable on the
response variable. Of course, this (generalized) linear form is both transparent and interpretable.
To some extent, our architecture preserves this linear structure of GLMs, but we make the
coefficients of the linear predictors feature dependent, too. Such an approach follows a similar strategy
as the ResNet proposal of He et al.~\cite{HeEtAl} that considers a linear term and then builds the network
around this linear term. The LassoNet of Lemhadri et al.~\cite{Lemhadri} follows a similar philosophy, too,
by performing Lasso regularization
on network features. Both proposals have in common that they use a so-called skip connection in the
network architecture that gives a linear modeling part around which the network model is built. Our proposal uses
such a skip connection, too, which provides the linear modeling part, and we weight these linear terms with
potentially non-linear weights. This allows us to generate non-linear regression functions with arbitrary 
interactions. In spirit, our non-linear weights are similar to the attention layers recently introduced
by Bahdanau et al.~\cite{Bahdanau} and Vaswani et al.~\cite{VaswaniEtAl}. Attention layers are a rather
successful new way of building powerful networks by  extracting more important feature components
from embeddings
by giving more weight (attention) to them. From this viewpoint we construct (network regression) attention weights that
provide us with a local GLM for tabular data, and we therefore call our proposal
LocalGLMnet. These regression attention weights also provide us with a possibility of explicit variable selection, which is a novel and unique property within network regression models. Moreover, we can explicitly
explore interactions between feature components. We mention that a similar approach is studied
by Ahn et al.~\cite{Ahn} in Bayesian credibility theory, where the credibility weights are modeled by
attention weights.

There is another stream of literature that tries to choose regression structures that have interpretable features, e.g.,
the explainable neural networks (xNN) and the 
neural additive models (NAM) make restrictions to the structure of the network regression function by running different
subsets of feature components through (separated) parallel networks, see 
Vaughan et al.~\cite{Vaughan} and
Agarwal et al.~\cite{Agarwal}. The drawback
of these proposals is that representation learning can only occur within the parallel networks. Our proposal overcomes
this issue as it allows for general interactions. We also mention Richman \cite{Richman3} who extends the xNN approach by
explicitly including linear features to a combined
model called CAXNN. 
Another interpretable network approach is the TabNet proposal of Arik--Pfister \cite{Arik}. TabNet uses networks
to create attention to important features for the regression task at hand, however, this proposal
has the drawback that it may lead to heavy
computational burden.

Our LocalGLMnet approach overcomes the limitations of these explainable network approaches. As we will see  below, our proposal is computationally efficient and it leads to a nice explanation, in the sense that we can also interpret
our model in terms of Shapley \cite{Shapley} values, we refer to Lundberg--Lee \cite{LundbergLee}
and Sundararajan--Najmi \cite{Sundararajan2} for SHAP. In fact, we could also argue that our idea makes the
SHAP interpretation to a regression model assumption, this will be further explored
in Section \ref{Interpretation and extension of LocalGLMnet}, below.

~

{\bf Organization of this manuscript.}
In the next section we introduce and discuss the LocalGLMnet. We therefore first recall the GLM framework which
will give us the right starting point and intuition for the LocalGLMnet. In Section 
\ref{Fully-connected feed-forward neural network} we extend GLMs to feed-forward neural networks
that form the basis of our regression attention weight construction, and Section \ref{Local generalized linear model network}
presents our LocalGLMnet proposal that combines GLMs with regression attention weights.
Section \ref{Interpretation and extension of LocalGLMnet} discusses
the LocalGLMnet,  and it relates our proposal to SHAP.
Section \ref{Examples} presents two examples, a synthetic data example and a real data example. The former
will give us a proof of concept, a verification is obtained because we know the true data generating mechanism
in the synthetic data example. Moreover, in Section \ref{Variable selection} we discuss how the LocalGLMnet
allows for variable selection, which is a novel and unique property within network regression modeling.
Section \ref{Interactions sections} explains how we can find interactions.
In Section \ref{Real data example} we present a real data example, 
Section \ref{Variable importance} discusses variable importance, and in Section \ref{Categorical feature components}
we  discuss
how categorical feature components can be treated within our proposal. Finally, in Section \ref{Conclusion section}
we conclude.

\section{Model architecture}
\label{Model architecture section}
\subsection{Generalized linear model}
\label{Generalized linear model}
The starting point of our proposal is a GLM which typically is based on the exponential dispersion family (EDF). 
GLMs have been introduced by Nelder--Wedderburn \cite{Nelder} and McCullagh--Nelder \cite{MN}, and the
EDF has been analyzed in detail by Barndorff-Nielsen \cite{Barndorff} and J{\o}rgensen \cite{Jorgensen3};
the present paper uses the notation and terminology of W\"uthrich--Merz \cite{WM2021}, and for a detailed 
treatment of GLMs and the EDF we also refer to Chapters 2 and 5 of that latter reference.

Assume we have a datum $(Y,\bx,v)$
with a given exposure $v>0$, a vector-valued feature $\bx \in  \R^q$, and a response variable $Y$ following
a member of the (single-parameter linear) EDF having density (w.r.t.~to a $\sigma$-finite measure on $\R$)
\begin{equation}\label{exponential dispersion family}
Y~\sim~
f(y; \theta, v/\varphi)= \exp \left\{ \frac{y\theta - \kappa(\theta)}{\varphi/v} + a(y;v/\varphi)\right\} ,
\end{equation}
with dispersion parameter $\varphi>0$, canonical parameter $\theta \in \bTheta$, where the effective
domain $\bTheta \subseteq \R$ is a non-empty interval, with cumulant function $\kappa:\bTheta\to \R$
and with normalizing function $a(\cdot; \cdot)$. By construction of the EDF, the cumulant function $\kappa$ is a smooth and
convex function on the interior of the effective domain $\bTheta$. This then implies that $Y$ has first and second moments,
we refer to Chapter 2 in  W\"uthrich--Merz \cite{WM2021},
\begin{equation*}
\mu=  \E\left[Y\right] = \kappa'(\theta)
  \qquad \text{ and } \qquad {\rm Var}\left(Y\right) = \frac{\varphi}{v}\kappa''(\theta)>0.
\end{equation*}
A GLM is obtained by making a specific regression assumption on the mean $\mu= \kappa'(\theta)$ of $Y$. Namely,
choose a strictly monotone and continuous link function $g:\R \to \R$ and assume that the mean of $Y$, given $\bx$, 
satisfies 
\begin{equation}\label{GLM}
\bx ~\mapsto~
g(\mu) = g\left( \mu (\bx) \right) = \beta_0 + \langle \bbeta , \bx \rangle=
\beta_0 + \sum_{j=1}^q \beta_j x_j,
\end{equation}
with GLM regression parameter $\bbeta=(\beta_1,\ldots, \beta_q)^\top \in \R^{q}$, bias (intercept) $\beta_0\in \R$, and where $\langle \cdot, \cdot \rangle$ denotes
the scalar product in the Euclidean space $\R^{q}$. This GLM assumption implies that the canonical
parameter takes the following form (on the canonical scale of the EDF)
\begin{equation*}
\theta = (\kappa')^{-1}\left(g^{-1}\left(\beta_0+ \langle \bbeta , \bx \rangle \right)\right),
\end{equation*}
where $(\kappa')^{-1}$ is the canonical link of the chosen EDF \eqref{exponential dispersion family}.

The GLM regression function \eqref{GLM} is very appealing because it leads to a linear predictor
$\eta(\bx)=\beta_0 + \langle \bbeta , \bx \rangle$
after applying the link function $g$ to the mean $\mu(\bx)$, and the regression parameter $\beta_j$
directly explains how the individual feature component $x_j$ influences 
the linear predictor $\eta(\bx)$ and the expected value $\mu(\bx)$ of $Y$, respectively.
Our goal is to benefit from this transparent structure as far as possible.

\subsection{Fully-connected feed-forward neural network}
\label{Fully-connected feed-forward neural network}
The neural network extension of a GLM can be obtained rather easily by allowing for feature engineering
before considering the scalar product in the linear predictor \eqref{GLM}. A fully-connected feed-forward neural
(FFN) network builds upon engineering feature information $\bx$ through non-linear transformations before entering 
the scalar product. A composition of FFN layers performs these non-linear transformations. Choose
a non-linear activation function $\phi_m:\R \to \R$ and integers (dimensions) $q_{m-1},q_m \in \N$. The $m$-th FFN layer 
of a deep FFN network 
is defined by the mapping
\begin{eqnarray}\label{FN layer}
\bz^{(m)}:\R^{q_{m-1}} &\to& \R^{q_m} \qquad\\
 \bx &\mapsto& \bz^{(m)}(\bx)=
\left(z_1^{(m)}(\bx), \ldots, z_{q_m}^{(m)}(\bx)\right)^\top,\nonumber
\end{eqnarray} 
having neurons $z_j^{(m)}(\bx)$, $1\le j \le q_m$, for $\bx=(x_1,\ldots, x_{q_{m-1}})^\top \in \R^{q_{m-1}}$,
\begin{equation*}
z_j^{(m)}(\bx)=\phi_m\left(w^{(m)}_{0,j} + \left\langle \bw^{(m)}_j, \bx
\right\rangle \right)=
\phi_m \left( w^{(m)}_{0,j} + \sum_{l=1}^{q_{m-1}} w^{(m)}_{l,j} x_l \right),
\end{equation*}
for given network weights $\bw^{(m)}_j=(w^{(m)}_{l,j})^\top_{1\le l \le q_{m-1}} \in \R^{q_{m-1}}$
and bias $w^{(m)}_{0,j} \in \R$.

A FFN network of depth $d \in \N$ is obtained by composing $d$ FFN layers \eqref{FN layer}
to provide a deep learned representation, we set input dimension $q_0=q$,
\begin{eqnarray}\label{deep representation learning}
\bz^{(d:1)}:\R^{q} &\to& \R^{q_d} \qquad\\
 \bx &\mapsto& \bz^{(d:1)}(\bx)=
 \left(\bz^{(d)} \circ \cdots \circ \bz^{(1)}\right)(\bx).\nonumber
\end{eqnarray} 
This $q_d$-dimensional learned representation $\bz^{(d:1)}(\bx) \in \R^{q_d}$ then enters a GLM  of type \eqref{GLM} providing
FFN network regression function
\begin{equation}\label{FN network regression}
\bx ~\mapsto~
g(\mu) = g\left( \mu (\bx) \right) = \beta_0 + \left\langle \bbeta , \bz^{(d:1)}(\bx) \right\rangle,
\end{equation}
with GLM regression (output) parameter $\bbeta=(\beta_1,\ldots, \beta_{q_d})^\top \in \R^{q_d}$ and bias $\beta_0\in \R$.
From \eqref{FN network regression} we see that the ``raw'' feature $\bx$ is first suitably transformed before
entering the GLM structure. The standard reference for neural networks is Goodfellow et al.~\cite{Goodfellow},
for more insight, interpretation and model fitting we refer to Section 7.2 of W\"uthrich--Merz \cite{WM2021}.

\subsection{Local generalized linear model network}
\label{Local generalized linear model network}
The disadvantage of deep representation learning \eqref{deep representation learning} is that
we can no longer track how individual feature components $x_j$ of $\bx$ influence 
regression function \eqref{FN network regression} because the composition of FFN layers acts
rather as a black box, e.g.,  in general, it is not clear how each feature component $x_j$ influences
the response $\mu(\bx)$, whether 
a certain component $x_j$ needs to be
included in the regression function or whether it could be dropped because it does not contribute.
This is neither clear for an individual example $\mu(\bx)$ (locally) nor at a global level.

The key idea of our LocalGLMnet proposal is to retain the GLM structure \eqref{GLM} as far as possible, but to let the regression parameters
$\beta_j=\beta_j(\bx)$ become feature $\bx$ dependent; we call $\bbeta$ {\it regression parameter} if it does
not depend on $\bx$, and we call $\bbeta(\bx)$ {\it regression attention} if it is $\bx$-dependent.
Our proposal of the LocalGLMnet architecture can
be interpreted as a local GLM with network learned regression attention, as we are going to model the
regression attentions
$\beta_j(\bx) \in \R$ by networks.
Strictly speaking, we typically lose the linearity if we let $\beta_j(\bx)$ be feature dependent, however, if
this dependence is smooth, we have a sort of a local GLM parameter which justifies our terminology, 
see also Section \ref{Interpretation and extension of LocalGLMnet}, below.

\begin{ass}[LocalGLMnet]
Choose a FFN network architecture of depth $d\in\N$ with input and output dimensions being equal to $q_0=q_d=q$
to model the {\it attention weights}
\begin{eqnarray}\label{LocalGLMnet weight}
\bbeta:\R^{q} &\to& \R^{q} \qquad\\
 \bx &\mapsto& \bbeta(\bx)=\bz^{(d:1)}(\bx)=
 \left(\bz^{(d)} \circ \cdots \circ \bz^{(1)}\right)(\bx).\nonumber
\end{eqnarray} 
The {\it LocalGLMnet} is defined by the {\it additive decomposition}
\begin{equation}\label{LocalGLMnet}
\bx ~\mapsto~
g(\mu) = g\left( \mu (\bx) \right) = \beta_0+ \langle \bbeta(\bx) , \bx \rangle.
\end{equation}
\end{ass}
Network architecture \eqref{LocalGLMnet} is a FFN  network architecture with
a skip connection: firstly, feature $\bx$ is processed through the deep FFN network providing us with
learned representation $\bbeta(\bx) \in \R^q$, and secondly, $\bx$ has a direct link to the output layer
(skipping all FFN layers) providing an (untransformed) linear term $\bx$. The LocalGLMnet then 
scalar multiplies these two different components, see \eqref{LocalGLMnet}. In the next section we give extended remarks and interpretation.

This model can be fitted to data by state-of-the art stochastic gradient descent (SGD) methods using
training and validation data for performing early stopping to not over-fit to the training data, for details
we refer to Goodfellow et al.~\cite{Goodfellow} and Section 7.2 in W\"uthrich--Merz \cite{WM2021}.

\subsection{Interpretation and extension of LocalGLMnet}
\label{Interpretation and extension of LocalGLMnet}
We call \eqref{LocalGLMnet} a LocalGLMnet because in a small environment ${\cal B}(\bx)$ around
$\bx$ we may approximate regression attention
$\bbeta(\bx')$, $\bx' \in {\cal B}(\bx)$, by a constant  regression parameter giving us the interpretation of a local GLM.
Network architecture \eqref{LocalGLMnet} can also be interpreted as an attention mechanism because $\beta_j(\bx)$
decides how much attention should be given to feature value $x_j$.

Yet, another interpretation of \eqref{LocalGLMnet} is given in terms of Shapley \cite{Shapley} values. We briefly
discuss SHAP to make this link; for references on SHAP we refer to Lundberg--Lee \cite{LundbergLee}, Sundararajan--Najmi \cite{Sundararajan2} and Aas et al.~\cite{Aas}. Shapley \cite{Shapley} values have their orgin in 
cooperative game theory by providing a ``fair'' allocation of a common gain to individual players. This concept has
been translated to deep learning models by aiming at attributing a joint response $\widetilde{\mu}(\bx)$ to
individual feature components $x_j$ so that we receive an additive decomposition
\begin{equation}\label{SHAP}
\widetilde{\mu}(\bx) ~\approx~ \alpha_0 + \sum_{j=1}^q \alpha_j x_j,
\end{equation}
where $\alpha_j$ describes the contribution of component $x_j$ of $\bx$ to response $\widetilde{\mu}(\bx)$. This
attribution is required to fulfill certain axioms of fairness, see Lundberg--Lee \cite{LundbergLee}.
A critical issue in the calculation of these attributions is the combinatorial complexity which can be very computational.
For this reason, approximations have been proposed, e.g., under the assumption of independence between feature components
a fair allocation in \eqref{SHAP} can be approximated efficiently.
Not surprisingly, these approximations have also been criticized as not being suitable and giving
wrong interpretations, see Aas et al.~\cite{Aas}. The LocalGLMnet solves the explanation problem
\eqref{SHAP} differently, namely, instead of fitting a complex model $\widetilde{\mu}(\cdot)$ to data that needs to be
interpreted in a subsequent step, our LocalGLMnet directly postulates an additive decomposition
\eqref{FN network regression} in response $\mu(\cdot)$ after applying the link function $g$. Thus, we could also say that
we make the interpretation an integral part our model assumptions, and we naturally obtain the regression attentions 
$\beta_j(\bx)$ playing the role of $\alpha_j$ by model fitting. We also mention the similarity to integrated
gradients of  Sundararajan et al.~\cite{Sundararajan}.

\begin{rems}\normalfont
\begin{itemize}
\item Formula \eqref{LocalGLMnet} defines the LocalGLMnet. If we replace the scalar product in  \eqref{LocalGLMnet}
by a Hadamard product $\otimes$ (component-wise product) we receive a {\it LocalGLM layer}
\begin{equation*}
\bx ~\mapsto~
\bbeta^{(1)}(\bx) \otimes \bx ~=~\left(\beta^{(1)}_1(\bx)x_1,\ldots,\beta^{(1)}_q(\bx)x_q\right)^\top \in \R^q.
\end{equation*}
A {\it deep LocalGLMnet} can be received by composing such LocalGLM layers, for instance, if we compose
two such layers we receive a regression function
\begin{equation*}
\bx ~\mapsto~
g(\mu) = g\left( \mu (\bx) \right) = \beta^{(2)}_0+ \left\langle \bbeta^{(2)}\left(\bbeta^{(1)}(\bx) \otimes \bx\right) , \bx 
\right\rangle.
\end{equation*}
Such an architecture may lead to increased predictive power and interpretable intermediate steps.
\item Above we have emphasized that the LocalGLMnet will lead to an interpretable regression model, and the
verification of this statement will be done in the examples, below. Alternatively, if one wants to rely
on a plain-vanilla deep FFN network, one can still fit a LocalGLMnet to the deep FFN network as
an interpretable surrogate model.
\item
The LocalGLMnet has been introduced for tabular data as we try to mimic a GLM that acts on a design
matrix which naturally is in  tabular form. If we extend the LocalGLMnet to unstructured
data, time series or image recognition it requires that this data is first encoded into tabular form, e.g.,
by using a convolutional module that extracts from spatial data relevant feature information and transforms
this into tabular structure. That is, it requires the paradigm of representation learning by first bringing raw
inputs into a suitable form before encoding this information by the LocalGLMnet for prediction.
\end{itemize}
\end{rems}

Before we study the performance of the LocalGLMnet we would like to get the right interpretation
and intuition for regression function \eqref{LocalGLMnet}. We select one component $1\le j \le q$ which
provides us on the linear scale (after applying the link $g$) with terms
\begin{equation}\label{interpretation of terms}
  \beta_j(\bx) x_j.
\end{equation}
We mention specific cases in the following remarks and how they should be interpreted.
\begin{rems}\label{interpretation of terms 2}\normalfont
\begin{itemize}
\item[(1)] A GLM term is obtained in component $x_j$ if $\beta_j(\bx) \equiv \beta_j$ is not feature dependent, providing
$\beta_j x_j$, we refer to GLM \eqref{GLM}. 
\item[(2)] Condition $\beta_j(\bx) \equiv 0$ proposes that the term $x_j$
should not be included. In Section \ref{Variable selection}, below, we are going to present an empirical method to test
for the null hypothesis of dropping a term.
\item[(3)] Property $\beta_j(\bx)=\beta_j(x_j)$ says that we have a term $\beta_j(x_j)x_j$ that 
does not interact with other terms. In general, we can analyze $\beta_j(\bx)$ over different
features $\bx$ with the $j$-th component being (constantly) equal to a fixed value $x_j$. If this  $\beta_j(\bx)$ does not
show any sensitivity in the components different from $j$, then we do not have interactions and otherwise we do.
Below, we extract this information by considering the gradients
\begin{equation}\label{gradient}
\nabla \beta_j(\bx) = \left( \frac{\partial}{\partial x_1}\beta_j(\bx), \ldots, \frac{\partial}{\partial x_q}\beta_j(\bx)
\right)^\top ~\in ~\R^q.
\end{equation}
The $j$-th component of this gradient $\nabla \beta_j(\bx)$ explores whether we have a linear term in $x_j$, and the components different from $j$ quantify the interaction strengths.
\item[(4)] One has to be a bit careful with these interpretations as we do not have full identifiability in model
calibration, as, e.g., we could also receive the following structure
\begin{equation*}
\beta_j(\bx) x_j = x_{j'},
\end{equation*}
by learning a regression attention $\beta_j(\bx)=x_{j'}/x_{j}$.
However, our tests on different configurations have not manifested any such issues as SGD fitting seems
rather pre-determined by the LocalGLMnet functional form \eqref{LocalGLMnet}.
\end{itemize}
\end{rems}

\section{Examples}
\label{Examples}
\subsection{Synthetic data example}
We start with a synthetic data example because this has the advantage of knowing the true data generating
model. This allows us to verify that we draw the right conclusions. We choose $q=8$ feature components
$\bx=(x_1,\ldots, x_8)^\top \in \R^8$. We generate two data sets, learning data ${\cal L}$ and test
data ${\cal T}$. The learning data will be used for model fitting and the test data for an out-of-sample
generalization analysis. We choose for both data sets $n=100,000$ randomly generated independent features $\bx \sim {\cal N}(\b0, \Sigma)$ being centered and having unit variance. Moreover, we assume that all components of $\bx$ are independent, 
except between $x_2$ and $x_8$ we assume a correlation of 50\%. Based on these features we choose
regression function
\begin{equation}\label{true synthetic}
\bx \in \R^8 ~\mapsto ~ \mu(\bx)=\frac{1}{2} x_1 - \frac{1}{4}x_2^2 + \frac{1}{2}|x_3| \sin(2x_3)
+ \frac{1}{2} x_4 x_5 + \frac{1}{8} x_5^2 x_6,
\end{equation}
thus, neither $x_7$ nor $x_8$ run into the regression function, $x_7$ is independent from the remaining components,
and $x_8$ has a 50\% correlation with $x_2$. Based on this regression function we generate independent 
Gaussian observations
\begin{equation}\label{Gaussian model}
 Y \sim {\cal N}(\mu(\bx),1).
 \end{equation}
This gives us the two data sets with all observations being independent
\begin{equation*}
{\cal L} = \{(Y_i,\bx_i);~1\le i \le n\} \qquad \text{ and } \qquad {\cal T} = \{(Y_t,\bx_t);~n+1\le t \le 2n\}.
\end{equation*}
The Gaussian model \eqref{Gaussian model} belongs to the EDF with cumulant function $\kappa(\theta)=\theta^2/2$,
effective domain $\bTheta=\R$, exposure $v=1$ and dispersion parameter $\varphi=1$. Thus, we can apply the theory
of Section \ref{Model architecture section}. 

We start with the GLM. 
As link function $g$ we choose the identity function which is the canonical
link of the Gaussian model. 
This provides us with linear predictor in the GLM case
\begin{equation}\label{GLM synthetic}
\bx ~\mapsto~
 \mu (\bx)  =\eta(\bx)  = \beta_0 + \langle \bbeta , \bx \rangle,
\end{equation}
with regression parameter $\bbeta \in \R^8$ and bias $\beta_0 \in \R$. This model is fit to the
learning data ${\cal L}$ using maximum likelihood estimation (MLE) which, in the Gaussian case, is equivalent
to minimizing the mean squared error (MSE) loss function for regression parameter $(\beta_0,\bbeta)$
\begin{equation*}
(\widehat{\beta}^{\rm MLE}_0,\widehat{\bbeta}^{\rm MLE})~=~
\underset{(\beta_0,\bbeta)}{\arg\min}~ \frac{1}{n}\sum_{i=1}^n \left(Y_i - \mu (\bx_i) \right)^2.
\end{equation*}
For this fitted model we calculate the in-sample MSE  on ${\cal L}$ and the out-of-sample 
MSE  on ${\cal T}$. We compare these losses to the MSEs  of the true model $\mu(\bx_i)$, which
is available here, and the corresponding MSEs  of the null model which only includes a bias $\beta_0$.
These figures are given in Table \ref{loss results synthetic} on lines (a)-(c).

\begin{table}[htb!]
\begin{center}
{\small
\begin{tabular}{|l|cc|}
\hline
&\multicolumn{2}{|c|}{MSE losses}\\
& in-sample on ${\cal L}$ & out-of-sample on ${\cal T}$\\\hline
(a) true regression function $\mu(\cdot)$ &1.0023&0.9955\\\hline
(b) null model (bias $\beta_0$ only) & 1.7907 & 1.7916\\
(c) GLM case & 1.5241 & 1.5274 \\\hline
(d) LocalGLMnet & 1.0023 & 1.0047 \\
\hline
\end{tabular}}
\end{center}
\caption{In-sample and out-of-sample MSEs in synthetic data example.}
\label{loss results synthetic}
\end{table}

\begin{figure}[htb!]
\begin{center}
\begin{minipage}[t]{0.45\textwidth}
\begin{center}
\includegraphics[width=.9\textwidth]{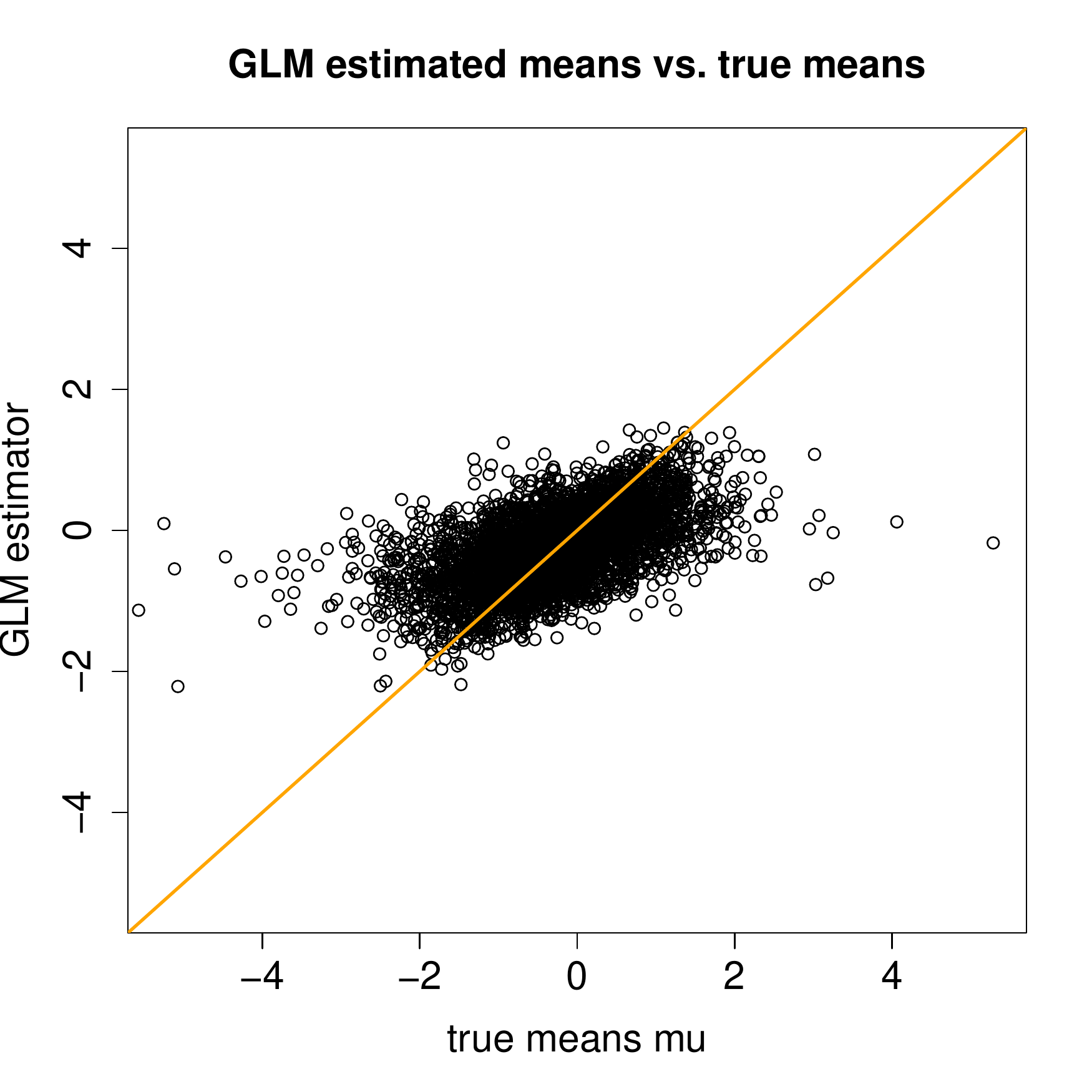}
\end{center}
\end{minipage}
\begin{minipage}[t]{0.45\textwidth}
\begin{center}
\includegraphics[width=.9\textwidth]{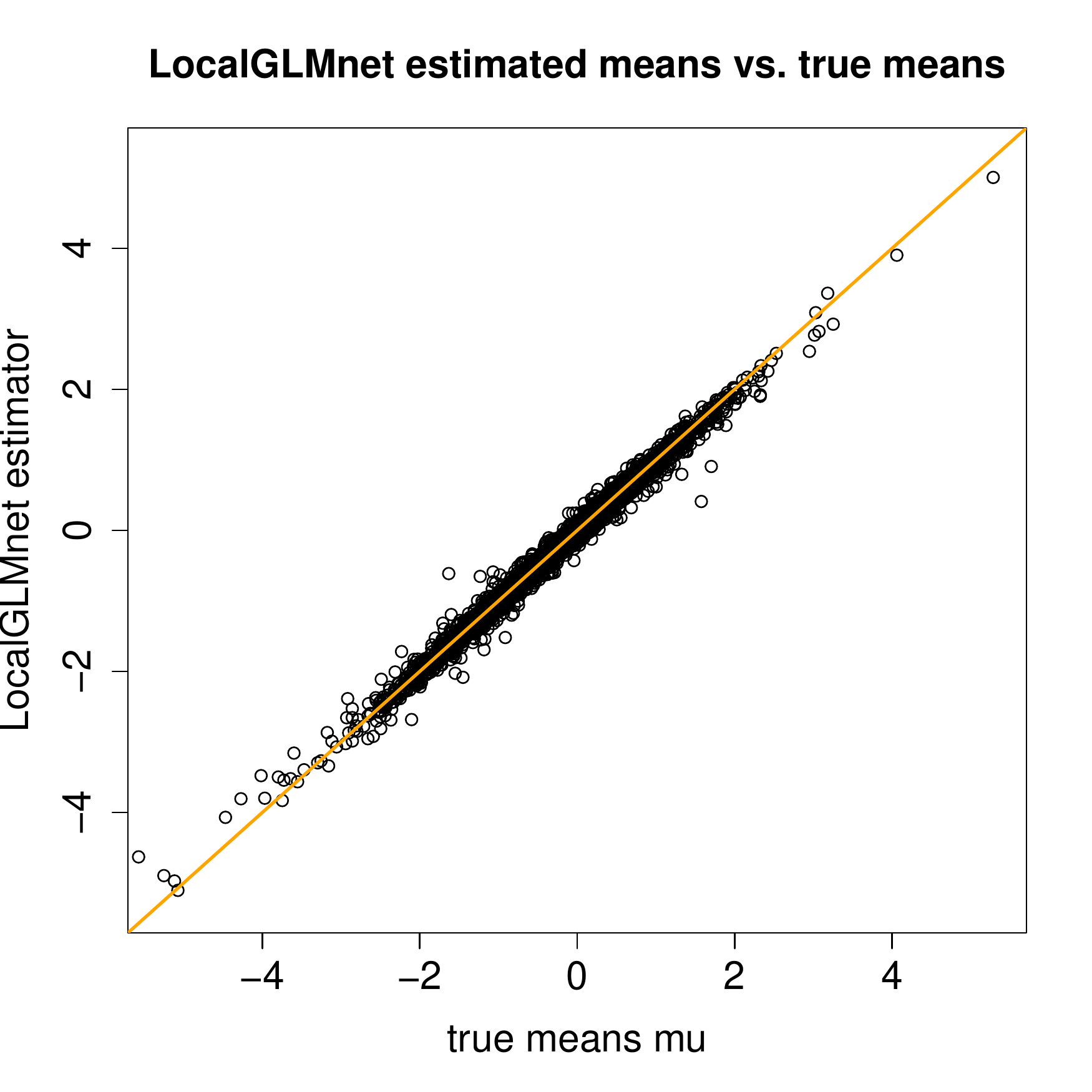}
\end{center}
\end{minipage}
\end{center}
\caption{Estimated means $\widehat{\mu}(\bx_t)$ vs.~true means $\mu(\bx_t)$: (lhs) fitted GLM 
\eqref{GLM synthetic} and (rhs) fitted LocalGLMnet of $5,000$ randomly selected out-of-sample
instances $\bx_t$ from ${\cal T}$.}
\label{accuracy of fitted models}
\end{figure}

The MSEs under the true regression function are roughly equal to 1 which exactly corresponds
to the fact that the responses $Y$ have been simulated with unit variance, see \eqref{Gaussian model}, the small differences to
1 correspond to the randomness implied by simulating. The loss figures of the GLM are between the null model (homogeneous model) and the correct model. Still these loss figures are comparably large because the true model \eqref{true synthetic}
has a rather different functional form compared to what we can capture by the linear function \eqref{GLM synthetic}.
This is also verified by Figure \ref{accuracy of fitted models} (lhs) which compares the 
GLM estimated means $\widehat{\mu}(\bx_t)$ to the true means $\mu(\bx_t)$ for different instances $\bx_t$. A perfect model would provide
points on the diagonal orange line, and we see rater big differences between the GLM and the true
regression function $\mu$.

We could now start to improve \eqref{GLM synthetic}, e.g., by including a quadratic term. We refrain from doing so,
but we fit the LocalGLMnet architecture \eqref{LocalGLMnet} using the identity link for $g$, a network of depth $d=4$
having $(q_0, q_1,q_2,q_3,q_4)=(8,20,15,10,8)$ neurons and as activation functions $\phi_m$ we choose the
hyperbolic tangent function for $m=1,2,3$ and the linear function for $m=4$. This architecture is illustrated in 
Listing \ref{LocalGLMnetCode1} in the appendix. This LocalGLMnet is fitted using the {\tt nadam} version of
SGD; early stopping is tracked by using 20\% of the learning data ${\cal L}$ as validation data ${\cal V}$ and the remaining 80\% as
training data ${\cal U}$, and the network calibration with the smallest MSE validation loss on ${\cal V}$ is selected;
note that ${\cal L}={\cal U}\cup{\cal V}$ is disjoint from the test data ${\cal T}$. This fitting approach is state-of-the-art,
for more details we refer to Section 7.2.3 in W\"uthrich--Merz \cite{WM2021}. The results are given 
on line (d) of Table \ref{loss results synthetic}. We observe that the MSEs (in-sample and out-of-sample)
are very close to 1 (and the MSEs of the true model $\mu$) which indicates that the LocalGLMnet is able to find
the true regression structure \eqref{GLM synthetic}. This is verified by Figure \ref{accuracy of fitted models} (rhs)
which plots the estimated means $\widehat{\mu}(\bx_t)$ against the true means $\mu(\bx_t)$ (out-of-sample)
for $5,000$ randomly selected instances $\bx_t$ from ${\cal T}$.
The fitted LocalGLMnet estimators lie on the  diagonal which says that we have very good accuracy.

\begin{figure}[htb!]
\begin{center}
\begin{minipage}[t]{0.32\textwidth}
\begin{center}
\includegraphics[width=.9\textwidth]{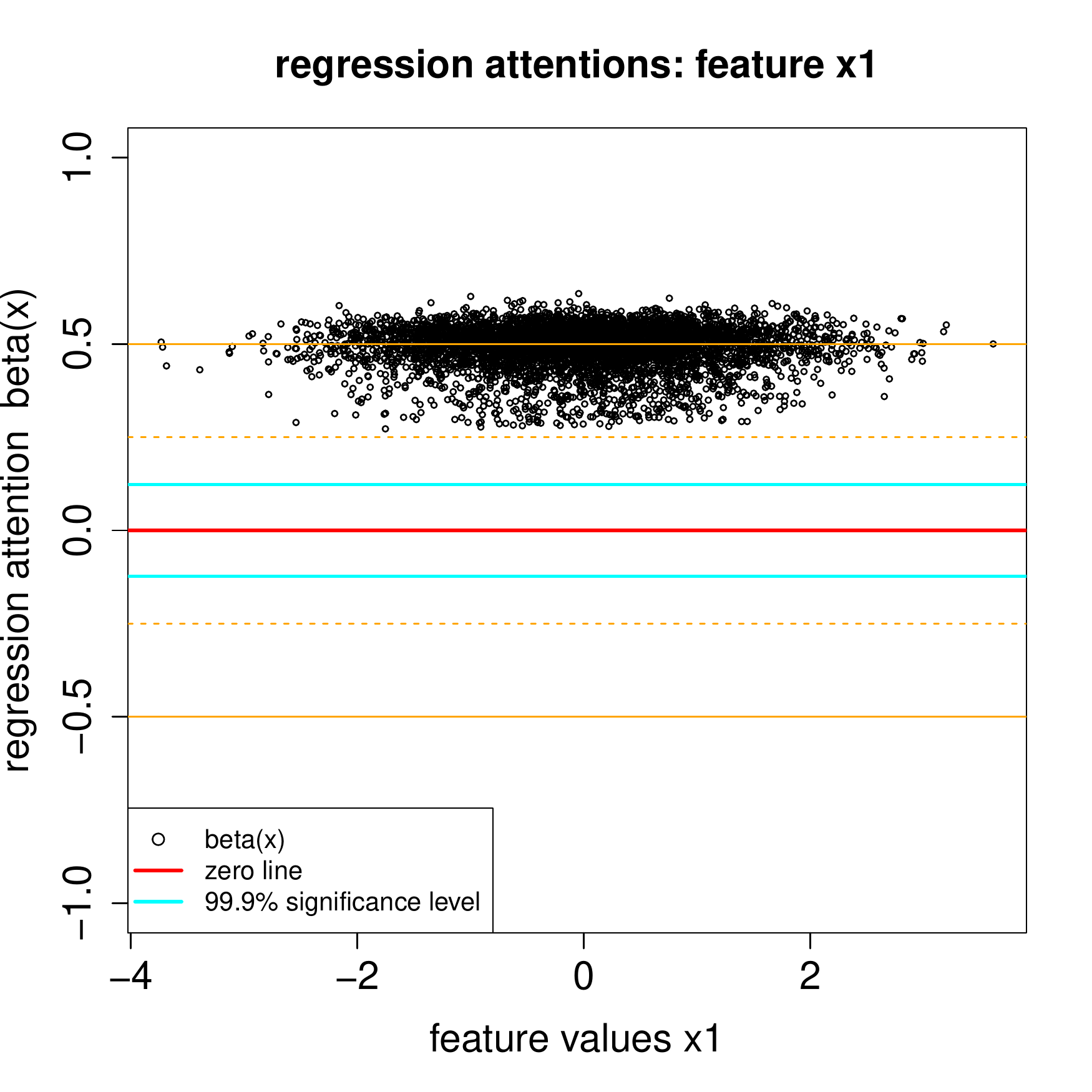}
\end{center}
\end{minipage}
\begin{minipage}[t]{0.32\textwidth}
\begin{center}
\includegraphics[width=.9\textwidth]{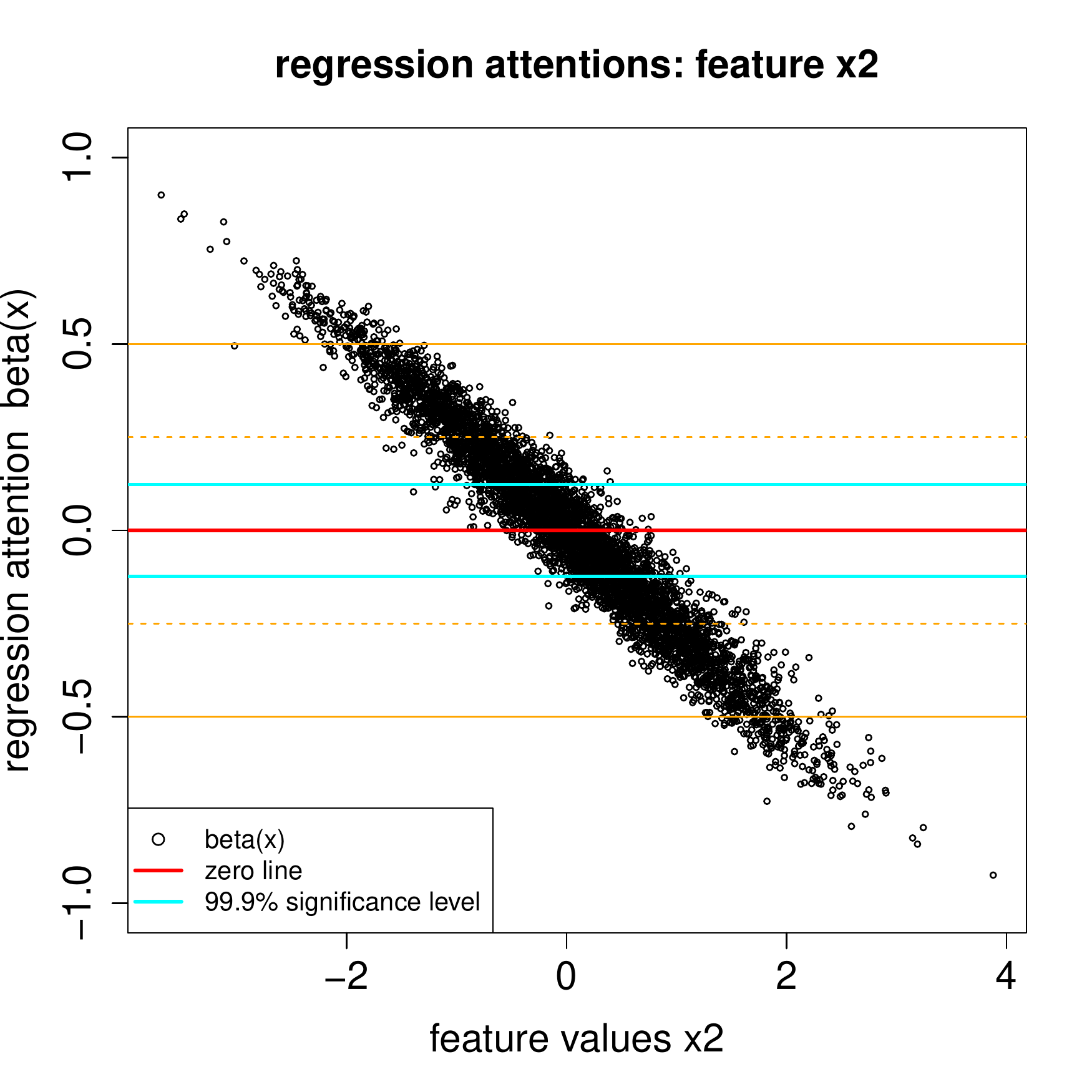}
\end{center}
\end{minipage}
\begin{minipage}[t]{0.32\textwidth}
\begin{center}
\includegraphics[width=.9\textwidth]{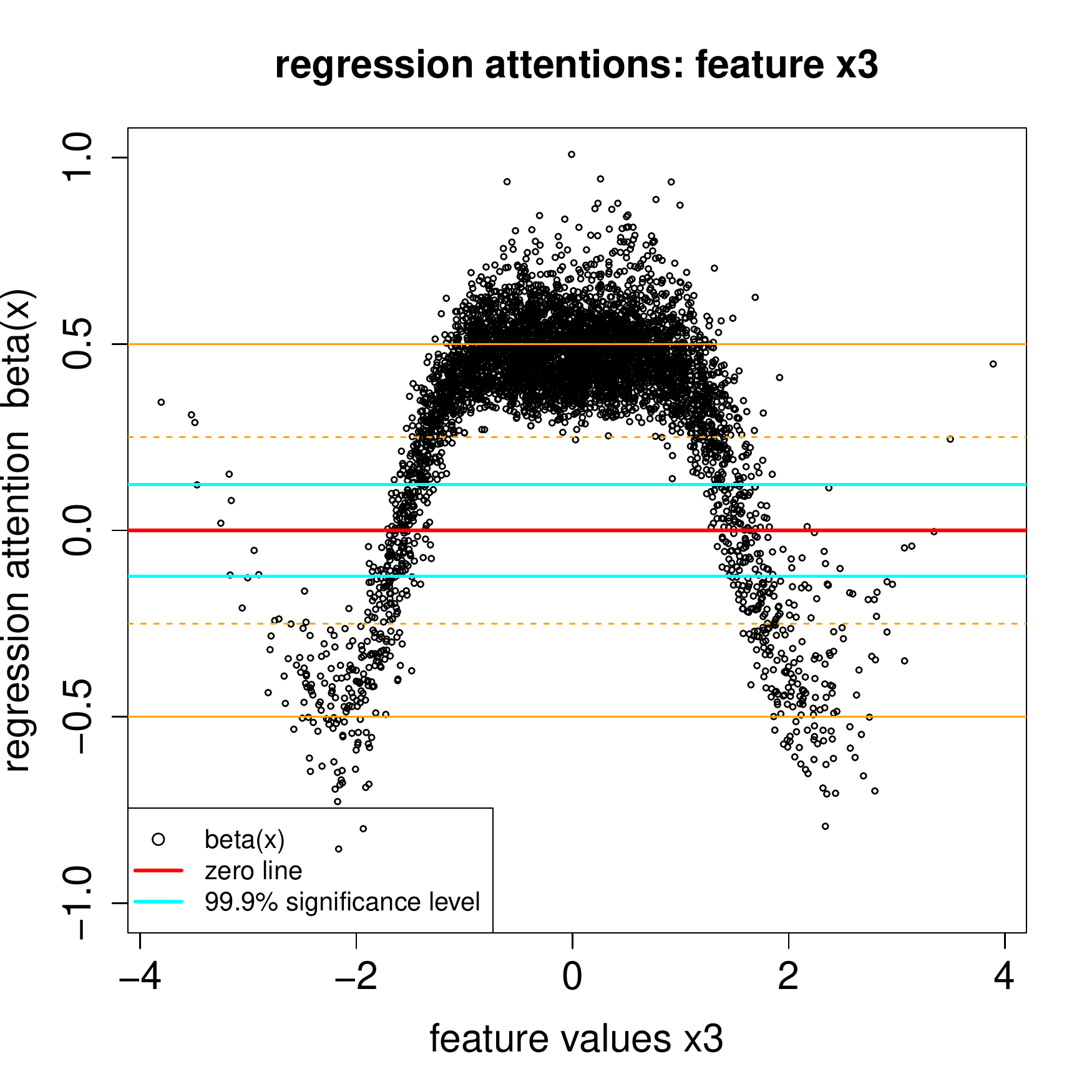}
\end{center}
\end{minipage}
\begin{minipage}[t]{0.32\textwidth}
\begin{center}
\includegraphics[width=.9\textwidth]{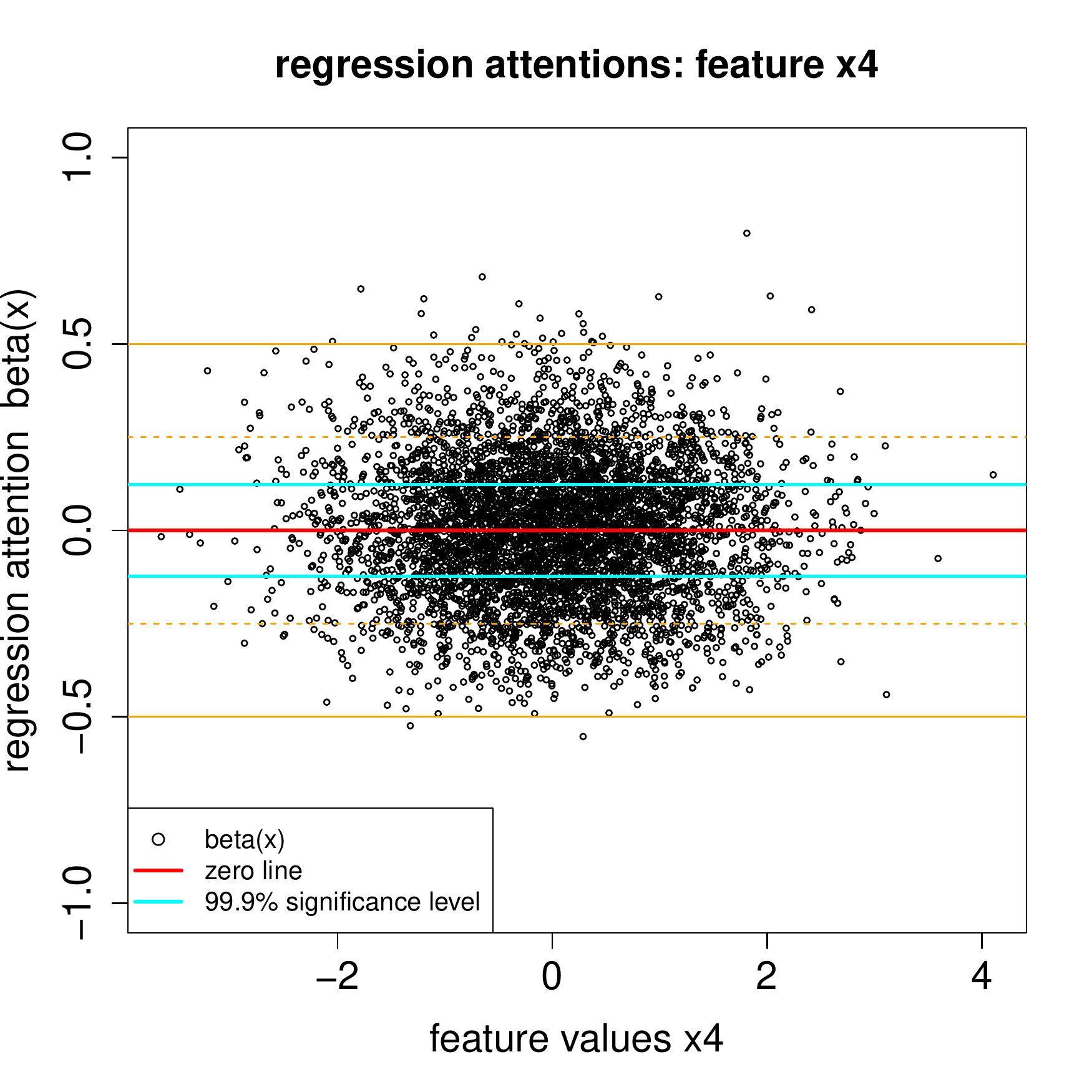}
\end{center}
\end{minipage}
\begin{minipage}[t]{0.32\textwidth}
\begin{center}
\includegraphics[width=.9\textwidth]{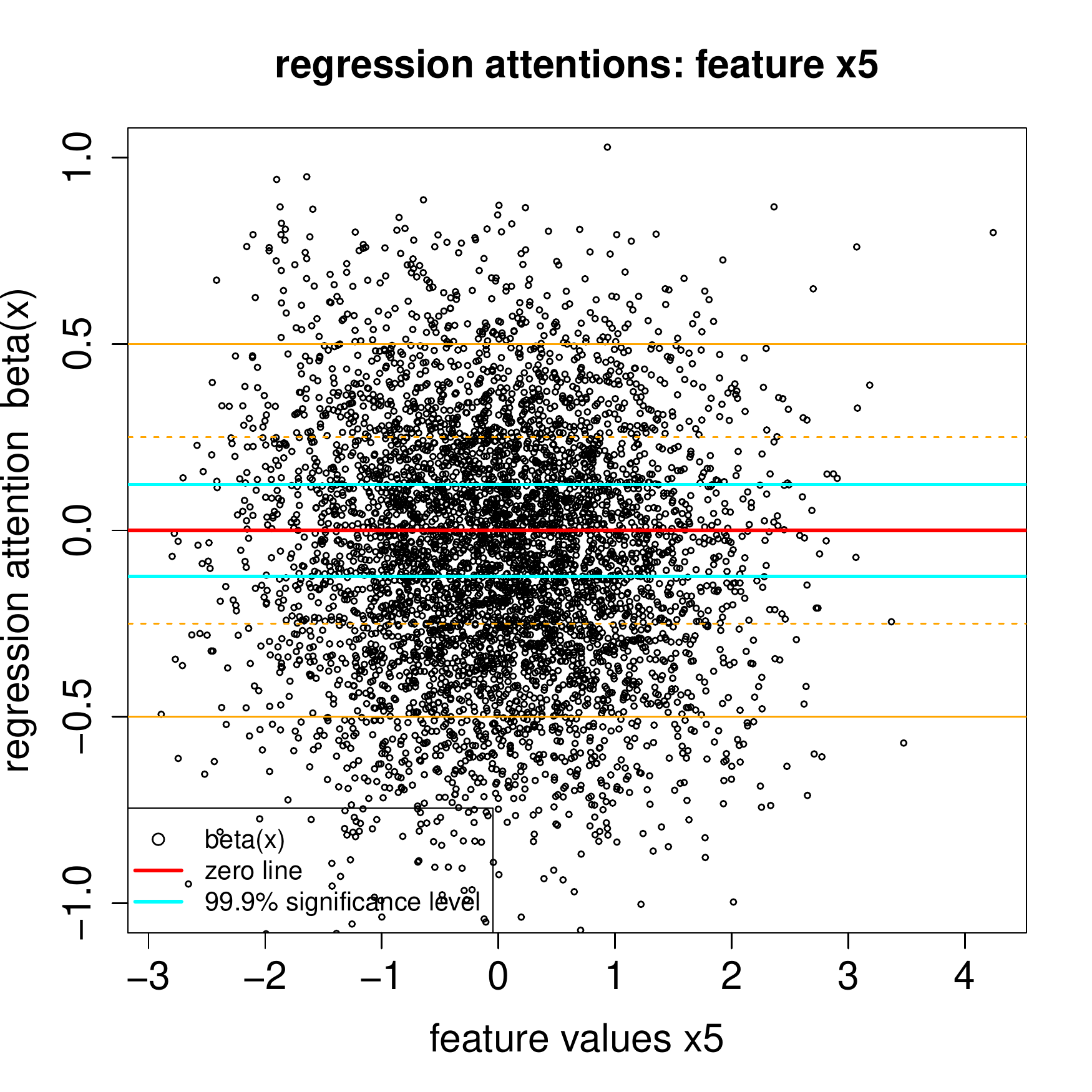}
\end{center}
\end{minipage}
\begin{minipage}[t]{0.32\textwidth}
\begin{center}
\includegraphics[width=.9\textwidth]{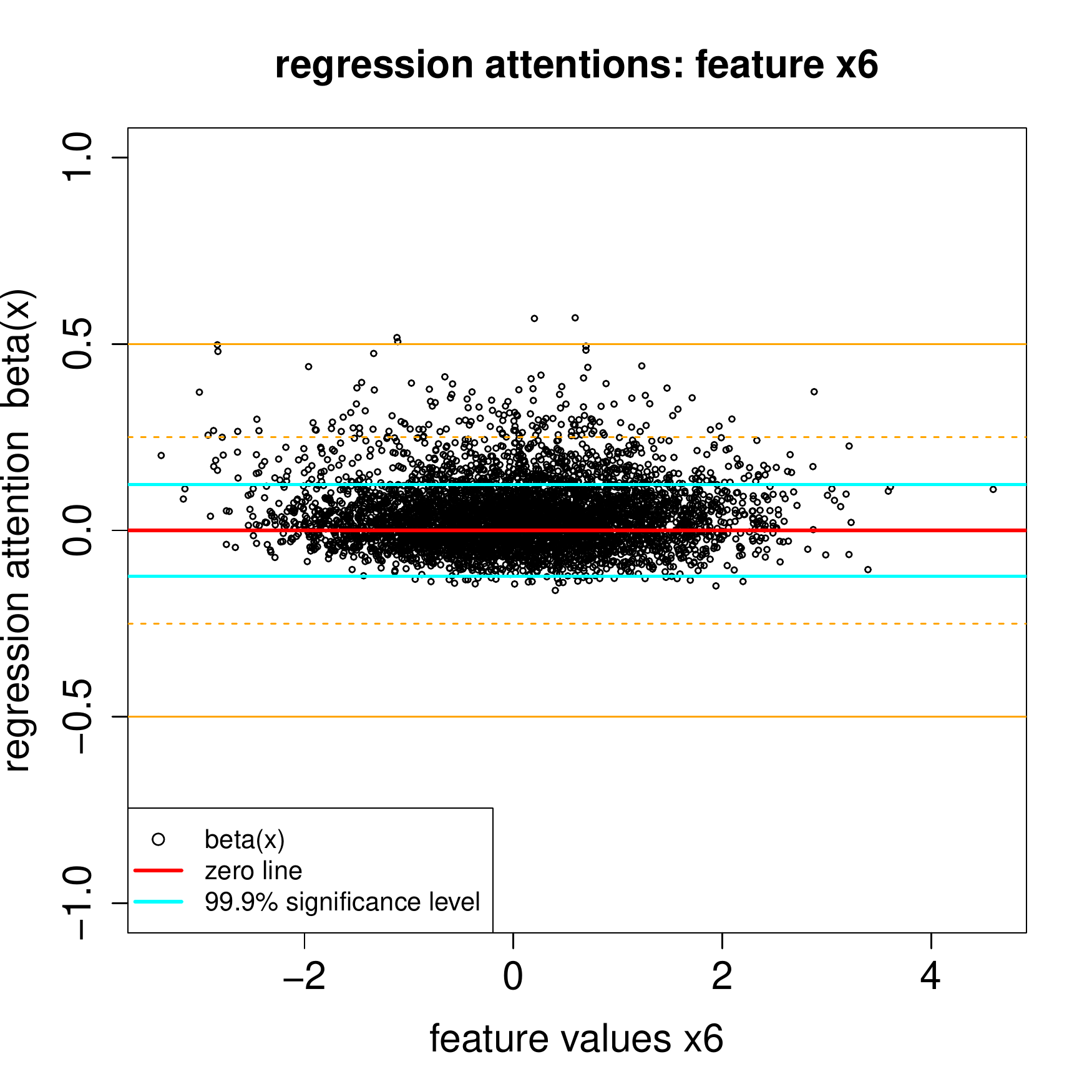}
\end{center}
\end{minipage}
\begin{minipage}[t]{0.32\textwidth}
\begin{center}
\includegraphics[width=.9\textwidth]{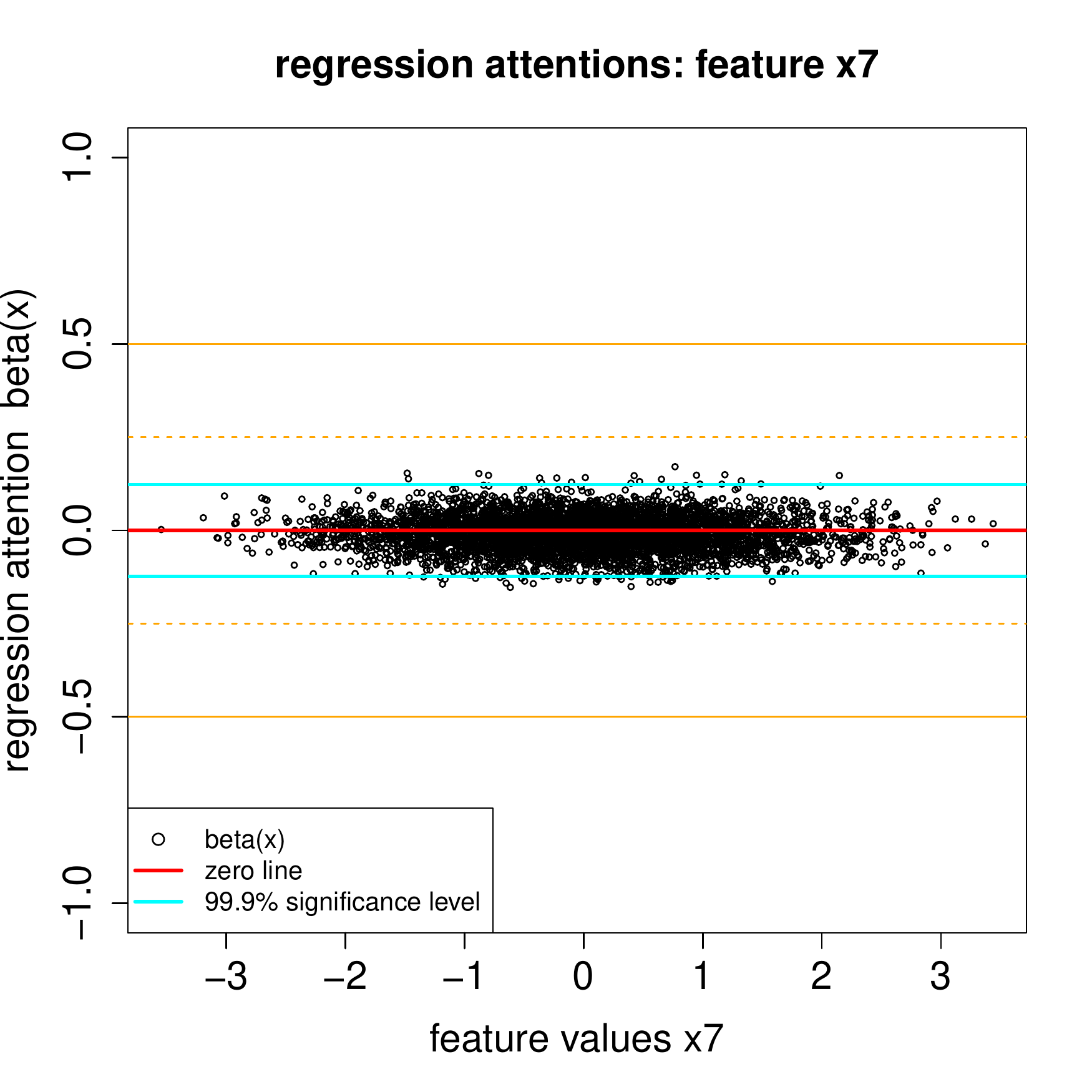}
\end{center}
\end{minipage}
\begin{minipage}[t]{0.32\textwidth}
\begin{center}
\includegraphics[width=.9\textwidth]{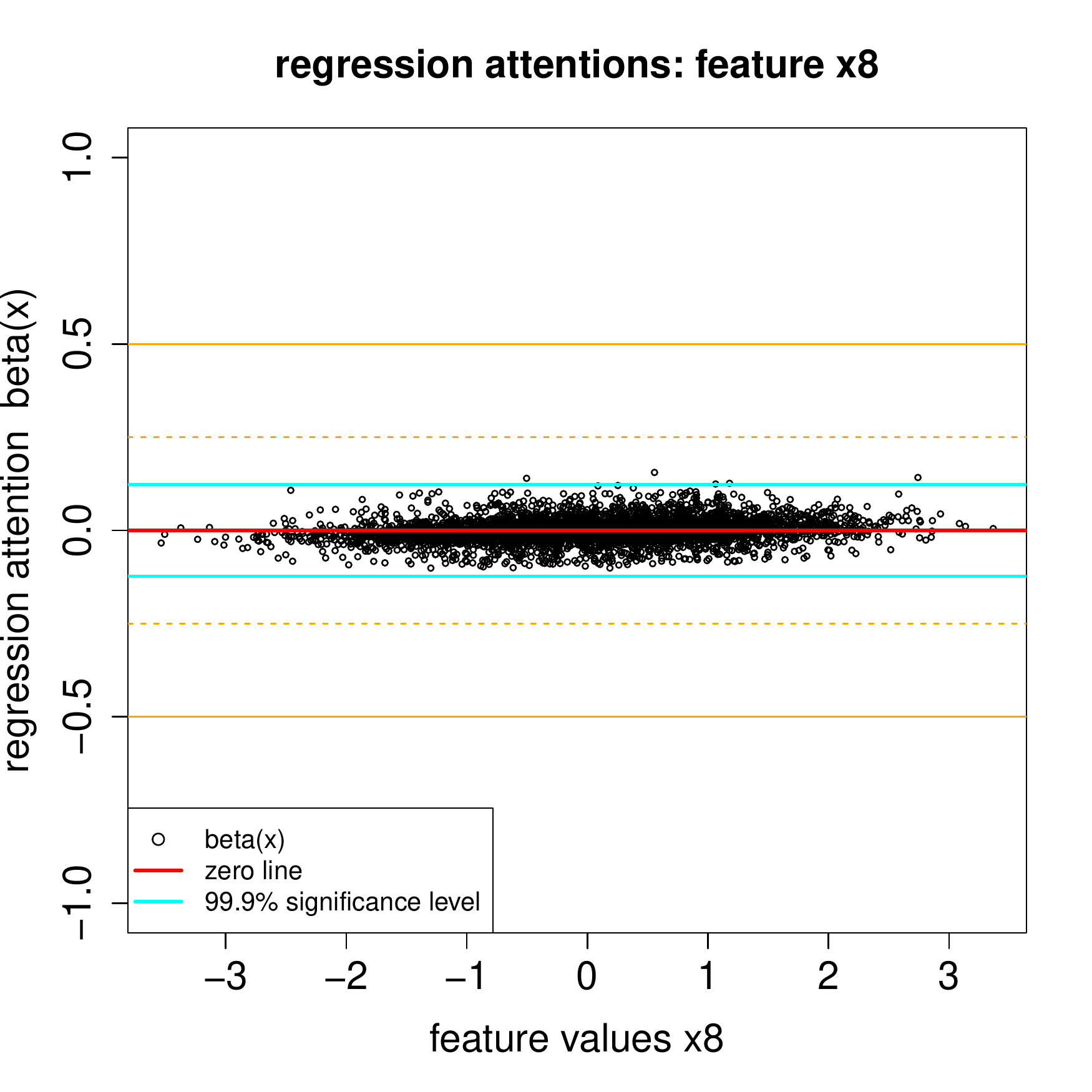}
\end{center}
\end{minipage}

\end{center}
\caption{Regression attentions $\widehat{\beta}_j(\bx_t)$, $1\le j \le q=8$, of $5,000$ randomly selected 
out-of-sample instances $\bx_t$ from ${\cal T}$; the $y$-scale is identical in all plots and on the $x$-scale we have $x_j$.}
\label{attentions weights synthetic}
\end{figure}

We are now in the situation where we can benefit from the LocalGLMnet architecture. This allows us
to study the estimated regression attentions and the resulting terms in the LocalGLMnet regression
function
\begin{equation*}
\bx \mapsto  \widehat{\beta}_j(\bx)
\qquad \text{ and } \qquad 
\bx \mapsto  \widehat{\beta}_j(\bx) x_j.
\end{equation*}
The interpretation to these terms has been given in Remarks \ref{interpretation of terms 2}. We use the 
code of Listing \ref{LocalGLMnetWeights1} to extract $\widehat{\bbeta}(\bx)$. These estimated
regression attentions $\widehat{\beta}_j(\bx)$
are illustrated in Figure \ref{attentions weights synthetic} for all components $1\le j \le q=8$.

We interpret Figure \ref{attentions weights synthetic}.
Regression attention $\widehat{\beta}_1(\bx)$ is concentrated around $1/2$ which describes
the first term in \eqref{true synthetic}. The regression attentions $\widehat{\beta}_2(\bx), \ldots, \widehat{\beta}_6(\bx)$
are quite different from 0 (red horizontal line) which indicates that $x_2,\ldots, x_6$ are important in the description of the
true regression function $\mu$. Finally, $\widehat{\beta}_7(\bx)$ and $\widehat{\beta}_8(\bx)$ are concentrated
around zero which indicates that feature information $x_7$ and $x_8$ may not be important for our regression function.
Thus, the last two variables could be dropped from the model, unless they play an important role in 
$\widehat{\bbeta}(\bx)$. This can be checked by just refitting the model without these variables.

\subsection{Variable selection}
\label{Variable selection}
In the previous example we have just said that we should drop variables $x_7$ and $x_8$ from the LocalGLMnet
regression because regression attentions $\widehat{\beta}_7(\bx)$ and $\widehat{\beta}_8(\bx)$  spread
around zero. Obviously, these two estimators
should be identically equal to zero because they do not appear in the true regression function $\mu(\cdot)$, but the noise in the data $Y_i$
is letting their estimators fluctuate around zero. This fluctuation is of comparable size for both 
$x_7$ (which is independent of all other variables) and $x_8$ (which is correlated with $x_2$). The main question
is: how much fluctuation around zero is still acceptable for allowing to drop a variable, i.e., does not reject the 
null hypothesis $H_0$
of setting ${\beta}_7(\bx)=0$, or how much fluctuation reveals real regression structure?
In GLMs this question is answered by either the Wald test or the likelihood ratio test (LRT) which use
asymptotic normality results of MLEs, see Section 2.2.2 in Fahrmeir--Tutz \cite{FahrmeirTutz}.
Here, we cannot rely on an asymptotic theory for MLEs because early stopping implies that we do 
not consider the MLE. An analysis of the results also shows that the volatility in 
$\widehat{\beta}_7(\bx)$ is bigger than the magnitudes used in the Wald test and the LRT. For these reason,
we propose an empirical way of determining the rejection region of the null hypothesis 
$H_0: {\beta}_7(\bx)=0$.

If we are given a statistical problem with features $\bx=(x_1,\ldots, x_q)^\top \in \R^q$, we propose to extend these features
by an additional variable $x_{q+1}$ which is completely random, independent of $\bx$ and which, of course, does
not enter the true (unknown) regression function $\mu(\bx)$ but is only included within the network. This additional random component $x_{q+1}$ will quantify the resulting fluctuations
in $\widehat{\beta}_{q+1}(\bx)$ of an independent component that does not enter the regression function.
In order to successfully apply this empirical test we need to normalize all feature components $x_j$, $1\le j \le q+1$, to have
zero empirical mean and unit variance, i.e., they should all live on the same scale. Note that such a normalization
should already have been done for successful SGD fitting, thus, this does not impose an additional step, here.
We then fit a LocalGLMnet \eqref{LocalGLMnet weight}-\eqref{LocalGLMnet} to the learning data ${\cal L}$ with
extended features $\bx^+=(\bx^\top,x_{q+1})^\top \in \R^{q+1}$ which gives us the estimated regression attentions
$\widehat{\beta}_1(\bx^+_i), \ldots, \widehat{\beta}_{q+1}(\bx^+_i)$. Using these estimated regression attentions
we receive empirical mean and standard deviation for the additional component
\begin{equation}\label{Wald test}
\bar{b}_{q+1} = \frac{1}{n} \sum_{i=1}^n \widehat{\beta}_{q+1}(\bx^+_i) 
\qquad \text{ and } \qquad 
\widehat{s}_{q+1} = \sqrt{\frac{1}{n-1} \sum_{i=1}^n \left(\widehat{\beta}_{q+1}(\bx^+_i)-\bar{b}_{q+1} \right)^2}.
\end{equation}
Since this additional component $x_{q+1}$ does not enter the true regression function we expect $\bar{b}_{q+1} \approx 0$,
and $\widehat{s}_{q+1}$ quantifies the expected fluctuation around zero.

The null hypothesis $H_0: {\beta}_j(\bx)=0$ for component $j$ on significance level $\alpha \in (0,1/2)$ can then
be rejected if the coverage ratio of the following
centered interval $I_\alpha$ for $\widehat{\beta}_j(\bx^+_i)$, $1\le i \le n$, 
\begin{equation}\label{empirical confidence bounds}
I_\alpha=\Big[ q_{\cal N}(\alpha/2)\cdot \widehat{s}_{q+1}  ,  ~q_{\cal N}(1-\alpha/2)\cdot \widehat{s}_{q+1} \Big]
=\Big[ q_{\cal N}(\alpha/2)\cdot \widehat{s}_{q+1}  , ~- q_{\cal N}(\alpha/2)\cdot \widehat{s}_{q+1} \Big]
\end{equation}
is substantially smaller than $\alpha$, where $q_{\cal N}(p)$ denotes the quantile of the standard Gaussian
distribution on quantile level $p \in (0,1)$.

We come back to our Figure \ref{attentions weights synthetic}. We do not add an additional component $x_{q+1}$,
but we directly use component $x_7$ to test the null hypotheses 
$H_0: {\beta}_j(\bx)=0$ for the remaining components $j \neq 7$. In our synthetic data example we have
\begin{equation*}
\bar{b}_{7} =  -0.0068 \approx 0
\qquad \text{ and } \qquad 
\widehat{s}_{7} = 0.0461.
\end{equation*}
We choose significance level $\alpha=0.1\%$ which provides us with $q_{\cal N}(0.05\%)=3.2905$.
The cyan lines in Figure \ref{attentions weights synthetic} show the resulting
interval $I_\alpha$ given in \eqref{empirical confidence bounds} for our example. Only for $x_8$ the black dots
$\widehat{\beta}_8(\bx_t)$ are within these confidence bounds $I_\alpha$ which implies that we should drop this
component and keep components $x_1,\ldots, x_6$ in the regression model. In a final step, the model with
dropped components should be re-fitted and the out-of-sample loss should not substantially change, this
re-fitting step verifies that the dropped components also do not play a significant role in the
regression attentions $\widehat{\beta}_j(\bx)$ of the remaining feature components $j$, i.e., contribute
by interacting with other variables.

\begin{figure}[htb!]
\begin{center}
\begin{minipage}[t]{0.32\textwidth}
\begin{center}
\includegraphics[width=.9\textwidth]{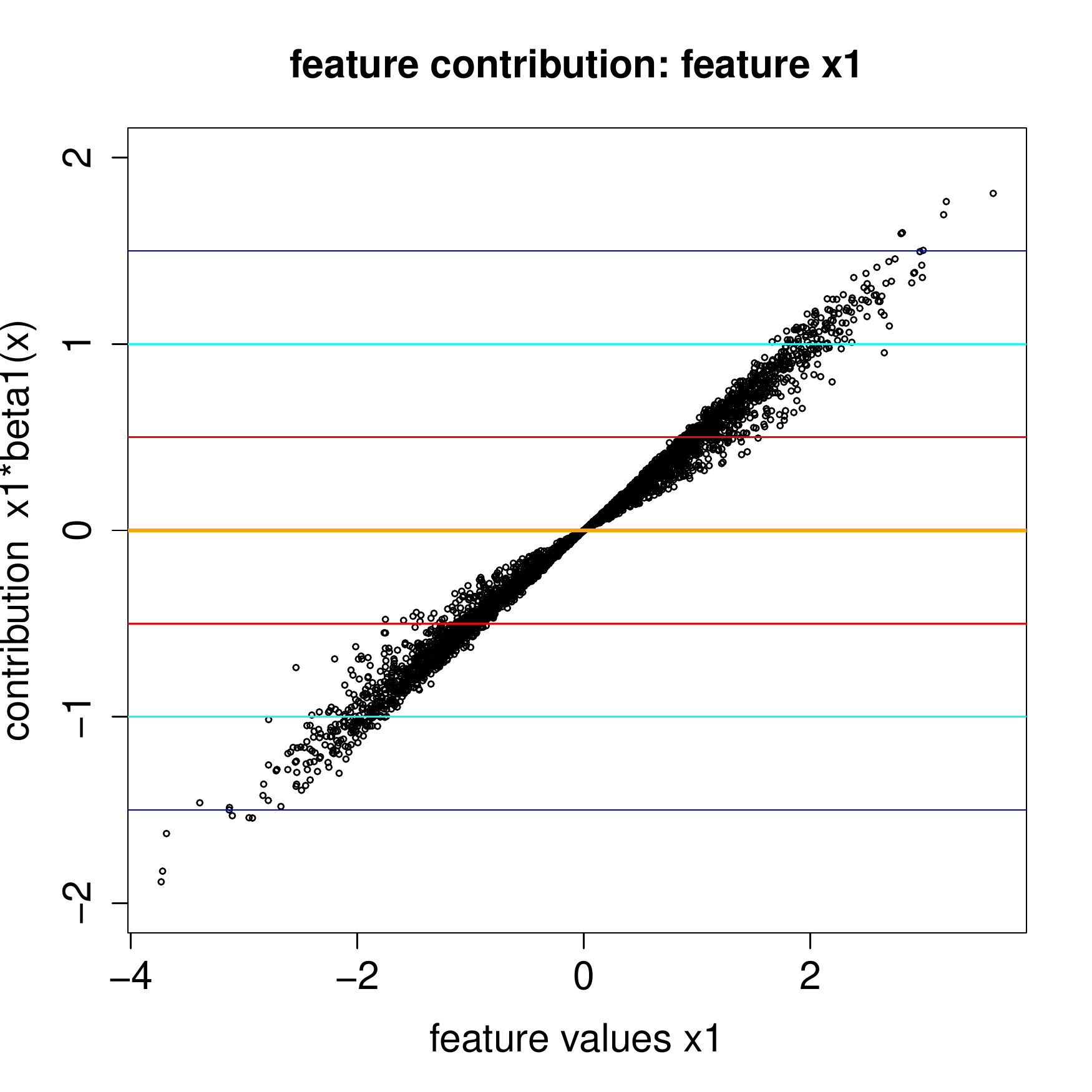}
\end{center}
\end{minipage}
\begin{minipage}[t]{0.32\textwidth}
\begin{center}
\includegraphics[width=.9\textwidth]{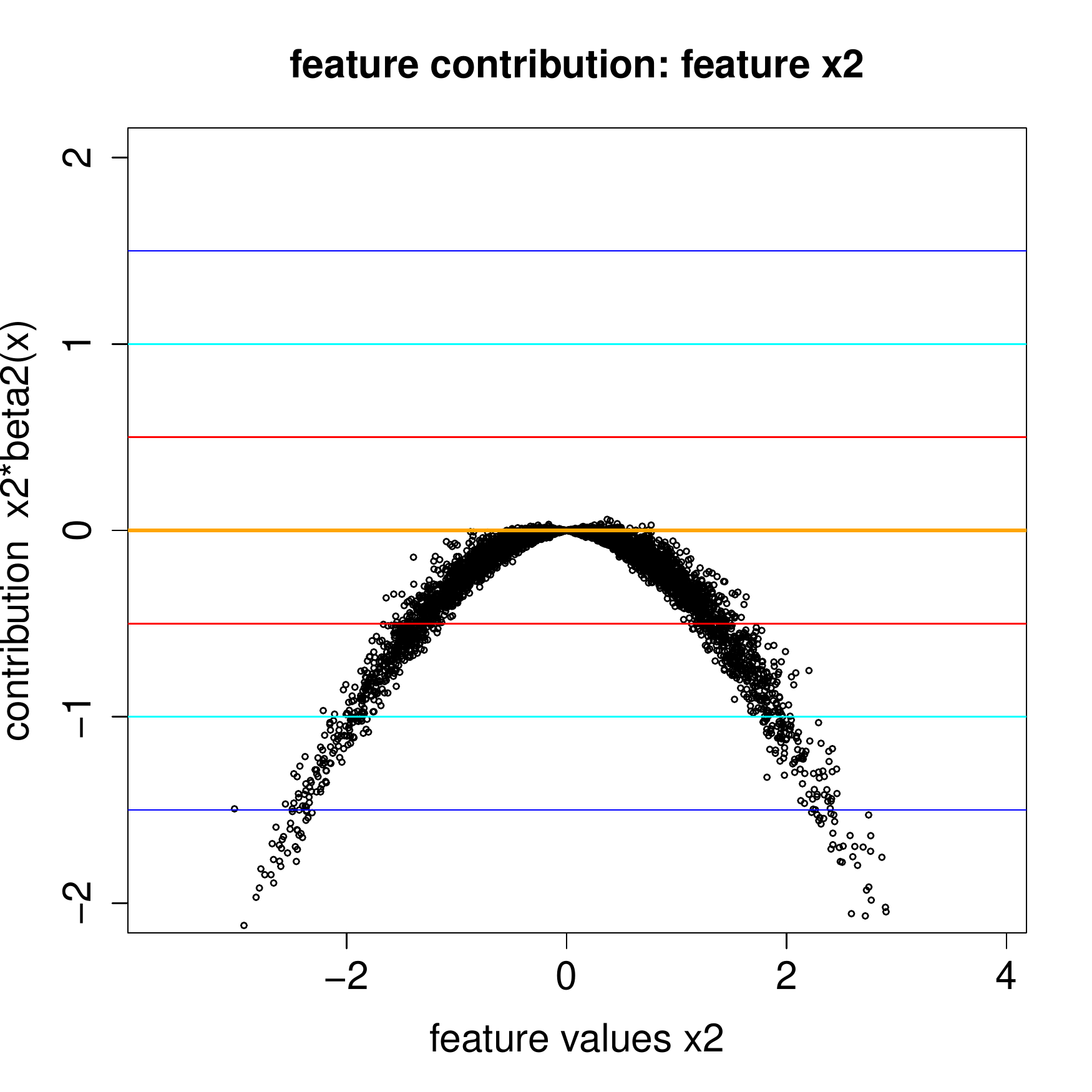}
\end{center}
\end{minipage}
\begin{minipage}[t]{0.32\textwidth}
\begin{center}
\includegraphics[width=.9\textwidth]{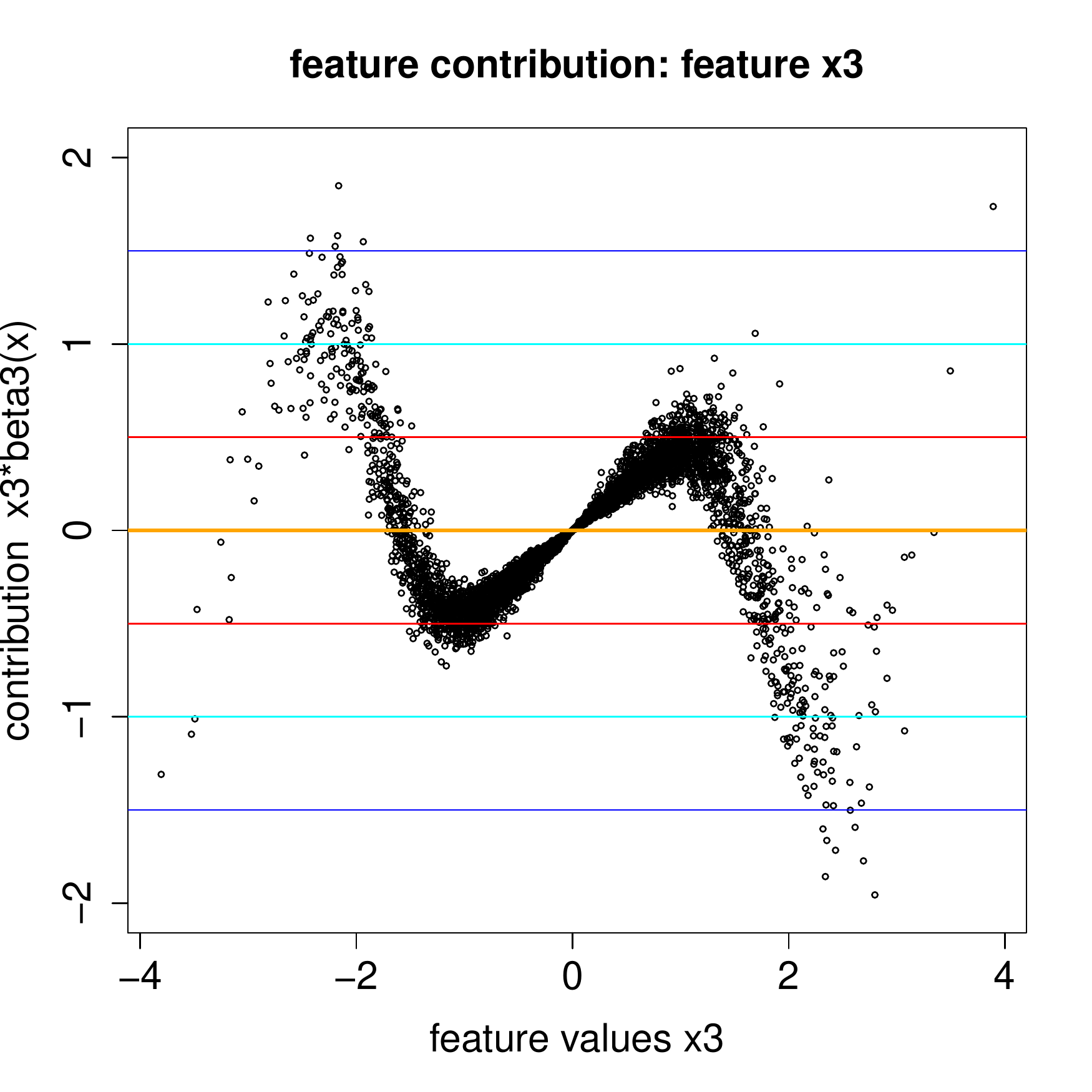}
\end{center}
\end{minipage}
\begin{minipage}[t]{0.32\textwidth}
\begin{center}
\includegraphics[width=.9\textwidth]{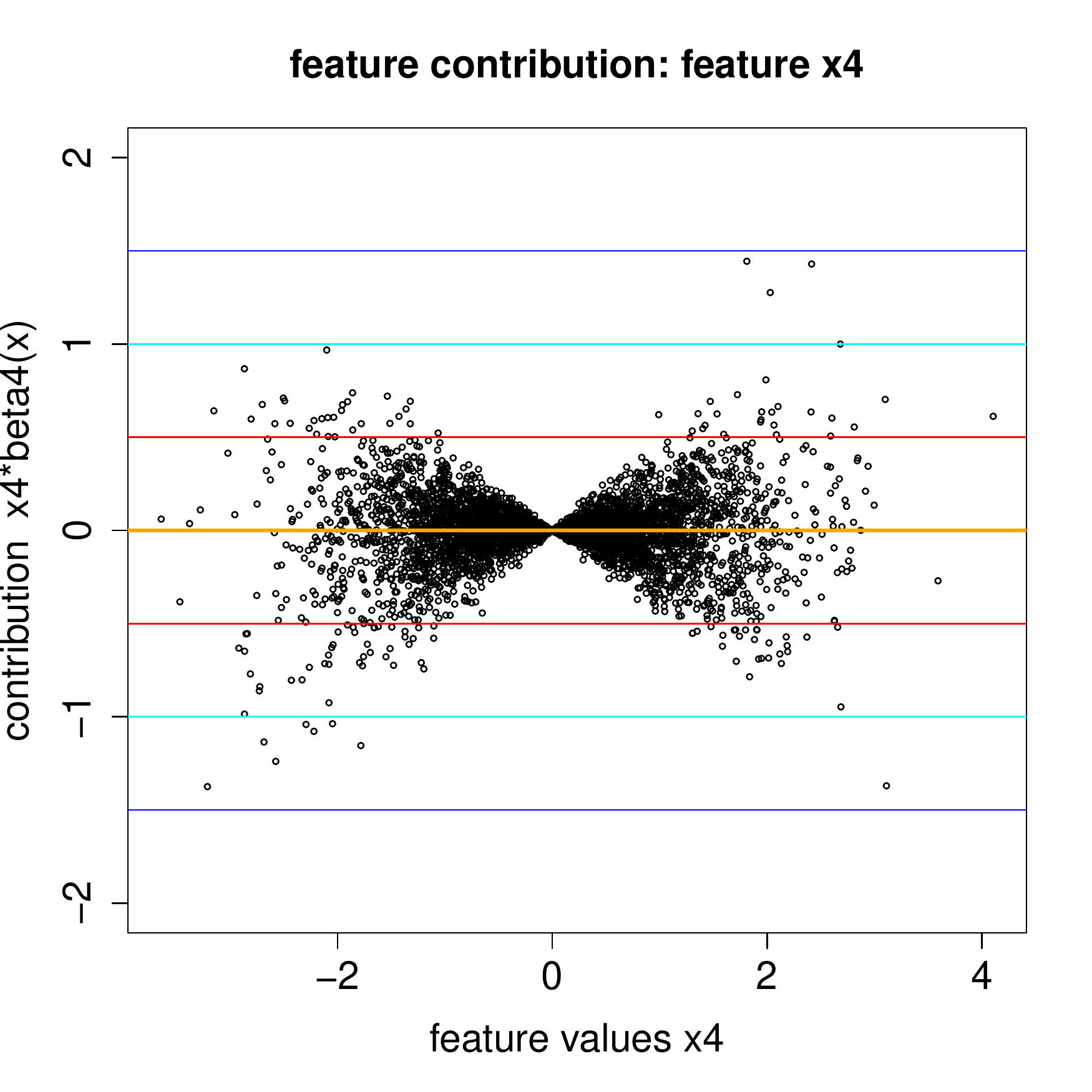}
\end{center}
\end{minipage}
\begin{minipage}[t]{0.32\textwidth}
\begin{center}
\includegraphics[width=.9\textwidth]{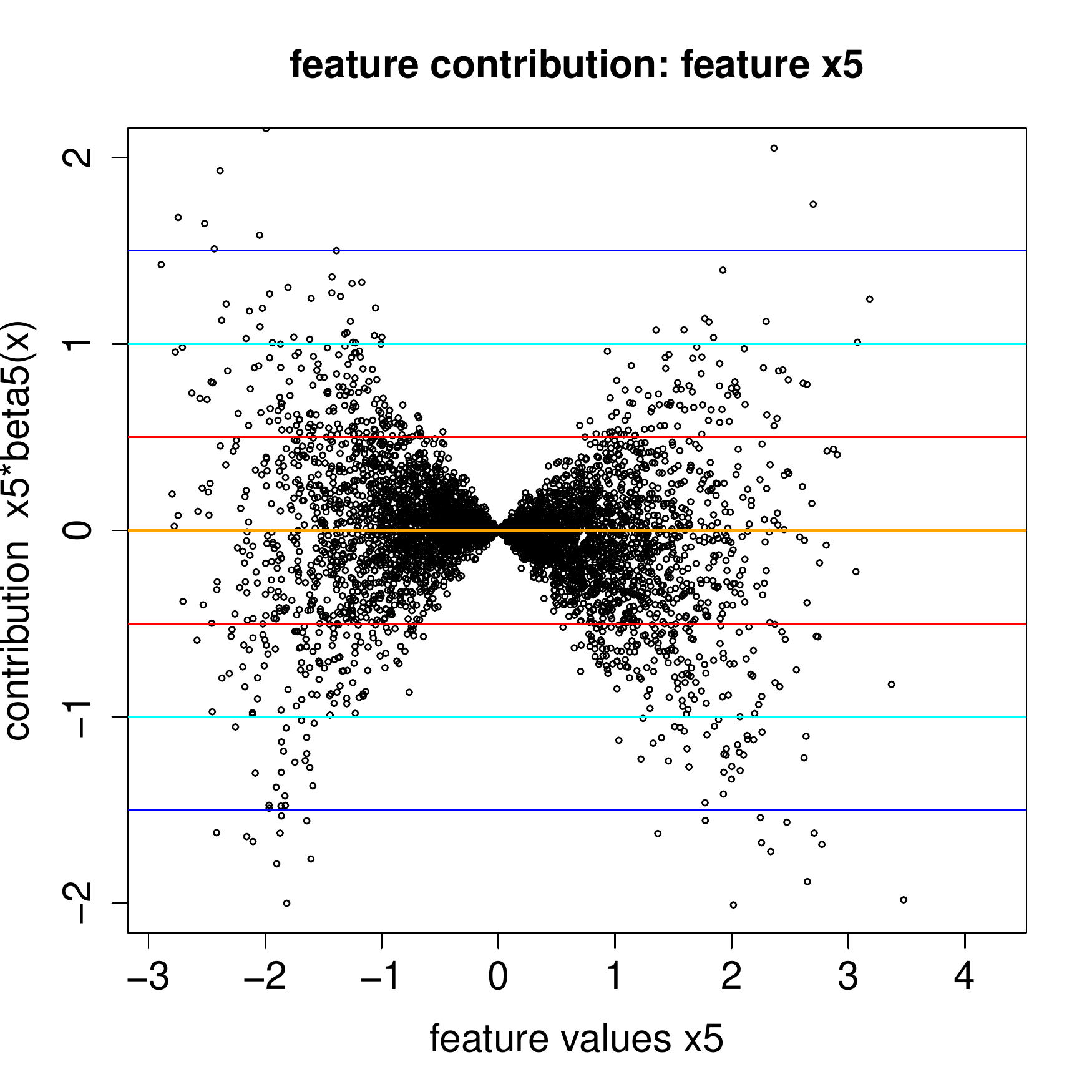}
\end{center}
\end{minipage}
\begin{minipage}[t]{0.32\textwidth}
\begin{center}
\includegraphics[width=.9\textwidth]{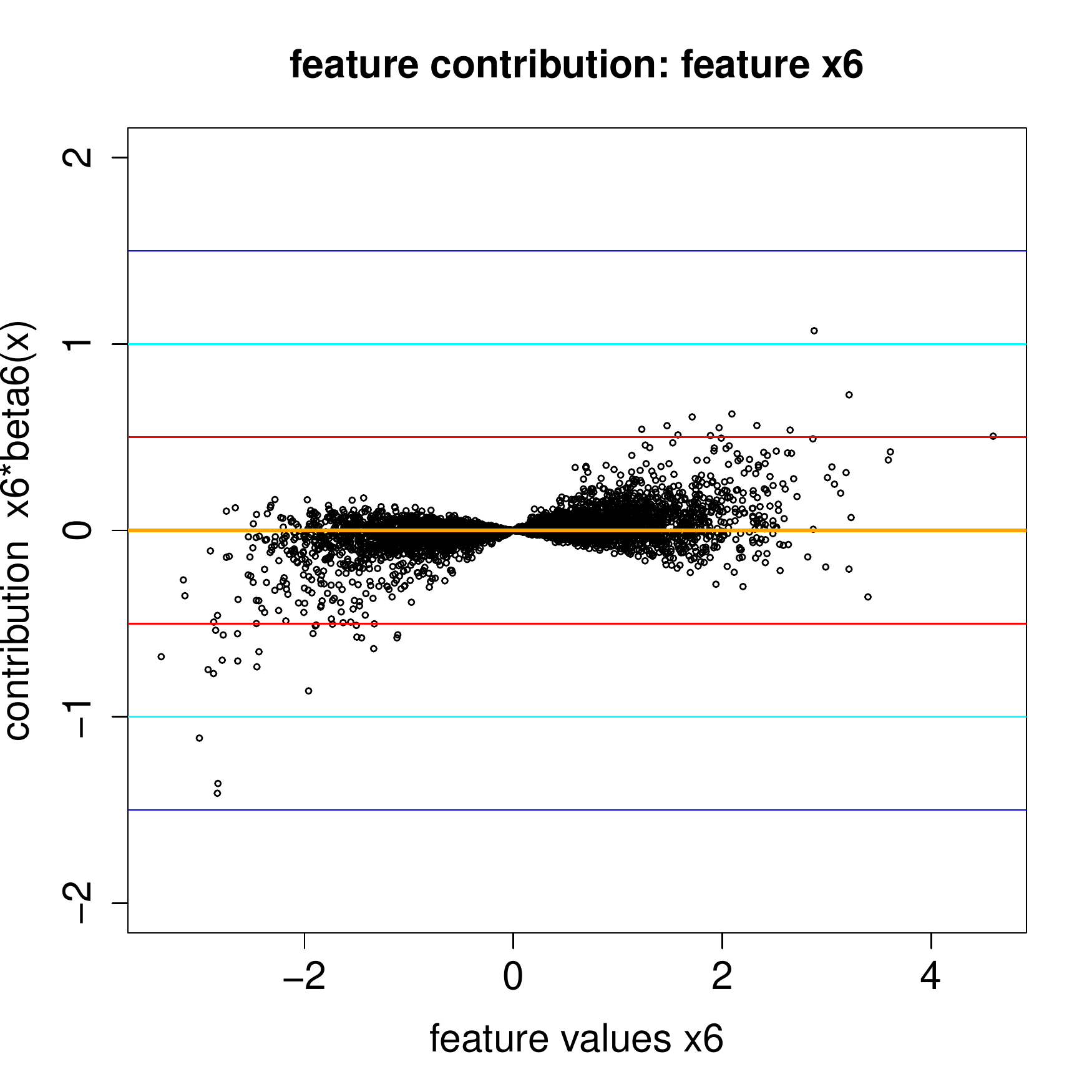}
\end{center}
\end{minipage}
\begin{minipage}[t]{0.32\textwidth}
\begin{center}
\includegraphics[width=.9\textwidth]{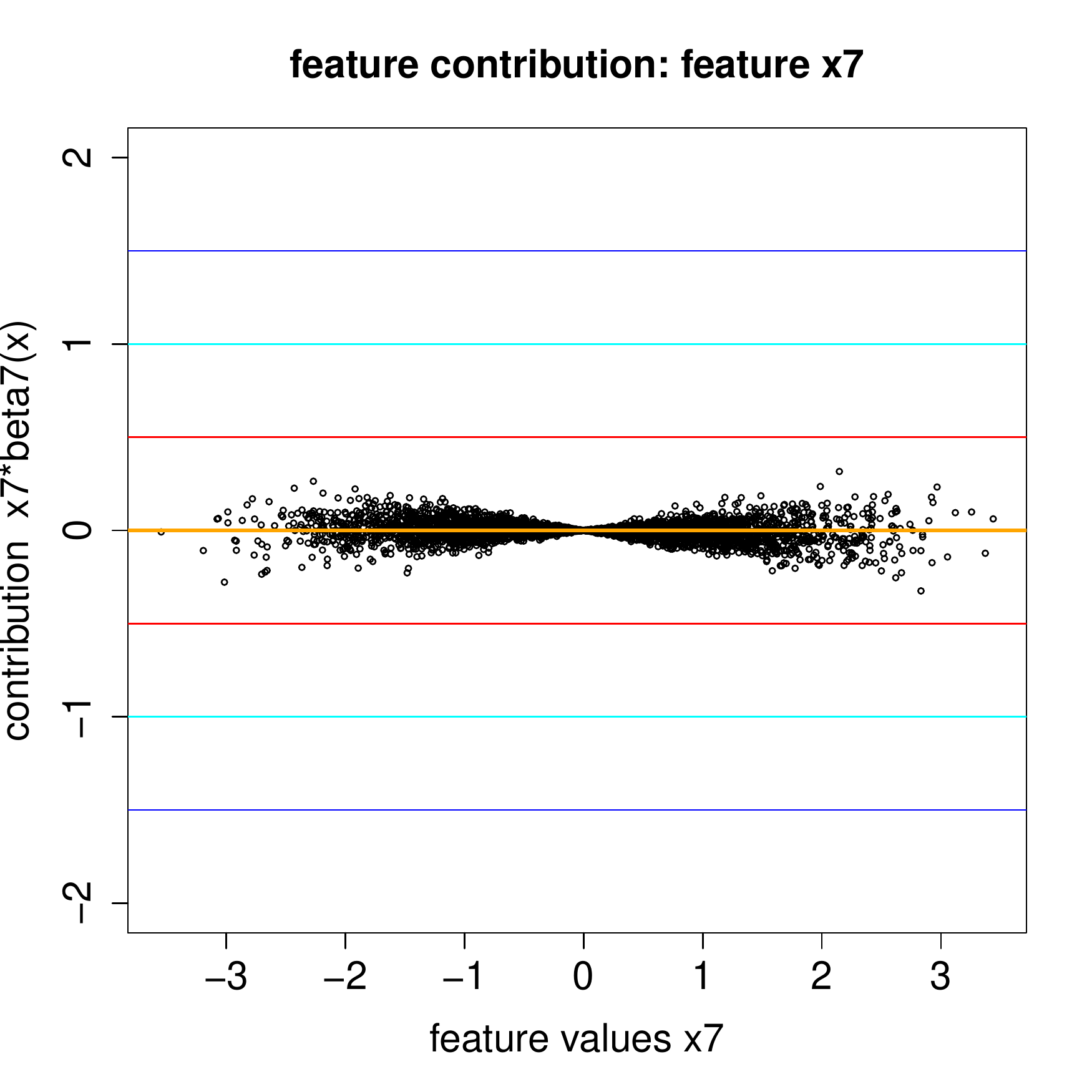}
\end{center}
\end{minipage}
\begin{minipage}[t]{0.32\textwidth}
\begin{center}
\includegraphics[width=.9\textwidth]{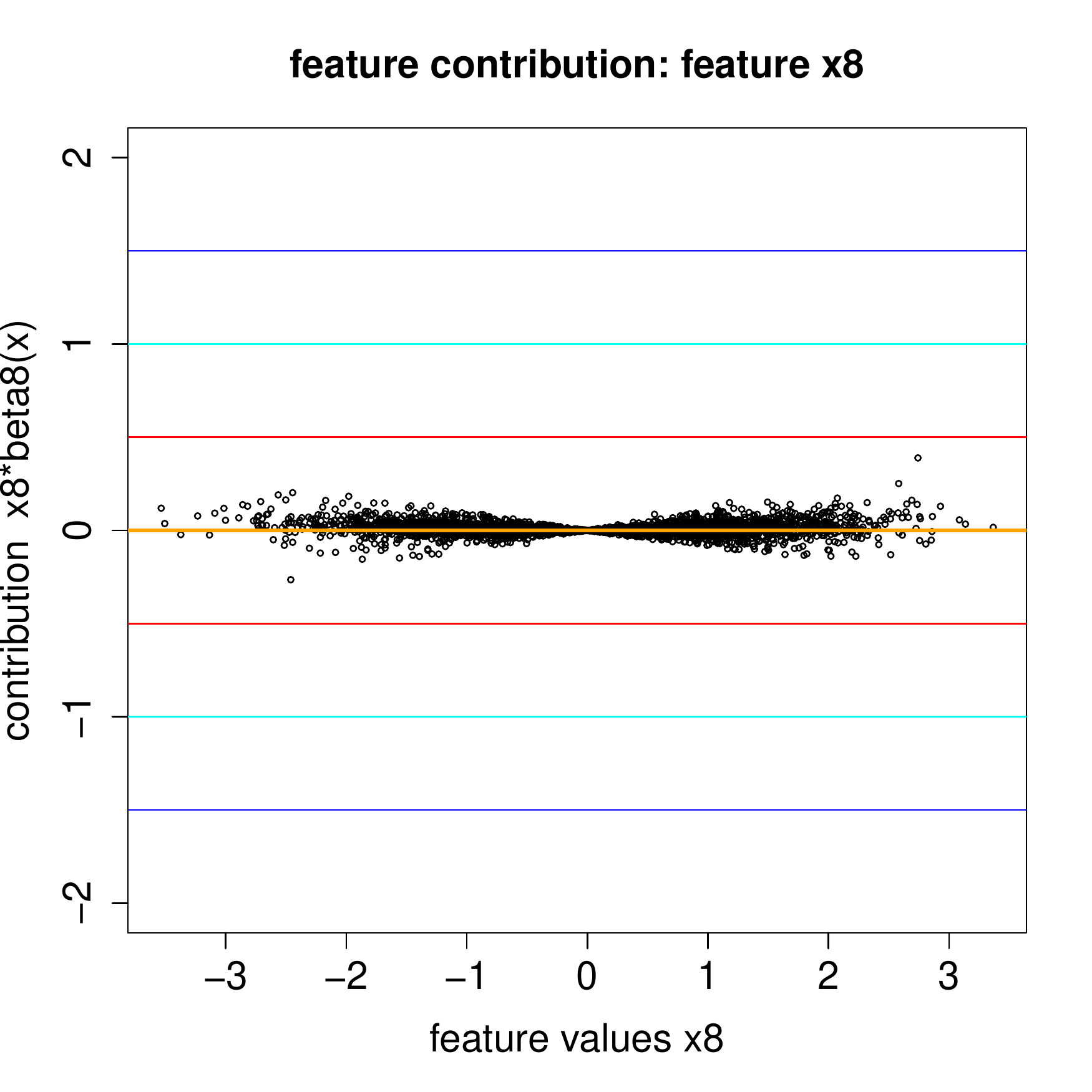}
\end{center}
\end{minipage}

\end{center}
\caption{Feature contributions $\widehat{\beta}_j(\bx_t)x_{t,j}$, $1\le j \le q=8$, to the LocalGLMnet estimated
regression function $\widehat{\mu}(\bx_t)$ of $5,000$ randomly selected 
out-of-sample instances $\bx_t$ from ${\cal T}$; the $y$-scale is identical in all plots and on the $x$-scale we have $x_j$.}
\label{attentions weights synthetic 2}
\end{figure}

Figure \ref{attentions weights synthetic 2} gives the resulting 
feature contributions $\widehat{\beta}_j(\bx_t)x_{t,j}$, $1\le j \le q=8$, to the LocalGLMnet estimated
regression function $\widehat{\mu}(\bx_t)$. We clearly see the linear term in $x_1$, the quadratic term
in $x_2$ and the sine term in $x_3$ (first line of Figure \ref{attentions weights synthetic 2}), see also
\eqref{true synthetic} for the true regression function $\mu$. The second line of Figure \ref{attentions weights synthetic 2}
shows the interacting feature components $x_4$, $x_5$ and $x_6$, and the last line those that should be dropped.

\subsection{Interactions}
\label{Interactions sections}
In the next and final step we explore interactions between different feature components. This is based
on analyzing the gradients $\nabla \beta_j(\bx)$ for $1\le j \le q$, see \eqref{gradient}.
The $j$-th component of this gradient $\nabla \beta_j(\bx)$ explores whether we have a linear term in $x_j$
or not. If there are no interactions of the $j$-th component
with other components, i.e.~$\beta_j(\bx)x_j =\beta_j(x_j)x_j$,  it will exactly provide
us with the right functional form of $\beta_j(\bx)$ since in that case $\partial \beta_j(\bx)/\partial x_{j'}=0$ for all
$j'\neq j$. In relation to Figure \ref{attentions weights synthetic} this means that the scatter plot resembles a line
that does not have any lateral dilation. In Figure \ref{attentions weights synthetic} this is the case
for components $x_1,x_2,x_7,x_8$, for component $x_3$ this is not completely clear from the scatter plot,
and $x_4,x_5,x_7$ show lateral extensions indicating interactions.
In the latter general case we have $\partial \beta_j(\bx)/\partial x_{j'}\neq 0$ for some $j'\neq j$.

\begin{figure}[htb!]
\begin{center}
\begin{minipage}[t]{0.45\textwidth}
\begin{center}
\includegraphics[width=.9\textwidth]{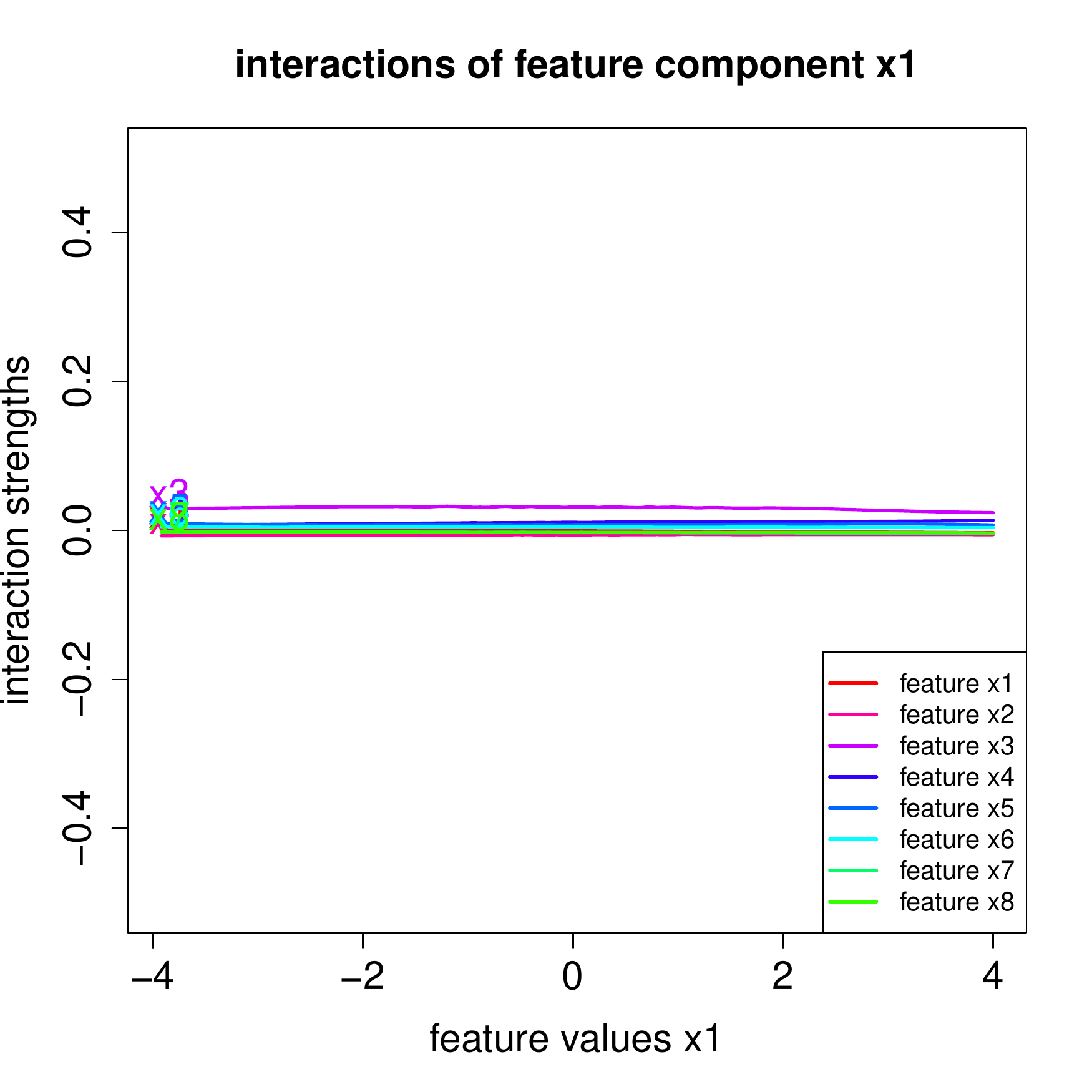}
\end{center}
\end{minipage}
\begin{minipage}[t]{0.45\textwidth}
\begin{center}
\includegraphics[width=.9\textwidth]{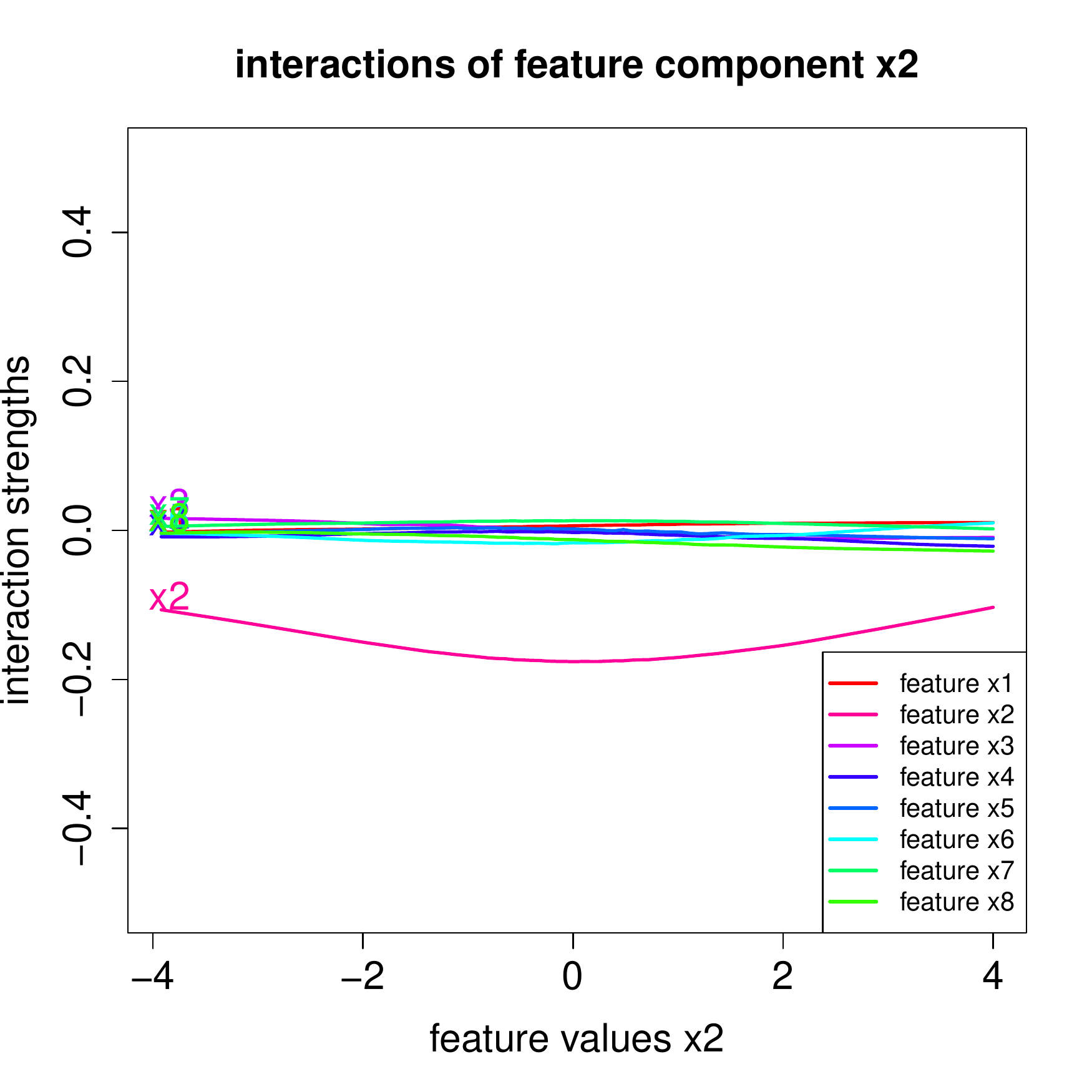}
\end{center}
\end{minipage}
\begin{minipage}[t]{0.45\textwidth}
\begin{center}
\includegraphics[width=.9\textwidth]{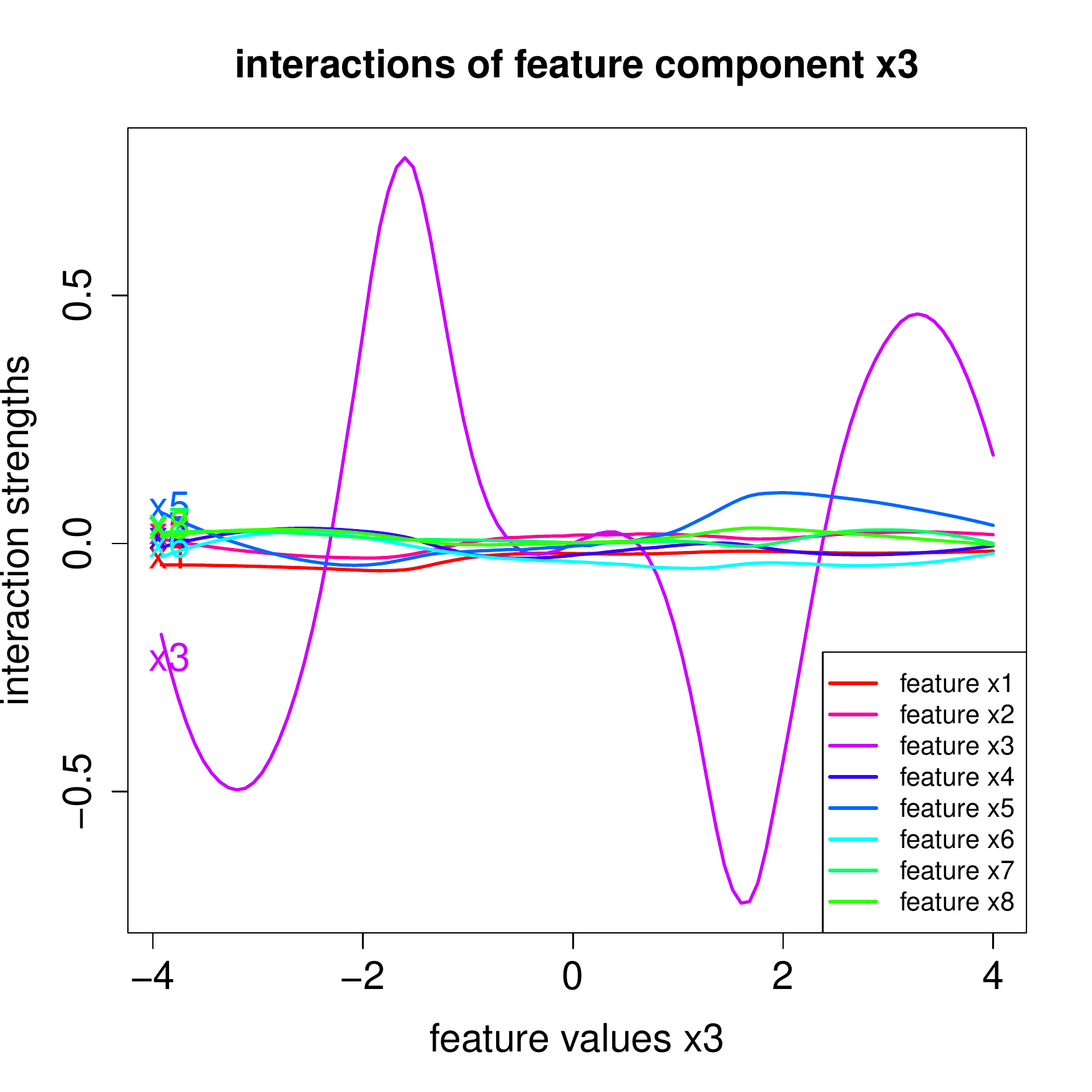}
\end{center}
\end{minipage}
\begin{minipage}[t]{0.45\textwidth}
\begin{center}
\includegraphics[width=.9\textwidth]{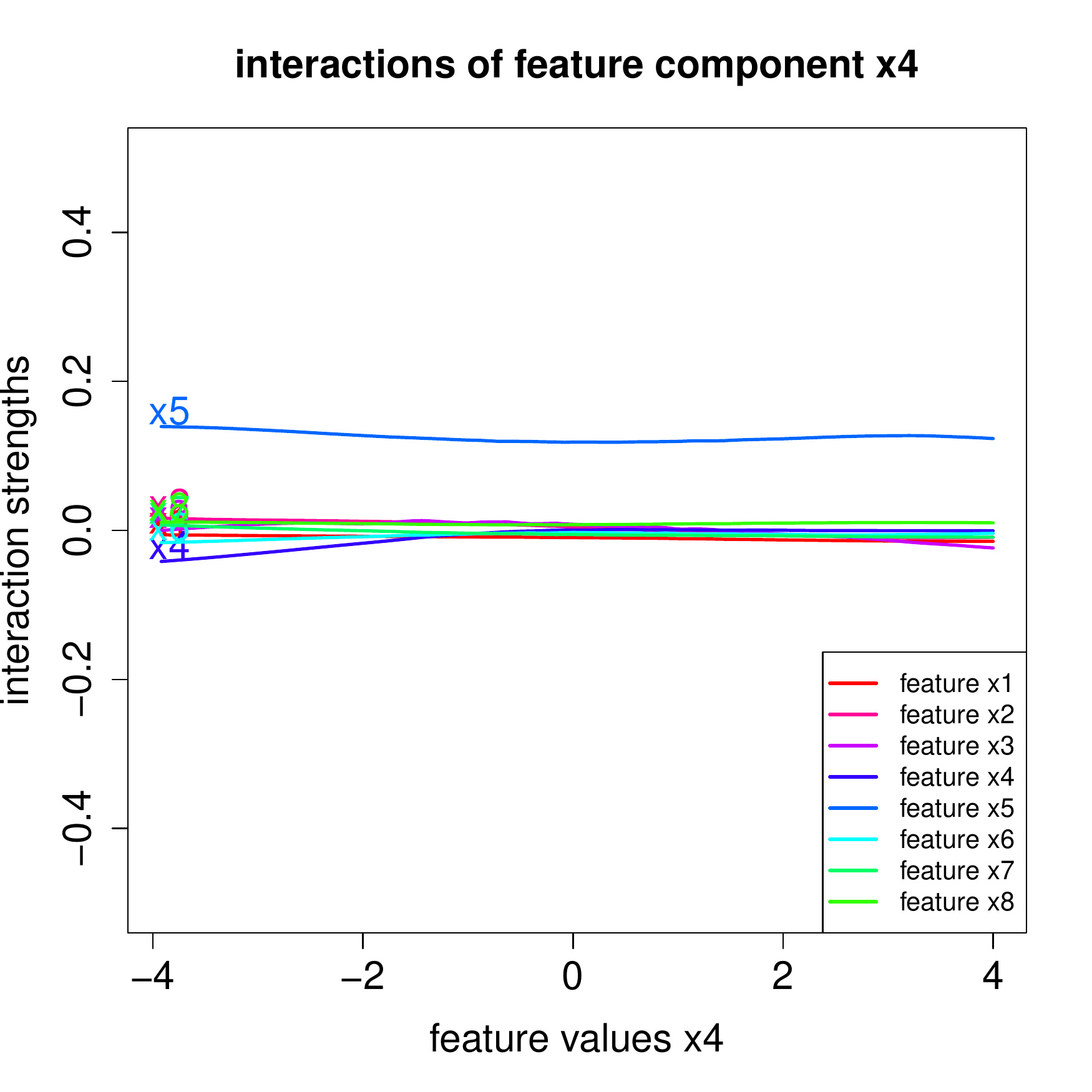}
\end{center}
\end{minipage}
\begin{minipage}[t]{0.45\textwidth}
\begin{center}
\includegraphics[width=.9\textwidth]{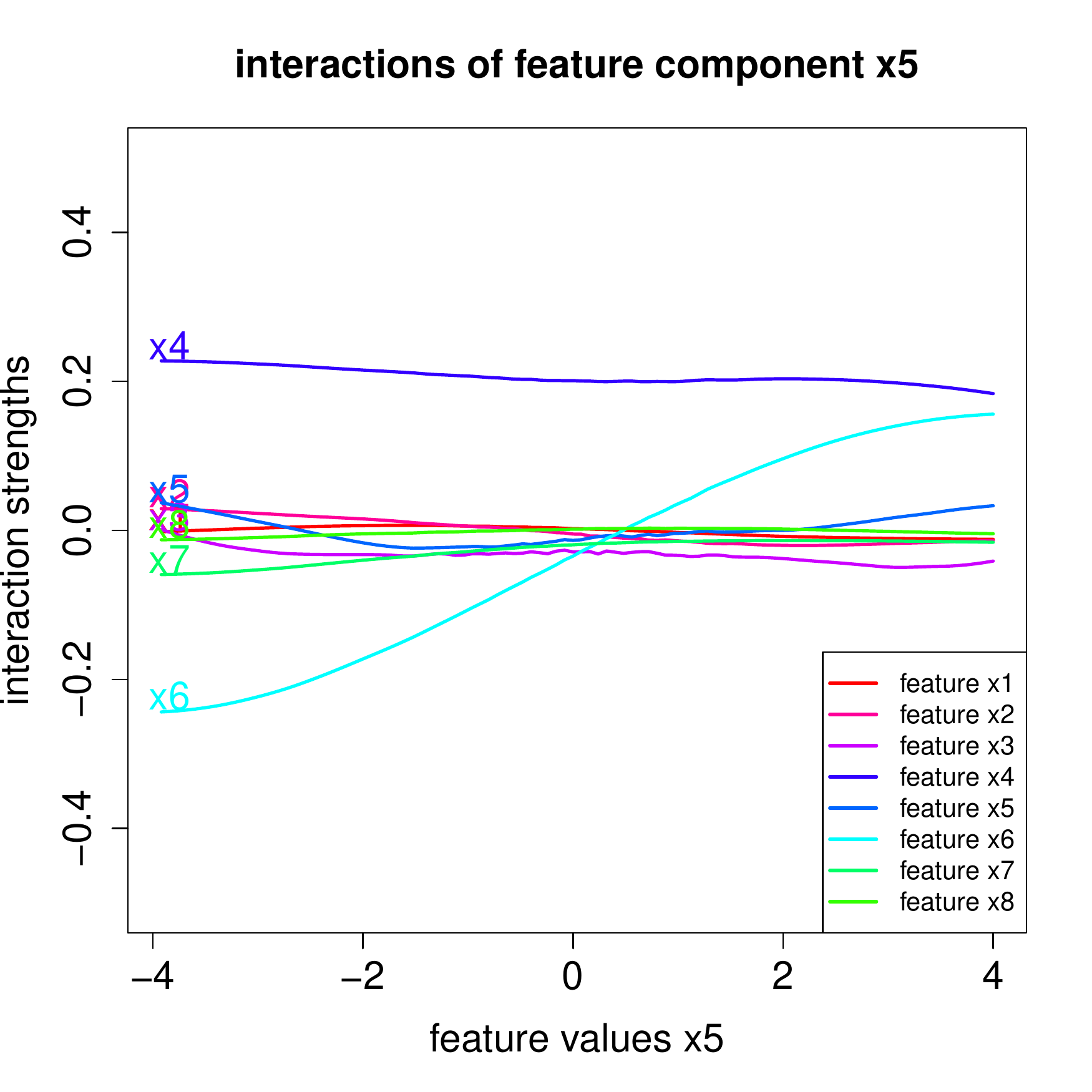}
\end{center}
\end{minipage}
\begin{minipage}[t]{0.45\textwidth}
\begin{center}
\includegraphics[width=.9\textwidth]{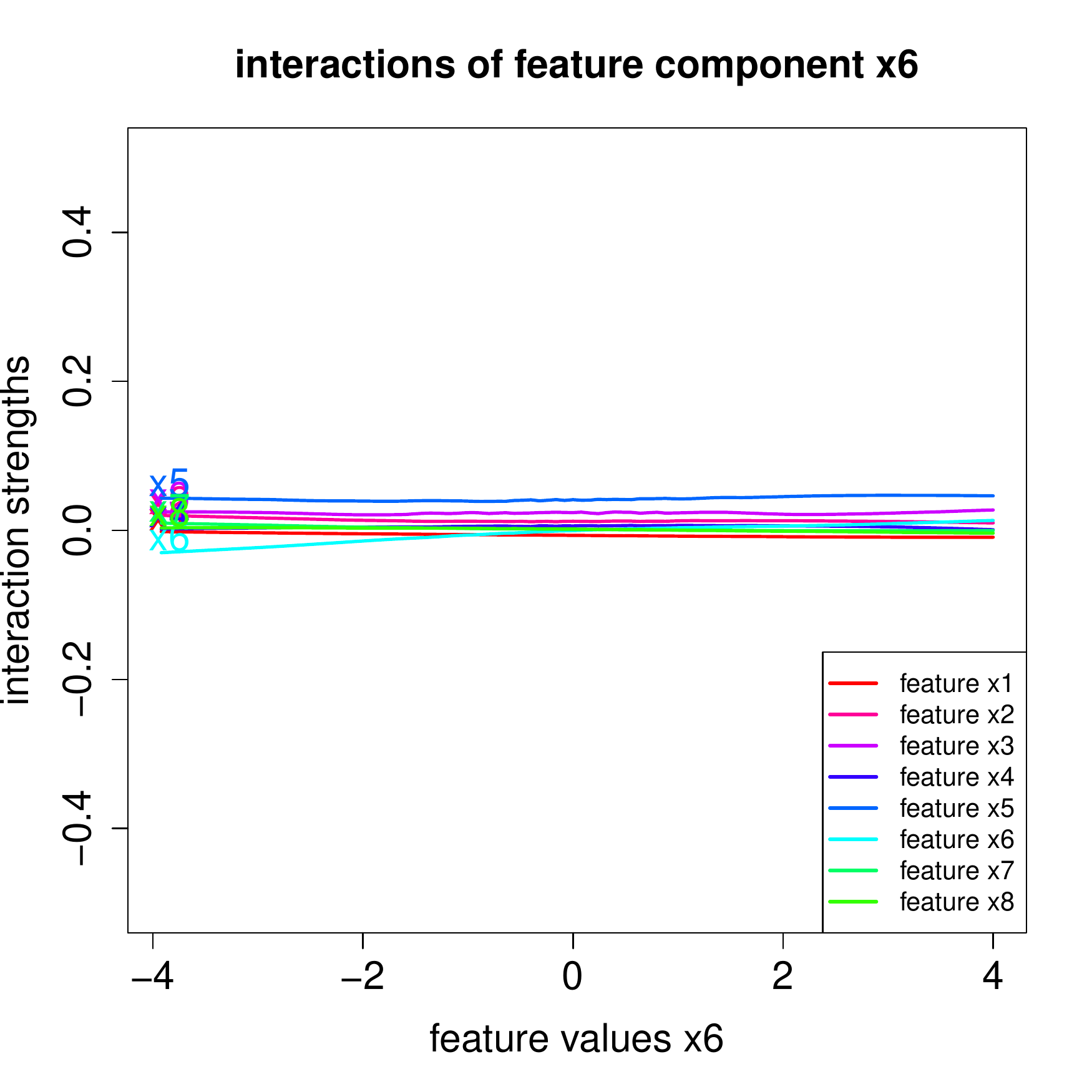}
\end{center}
\end{minipage}
\end{center}
\caption{Spline fits to the sensitivities $\partial_{ x_k} \widehat{\beta}_j(\bx_i)$, $1 \le j,k \le 6$,
over all instances $i=1,\ldots, n$.}
\label{interactions synthetic}
\end{figure}

We calculate the gradients $\nabla \widehat{\beta}_j(\bx)$, $1\le j \le q$, of the fitted
model. These can be obtained by the {\sf R} code of Listing \ref{GradientListing}.
This provides us for fixed $j$ with vectors $\nabla \widehat{\beta}_j(\bx_i)$ for all
instances $i=1,\ldots, n$. In order to analyze these gradients we fit a spline to these
observations by regressing 
\begin{equation}\label{spline regressions}
\partial_{ x_k} \widehat{\beta}_j(\bx_i)
=\frac{\partial}{\partial x_k} \widehat{\beta}_j(\bx_i) ~\sim ~ x_{i,j}.
\end{equation}
This studies the derivative of regression attention $\widehat{\beta}_j(\bx)$ w.r.t.~$x_k$ as a function of the
corresponding feature component $x_j$ that is considered
in the feature contribution $ \beta_j(\bx) x_j$, see  \eqref{LocalGLMnet}.
The code for these spline fits is also provided in Listing \ref{GradientListing} on lines 12-15,
and it gives us the results illustrated in Figure \ref{interactions synthetic}.

Figure \ref{interactions synthetic} studies the spline regressions \eqref{spline regressions}
only of the significant components $x_1,\ldots, x_6$ that enter regression function $\mu$,
see \eqref{true synthetic}. We interpret these plots.
\begin{itemize}
\item
Plot $x_1$ shows that all gradients are roughly zero, which means that $\beta_1(\bx)={\rm const}$,
which, indeed, is the case in the true regression function $\mu$.
\item
Plots $x_2$ and $x_3$ show one term
that is significantly different from 0. For the $x_2$ plot it is 
$\partial_{x_2} \widehat{\beta}_2(\bx)$, and for the $x_3$ plot it is 
$\partial_{x_3} \widehat{\beta}_3(\bx)$. This says that these two terms do not have any interactions
with other variables,
and that the right functional form is not the linear one, but 
$\widehat{\beta}_j(\bx)x_j=\widehat{\beta}_j(x_j)x_j$ is non-linear for $j=2,3$.
In fact, we have $\partial_{x_2} \widehat{\beta}_2(\bx)\approx {\rm const}$ which says that
we have a quadratic term in $x_2$, and $\partial_{x_3} \widehat{\beta}_3(\bx)$ shows
a sine like behavior.
\item Plot $x_4$ shows a linear interaction with $x_5$ because 
$\partial_{x_5} \widehat{\beta}_4(\bx)\approx {\rm const}$.
\item Plot $x_5$ shows a linear interaction with $x_4$ because 
$\partial_{x_4} \widehat{\beta}_5(\bx)\approx {\rm const}$, and it shows
an interaction with $x_6$.
\item Plot $x_6$ does not show any significant terms, as they have already been captured
by the previous plots. Note that this comes from the fact that terms $x_5^2 x_6/8$ do not lead to 
an identifiable decomposition, but this can be allocated either to $x_5$ or to $x_6$, or could
even be split among the two.
\end{itemize}


\subsection{Real data example}
\label{Real data example}
As a second example we consider a real data example. In this real data example we also discuss
how categorical feature components should be pre-processed for our LocalGLMnet approach.
We consider the French motor third party liability (MTPL) claims frequency data
{\tt FreMTPL2freq} which is available through the {\sf R} package {\tt CASdatasets} of 
Dutang--Charpentier \cite{DutangCharpentier}. This data is described in Appendix A
of W\"uthrich--Merz \cite{WM2021} and in the tutorials of Noll et al.~\cite{Noll}
and Lorentzen--Mayer \cite{LorentzenMayer}. We apply the data cleaning of Listing B.1
of W\"uthrich--Merz \cite{WM2021} to this data. 

After data cleaning,  we have observations $(Y_i,v_i,\bx_i)_i$ with claim counts $Y_i \in \N_0$, time exposures $v_i \in (0,1]$ and feature
information $\bx_i$. We have 6 continuous feature components (called `Area Code', `Bonus-Malus Level', `Density',
`Driver's Age', `Vehicle Age', `Vehicle Power'), 1 binary component (called `Vehicle Gas') and 2 categorical components with more then
two levels (called `Vehicle Brand' and `Region'). We pre-process these components as follows: we center and normalize to unit 
variance the 6 continuous and the binary components. We apply one-hot encoding to the 2 categorical variables, we emphasize
that we do not use dummy coding as it is usually done in GLMs. Below, 
in Section \ref{Categorical feature components}, we are going to motivate this one-hot encoding
choice (which does not lead to full rank design matrices); for one-hot encoding vs.~dummy coding we refer
to formulas (5.21) and (7.29) in W\"uthrich--Merz \cite{WM2021}.

As a control variable we add two random feature components that are i.i.d.~distributed, centered and with unit variance, 
the first one having a uniform distribution and the second one having a standard normal distribution, 
we call these two additional feature components `RandU' and `RandN'. We consider two additional independent
components to understand whether the distributional choice influences the results of hypothesis testing
using the empirical interval $I_\alpha$, see \eqref{empirical confidence bounds}. Altogether (and using one-hot encoding)
we receive $q=42$ dimensional  tabular feature variables $\bx_i \in \R^q$; this includes the
two additional components RandU and RandN.

We fit a Poisson network regression model to this French MTPL data. Note that the Poisson distribution belongs to the EDF
with cumulant function $\kappa(\theta) = \exp\{\theta\}$ on the effective domain $\bTheta = \R$. The canonical link is given
by the log-link, this motivates link choice $g(\cdot)=\log (\cdot)$. We then start by fitting a plain-vanilla
FFN network \eqref{FN network regression} to this data. This FFN network will give us the benchmark in terms
of predictive power. We choose a network of depth $d=3$ having $(q_1,q_2,q_3)=(20,15,10)$ FFN neurons;
the {\sf R} code for this FFN architecture is given in Listing 7.1 of W\"uthrich--Merz \cite{WM2021},
but we replace input dimension 40 by 42 on line 3 of that listing.
In order to do a proper out-of-sample generalization analysis we partition the data randomly into a
learning data set ${\cal L}$ and a test data set ${\cal T}$. The learning data ${\cal L}$ contains $n=610,206$ 
instances and the test data set ${\cal T}$ contains $67,801$ instances; we use exactly the same split as in 
Table 5.2 of W\"uthrich--Merz \cite{WM2021}. The learning data ${\cal L}$ will be used to learn the network
parameters and the test data ${\cal T}$ is used to perform an out-of-sample generalization analysis. As loss
function for parameter fitting and generalization analysis we choose the Poisson deviance loss, which is a distribution adapted
and strictly consistent loss function for the mean within the Poisson model, for details we refer to 
Section 4.1.3 in W\"uthrich--Merz \cite{WM2021}.

We fit this FFN network using the {\tt nadam} version
of SGD on batches of size 5,000 over 100 epochs, and we retrieve the network calibration that provides
the smallest validation loss on a training-validation partition ${\cal U}$ and ${\cal V}$ of the learning data ${\cal L}$.
The results are presented on line (b) of Table \ref{loss results}. The FFN network provides clearly
better results than the null model only using a bias $\beta_0$. This justifies regression modeling, here.

\begin{table}[htb!]
\begin{center}
{\small
\begin{tabular}{|l|cc|}
\hline
&\multicolumn{2}{|c|}{Poisson deviance losses in $10^{-2}$}\\
& in-sample on ${\cal L}$ & out-of-sample on ${\cal T}$\\\hline
(a) null model (bias $\beta_0$ only) & 25.213 & 25.445\\
(b) FFN network & 23.764 & 23.873 \\
(c) LocalGLMnet & 23.728 &23.945 \\
(d) reduced LocalGLMnet & 23.714 &23.912 \\
\hline
\end{tabular}}
\end{center}
\caption{In-sample and out-of-sample losses on the real MTPL data example.}
\label{loss results}
\end{table}

Next we fit the LocalGLMnet architecture using exactly the same set up as in the FFN network,
but having a depth of $d=4$ with numbers of neurons $(q_0, q_1,q_2,q_3,q_4)=(42,20,15,10,42)$.
The results on line (c) of Table \ref{loss results} show that we sacrifice a bit of predictive power (out-of-sample)
for receiving our interpretable network architecture. We now analyze the resulting estimated
regression attentions $\widehat{\beta}_j(\bx)$.

\begin{figure}[htb!]
\begin{center}
\begin{minipage}[t]{0.32\textwidth}
\begin{center}
\includegraphics[width=.9\textwidth]{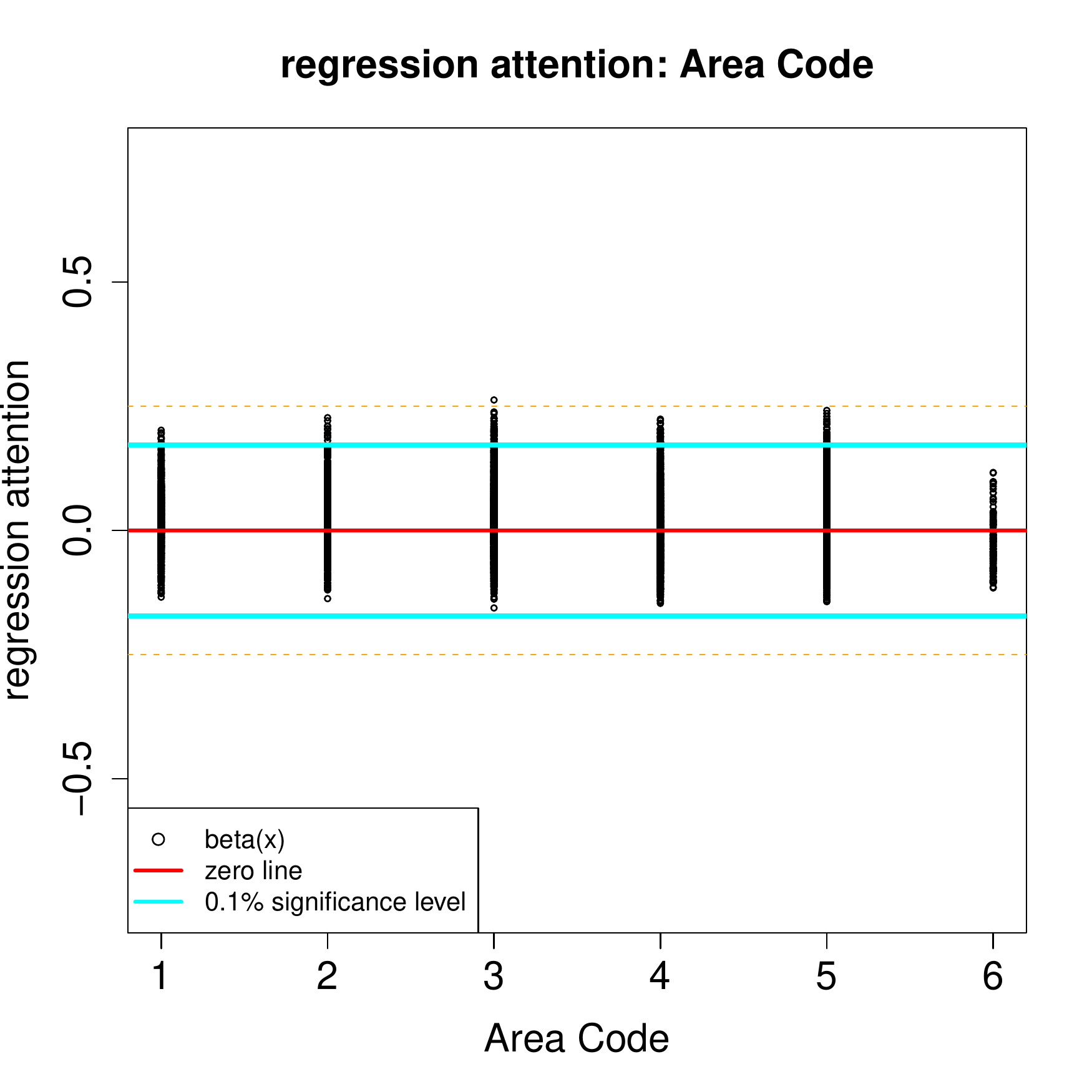}
\end{center}
\end{minipage}
\begin{minipage}[t]{0.32\textwidth}
\begin{center}
\includegraphics[width=.9\textwidth]{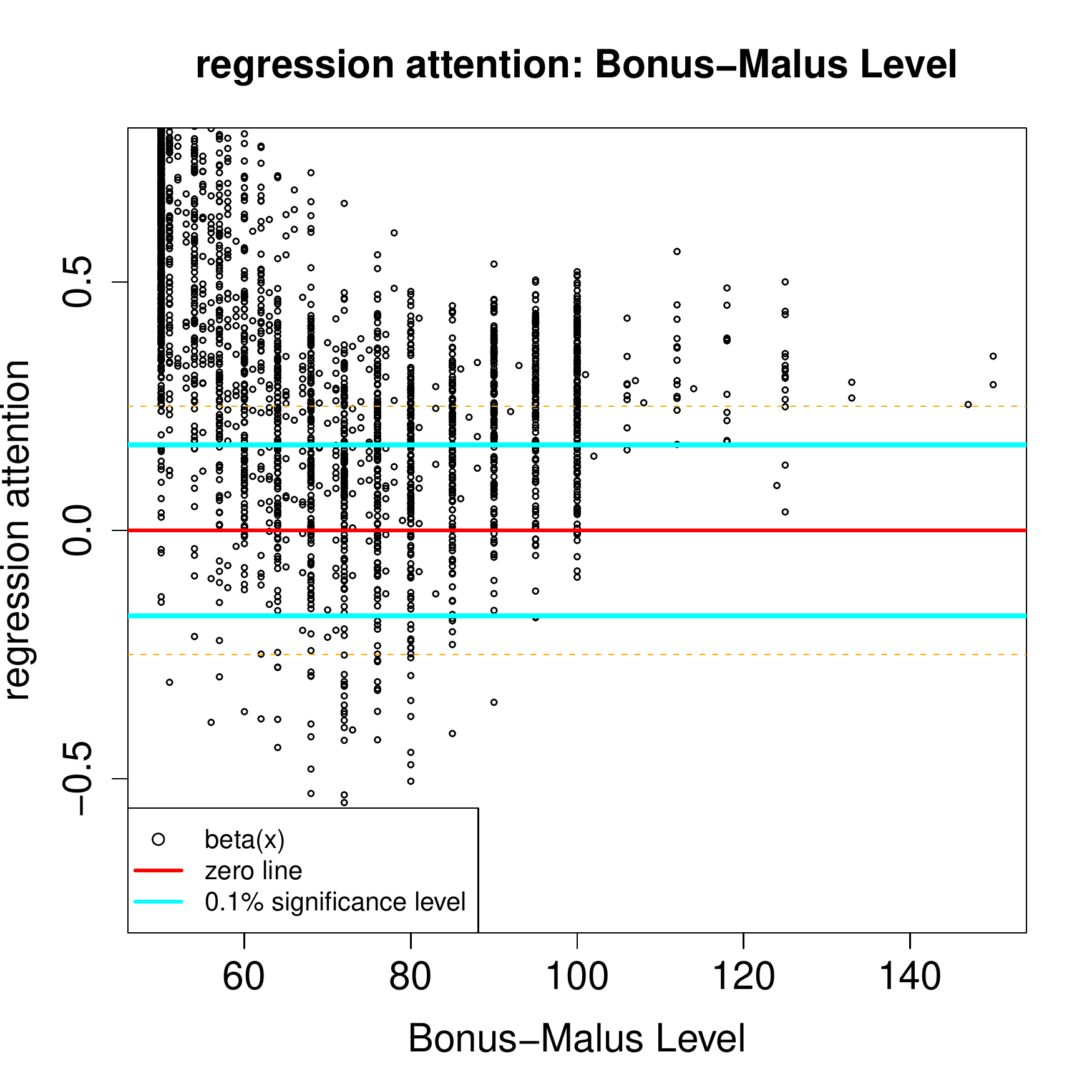}
\end{center}
\end{minipage}
\begin{minipage}[t]{0.32\textwidth}
\begin{center}
\includegraphics[width=.9\textwidth]{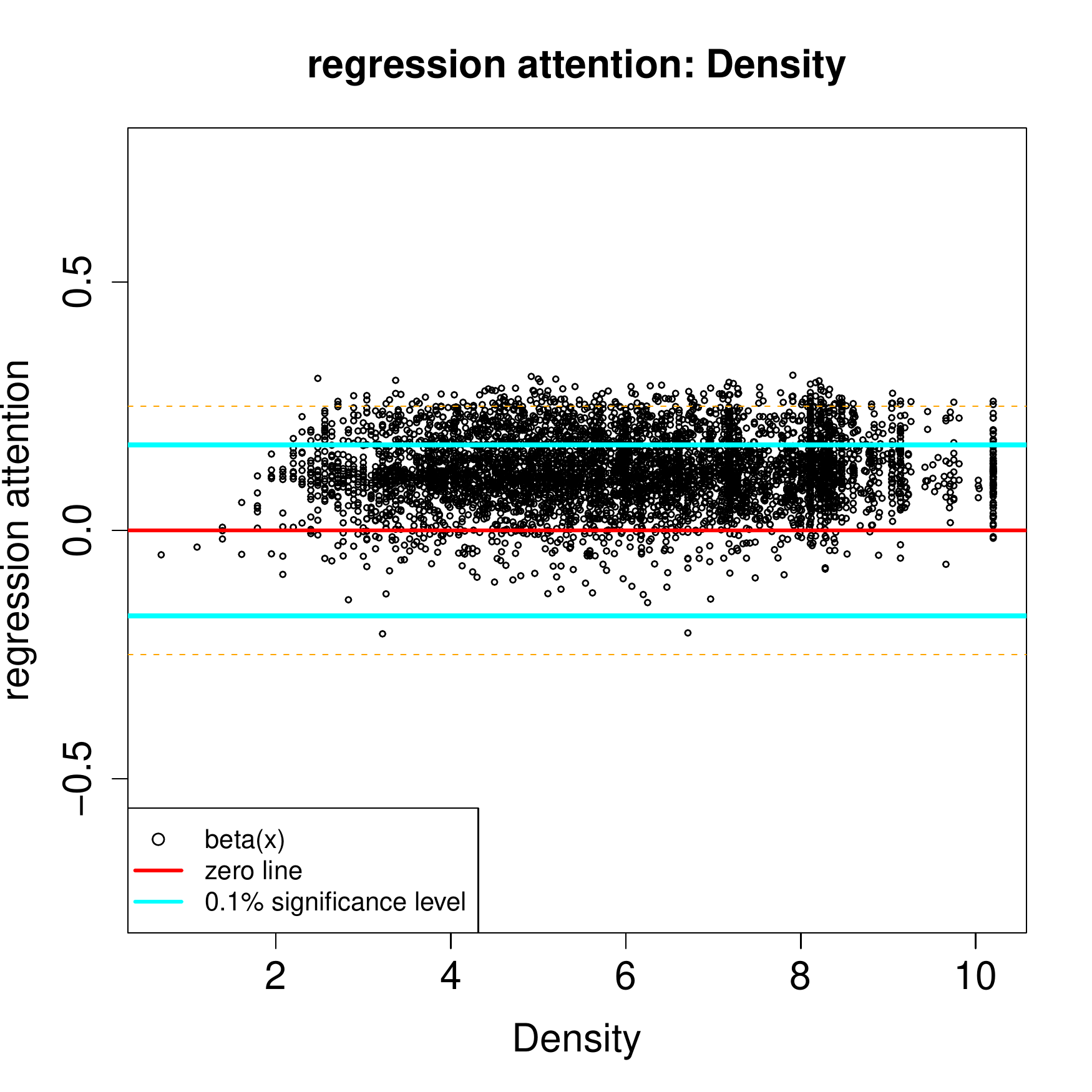}
\end{center}
\end{minipage}
\begin{minipage}[t]{0.32\textwidth}
\begin{center}
\includegraphics[width=.9\textwidth]{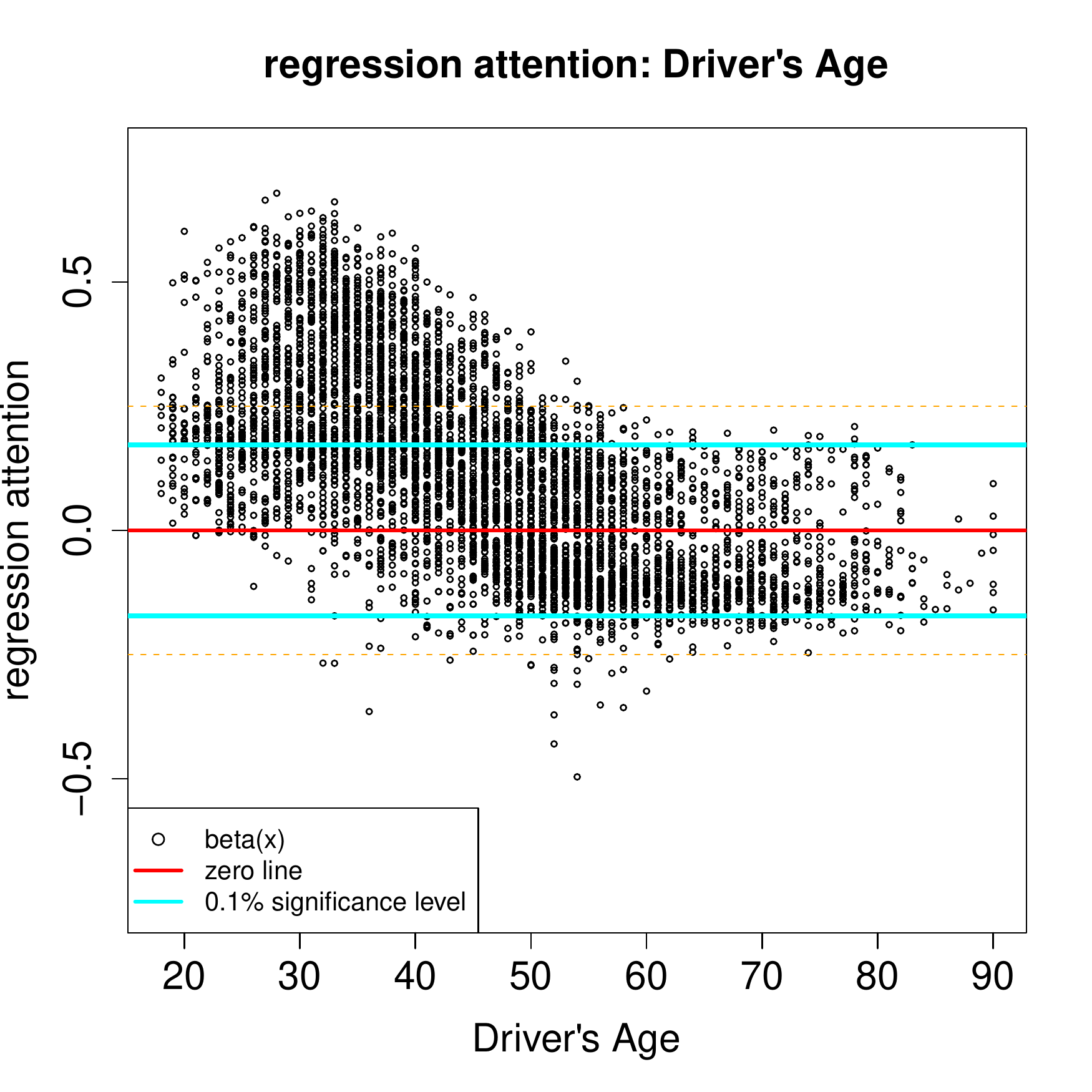}
\end{center}
\end{minipage}
\begin{minipage}[t]{0.32\textwidth}
\begin{center}
\includegraphics[width=.9\textwidth]{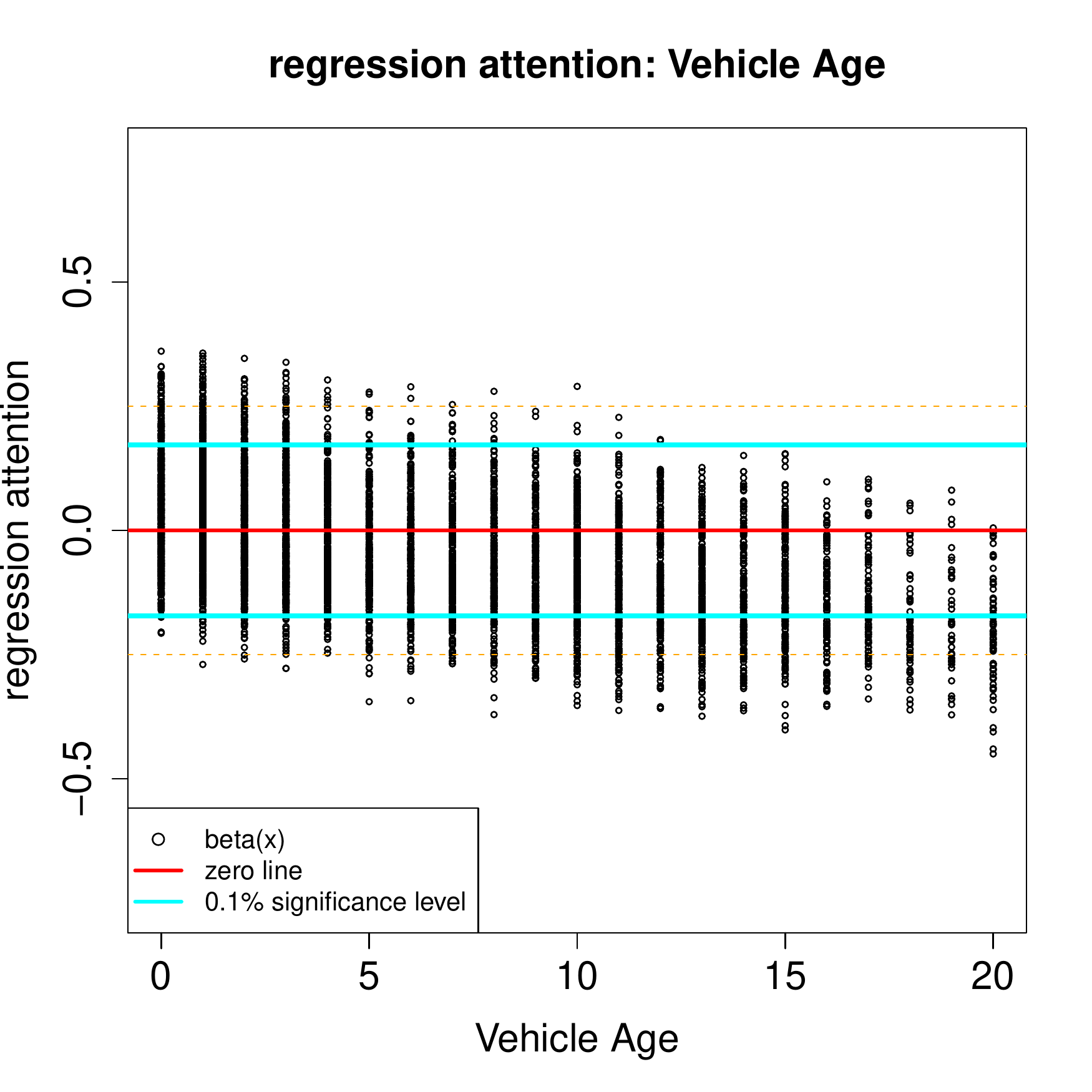}
\end{center}
\end{minipage}
\begin{minipage}[t]{0.32\textwidth}
\begin{center}
\includegraphics[width=.9\textwidth]{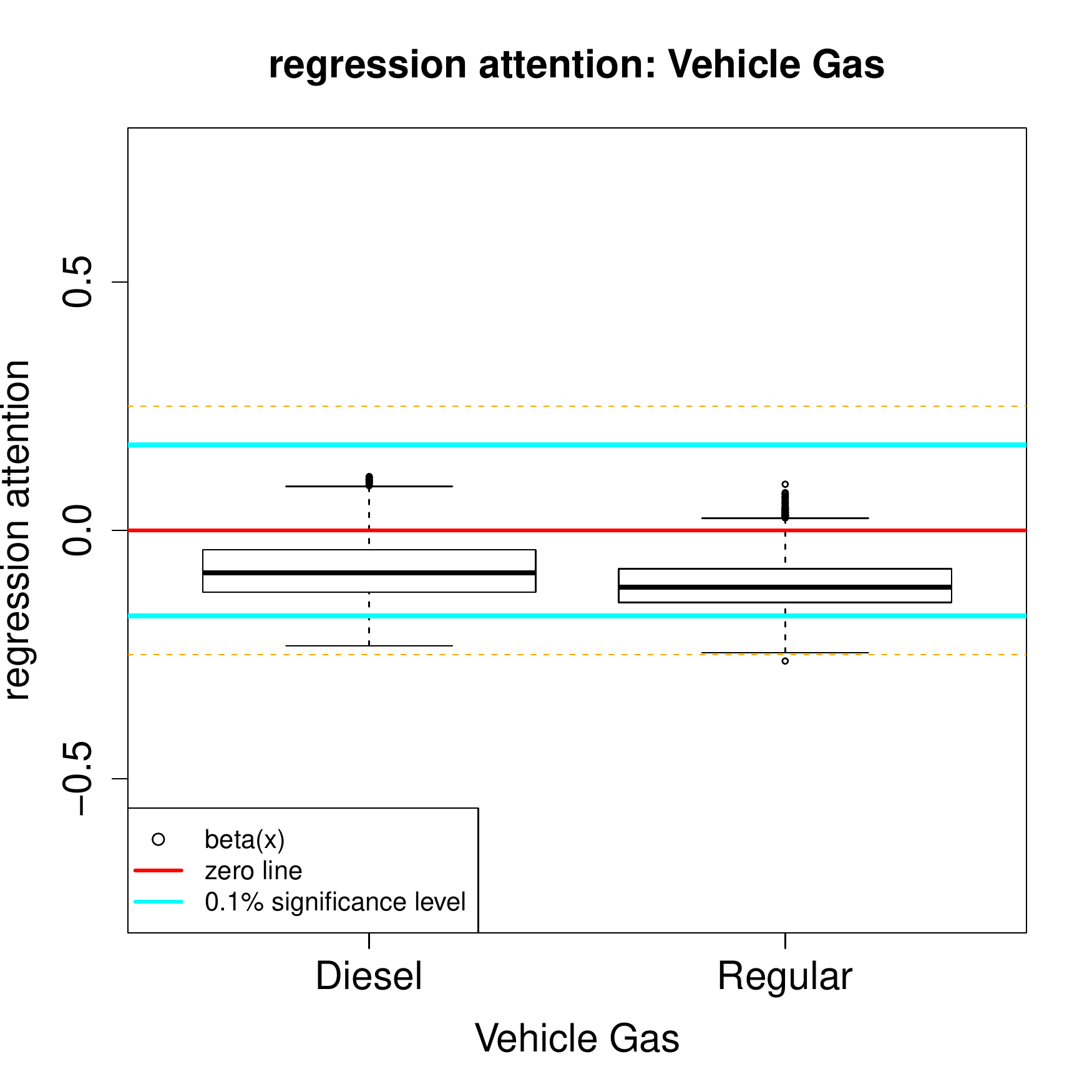}
\end{center}
\end{minipage}
\begin{minipage}[t]{0.32\textwidth}
\begin{center}
\includegraphics[width=.9\textwidth]{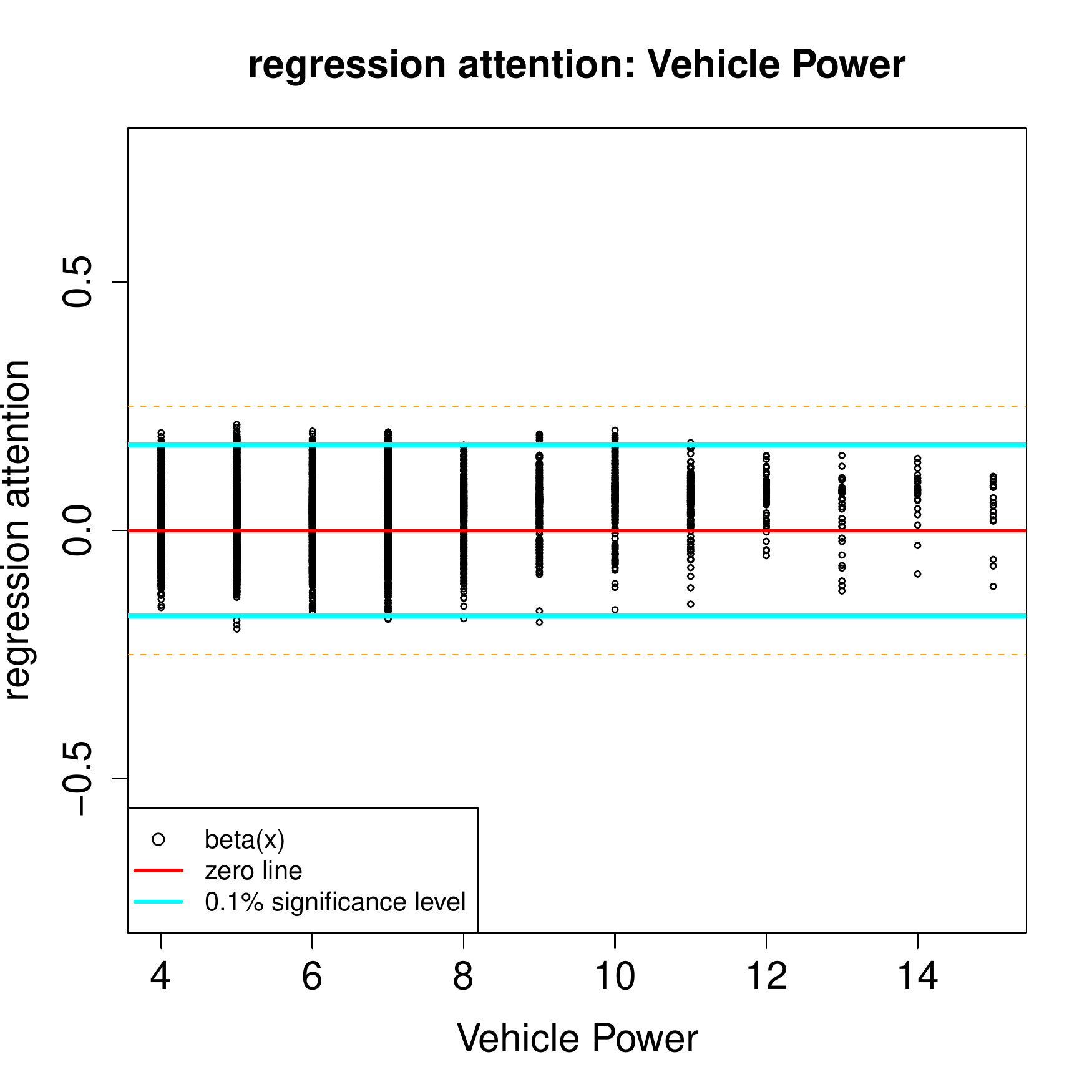}
\end{center}
\end{minipage}
\begin{minipage}[t]{0.32\textwidth}
\begin{center}
\includegraphics[width=.9\textwidth]{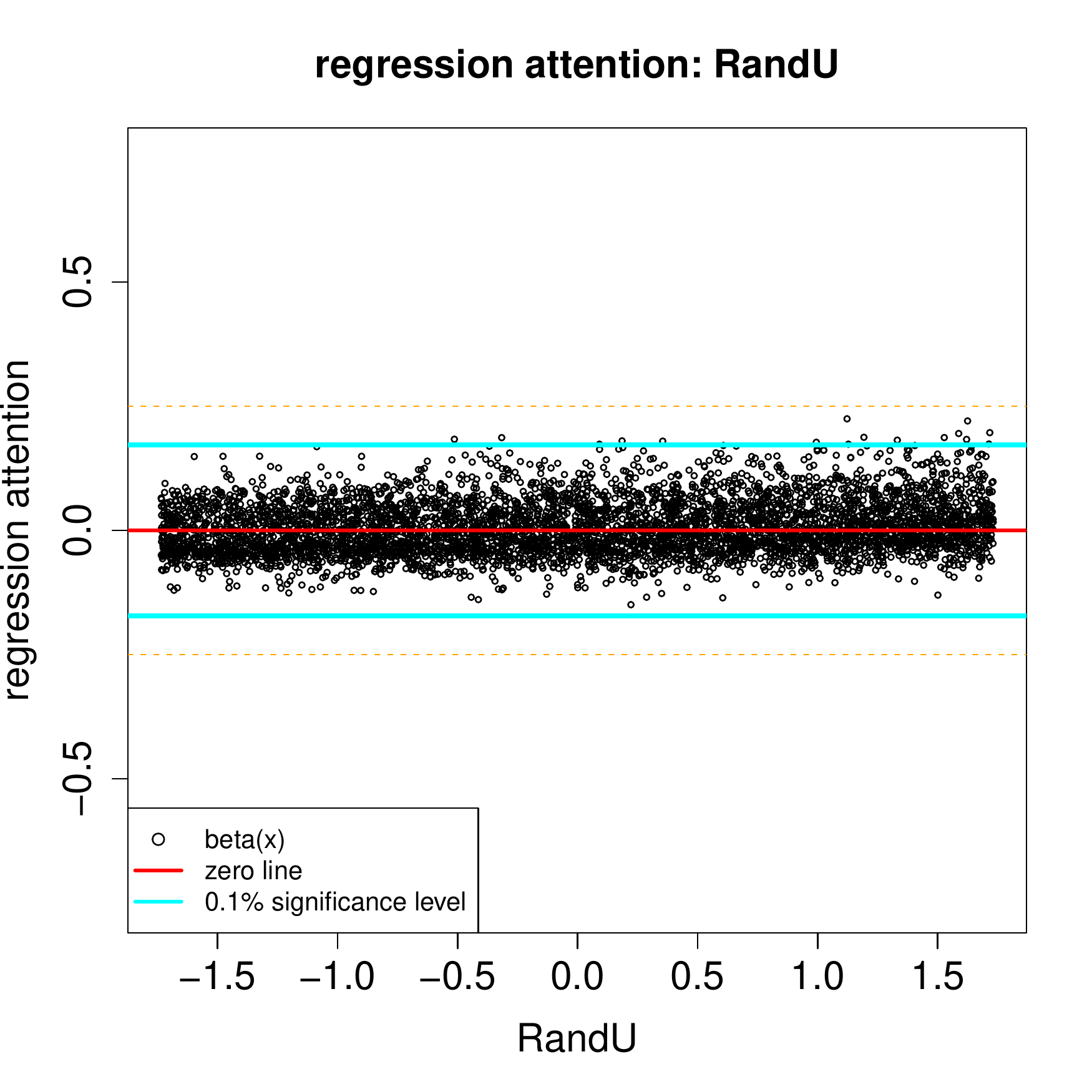}
\end{center}
\end{minipage}
\begin{minipage}[t]{0.32\textwidth}
\begin{center}
\includegraphics[width=.9\textwidth]{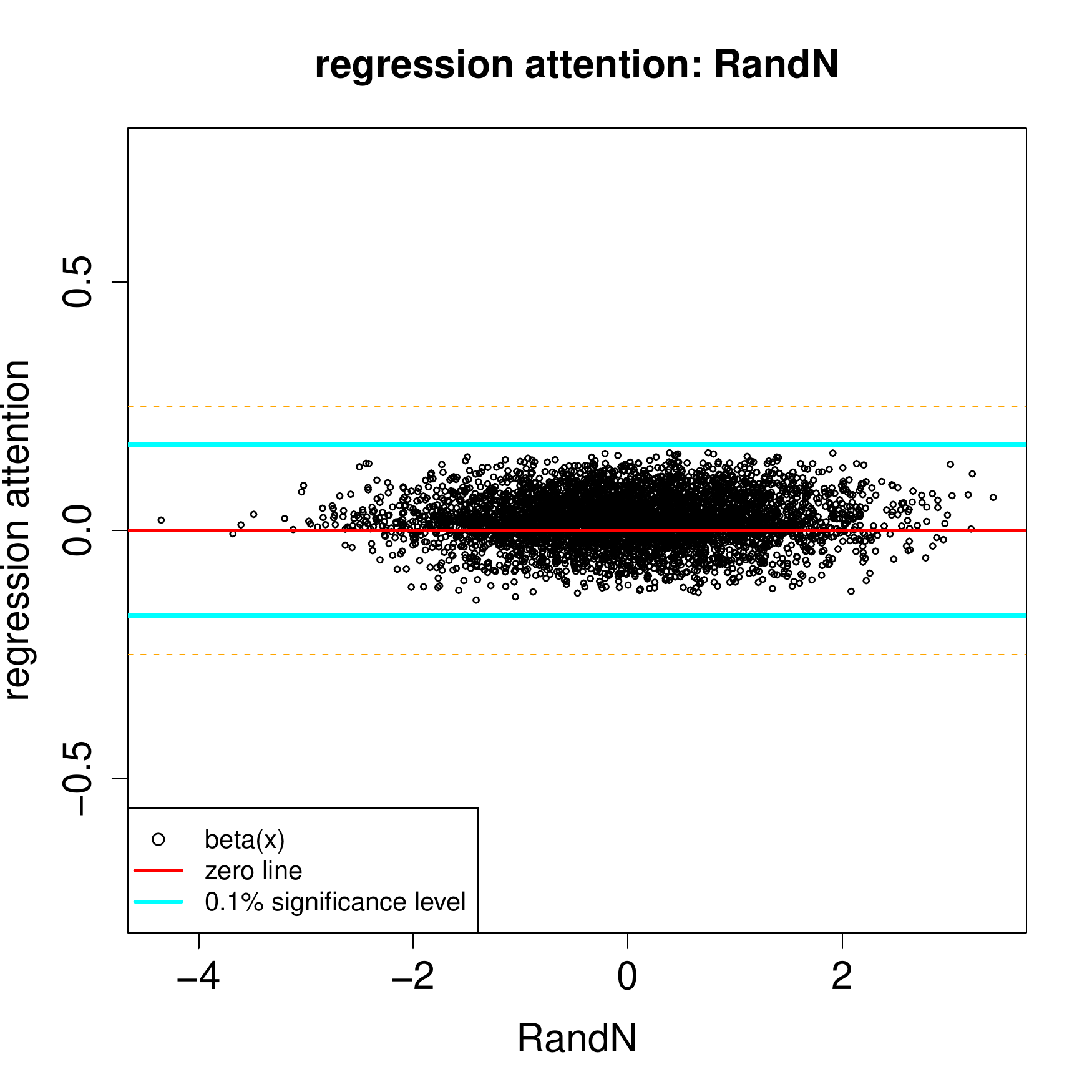}
\end{center}
\end{minipage}
\end{center}
\caption{Attention weights $\beta_j(\bx_t)$ of the continuous and binary feature components
Area Code, Bonus-Malus Level, Density, Driver's Age, Vehicle Age, Vehicle Gas, Vehicle Power, RandU
and RandN for 5,000 randomly selected instances $\bx_t$ of ${\cal T}$; the cyan lines show
the boundary of the rejection area $I^c_{\alpha}$ for dropping the term $x_j$ on significance level $\alpha =0.1\%$.}
\label{attentions 2}
\end{figure}

\begin{figure}[htb!]
\begin{center}
\begin{minipage}[t]{0.45\textwidth}
\begin{center}
\includegraphics[width=.9\textwidth]{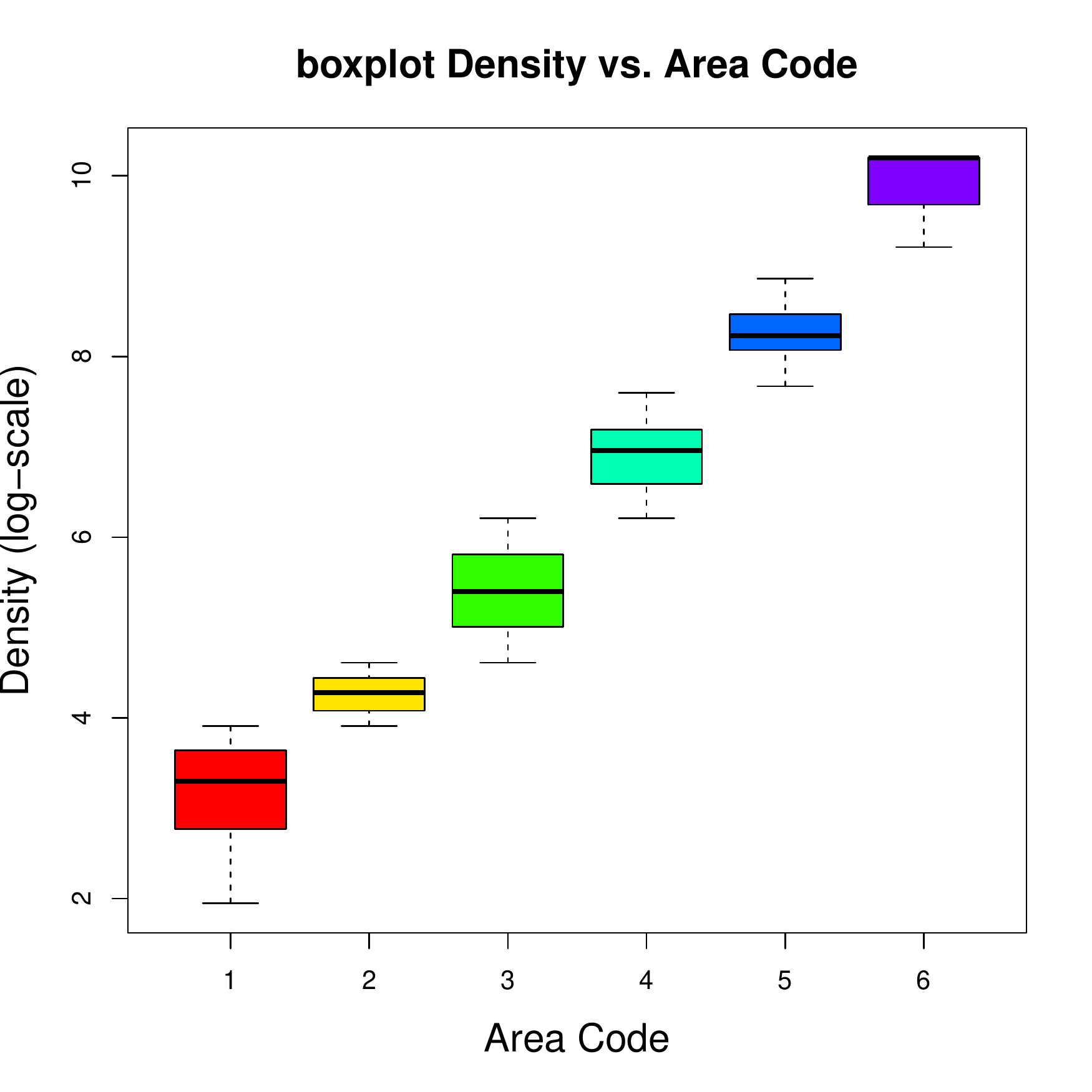}
\end{center}
\end{minipage}
\begin{minipage}[t]{0.45\textwidth}
\begin{center}
\includegraphics[width=.9\textwidth]{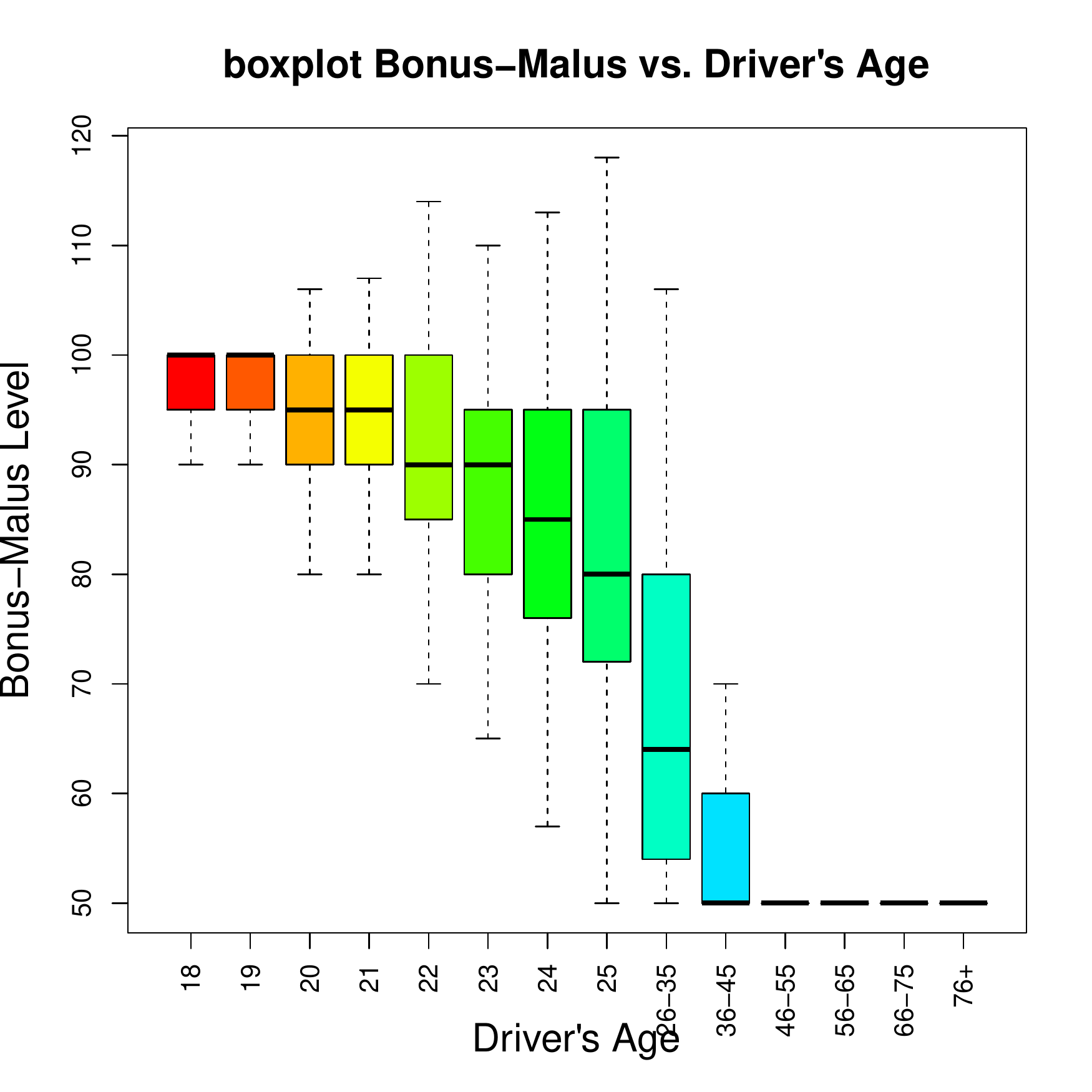}
\end{center}
\end{minipage}
\end{center}
\caption{(lhs) Boxplots of Density vs.~Area Code, and (rhs) Bonus-Malus vs.~Driver's Age.}
\label{box plots 1}
\end{figure}

We start by studying the regression attentions $\widehat{\beta}_j(\bx)$ of the continuous
and binary feature components Area Code, Bonus-Malus Level, Density, Driver's Age, Vehicle Age, Vehicle Gas, Vehicle Power,
RandU and RandN. First, we calculate the empirical standard deviations $\widehat{s}_j$ that we
receive from RandU and RandN, see \eqref{Wald test},
\begin{equation*}
\widehat{s}_{\rm RandU} = 0.052 \qquad \text{ and } \qquad
\widehat{s}_{\rm RandN} = 0.048.
\end{equation*}
Thus, these standard deviation estimates are rather similar, and in this case the specific distributional choice of the
control variable $x_{q+1}$ does not influence the results.
We calculate interval
$I_\alpha$ for significance level $\alpha =0.1\%$, see \eqref{empirical confidence bounds}.
The resulting confidence bounds are illustrated by the cyan lines in Figure \ref{attentions 2}.
We observe that for the variables Bonus-Malus Level, Density, Driver's Age, Vehicle Age and 
Vehicle Gas we clearly reject the null hypothesis $H_0: \beta_j(\bx)=0$ on the chosen significance
level $\alpha =0.1\%$, and Area Code and Vehicle Power need further analysis. For these two variables,
$I_\alpha$ provides a coverage ratio of 97.1\% and 98.1\%, thus, strictly speaking these numbers are
below $1-\alpha=99.9\%$ and we should keep these variables in the model. Nevertheless, we further
analyze these two variables. From the empirical analysis in Noll et al.~\cite{Noll} we know that
Area Code and Density are highly correlated. Figure \ref{box plots 1} (lhs) shows the boxplot of 
Density vs.~Area Code, and this plot highlights that Density almost fully explains Area Code. 
Therefore, it is sufficient to include the Density variable and we drop Area Code.

Thus, we drop the variables Area Code and VehPower, and we also drop the control variables
RandU and RandN, because these are no longer needed. This gives us reduced input dimension $q_0=q=38$,  and we run the same 
LocalGLMnet SGD fitting again.
Line (d) of Table \ref{loss results} gives the in-sample and out-of-sample results of this reduced 
LocalGLMnet model. We observe
a small out-of-sample improvement compared to line (c) which confirms that we can drop Area Code and VehPower
without losing predictive performance. In fact, the small improvement indicates that a smaller
model less likely over-fits, and we can consider more SGD steps before over-fitting, i.e., we receive
a later early stopping point.

\begin{figure}[htb!]
\begin{center}
\begin{minipage}[t]{0.32\textwidth}
\begin{center}
\includegraphics[width=.9\textwidth]{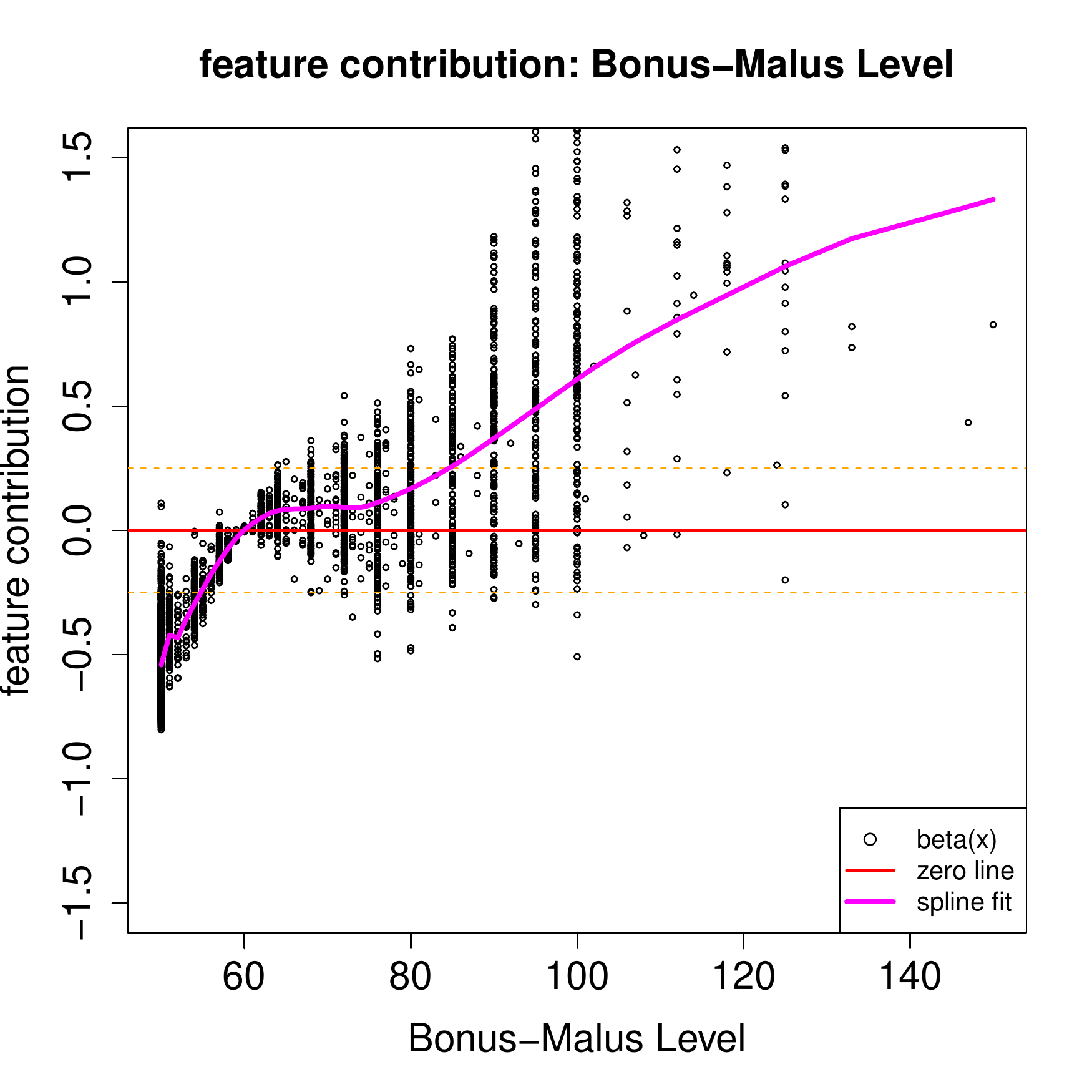}
\end{center}
\end{minipage}
\begin{minipage}[t]{0.32\textwidth}
\begin{center}
\includegraphics[width=.9\textwidth]{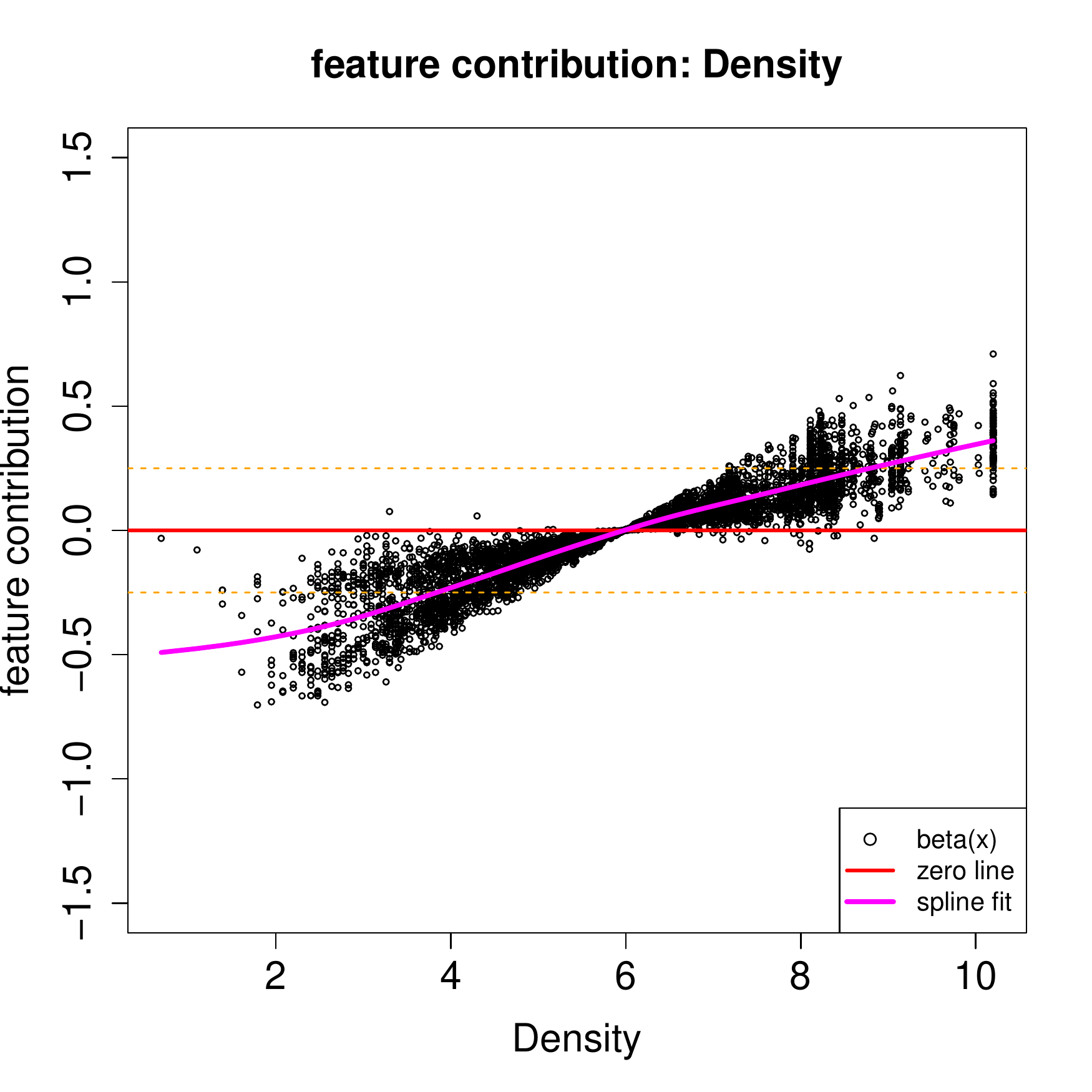}
\end{center}
\end{minipage}
\begin{minipage}[t]{0.32\textwidth}
\begin{center}
\includegraphics[width=.9\textwidth]{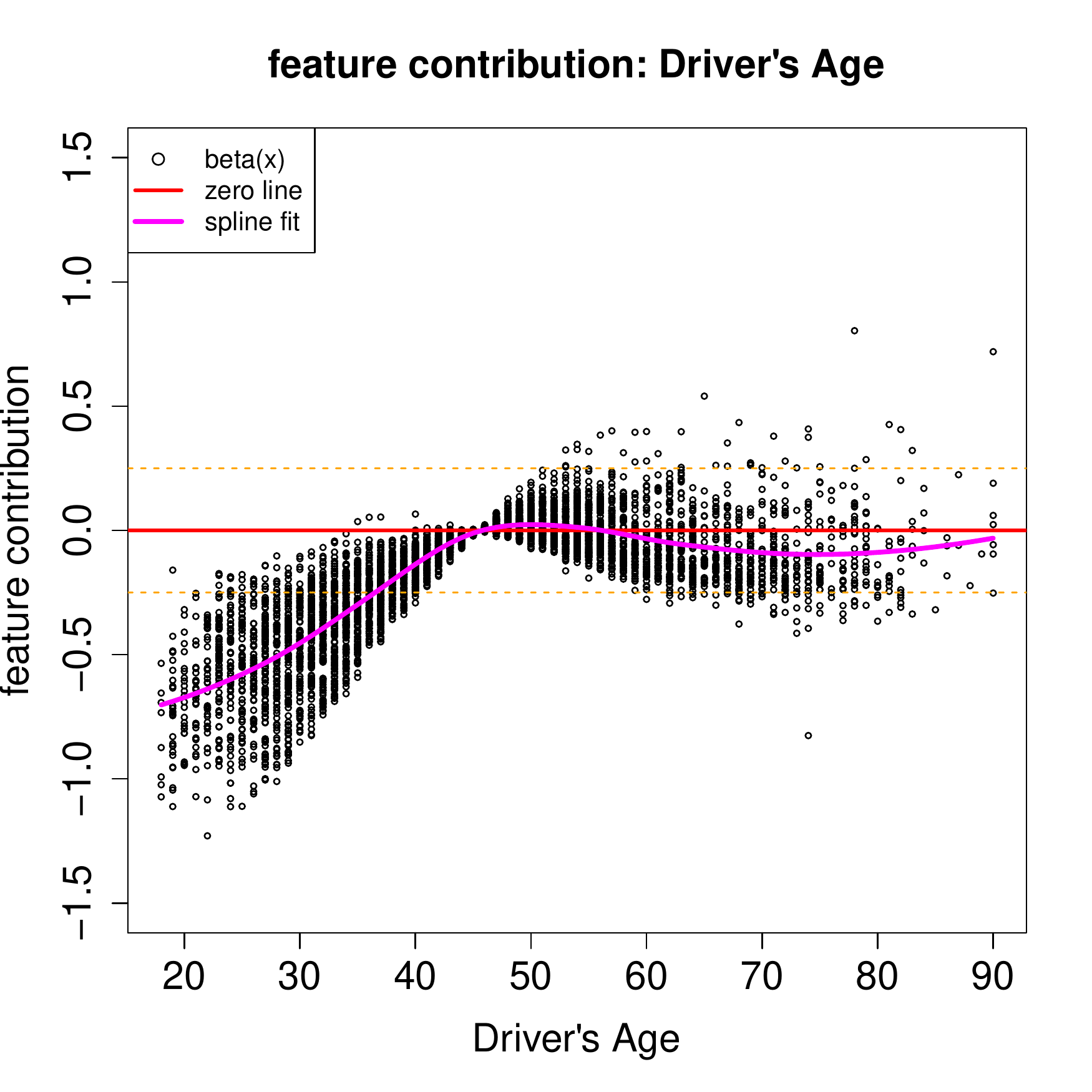}
\end{center}
\end{minipage}
\begin{minipage}[t]{0.32\textwidth}
\begin{center}
\includegraphics[width=.9\textwidth]{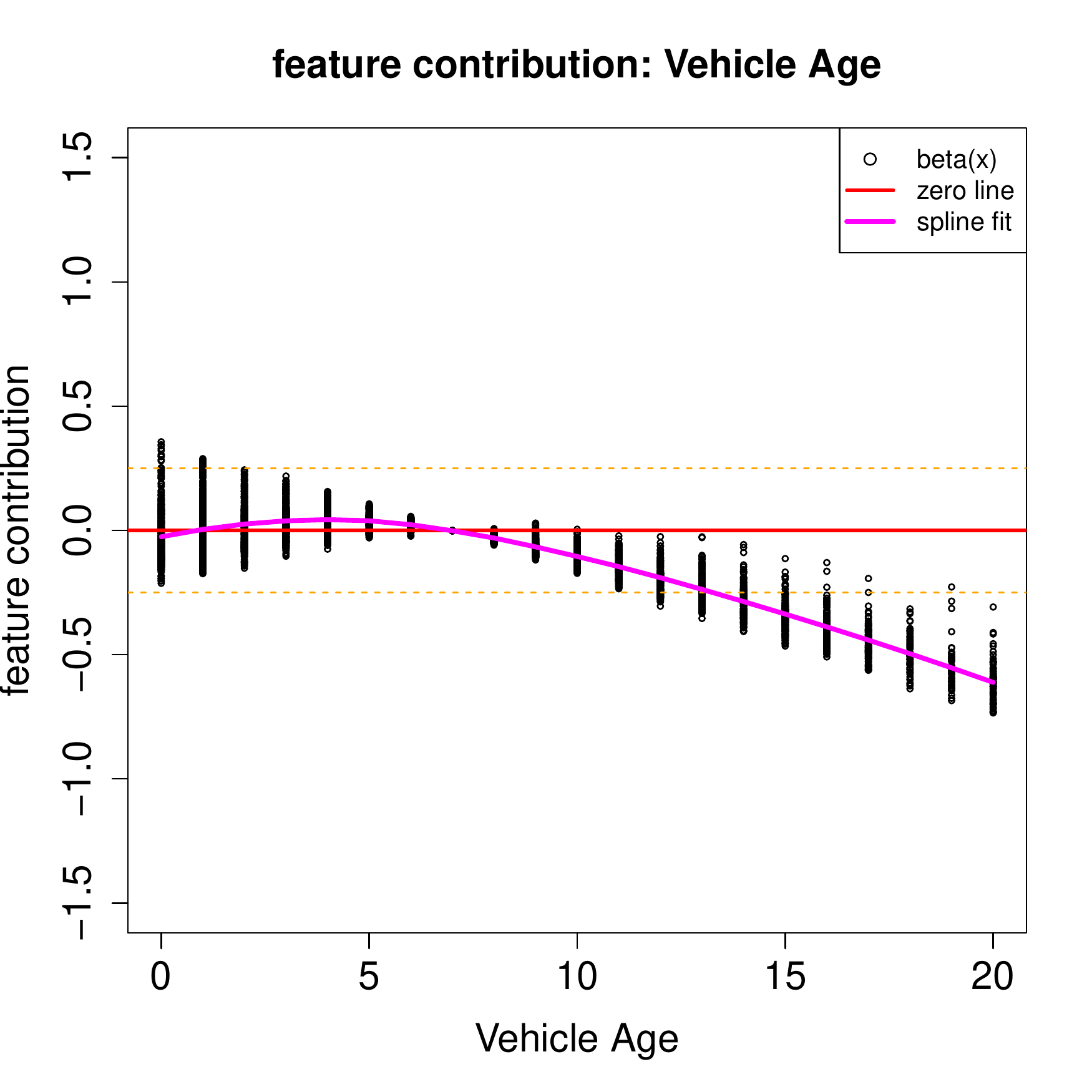}
\end{center}
\end{minipage}
\begin{minipage}[t]{0.32\textwidth}
\begin{center}
\includegraphics[width=.9\textwidth]{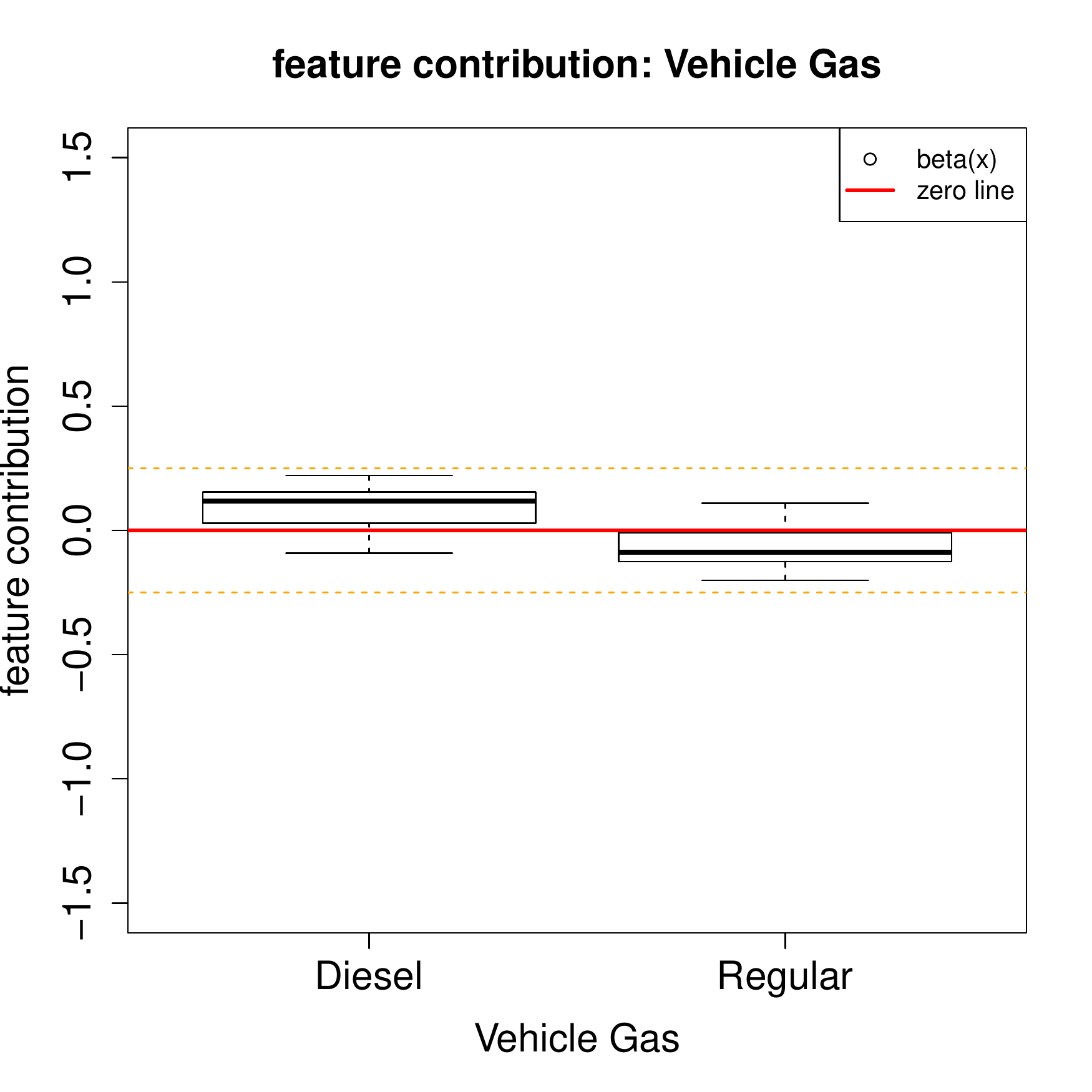}
\end{center}
\end{minipage}
\end{center}
\caption{Feature contributions $\widehat{\beta}_j(\bx_t)x_{j,t}$ of the continuous and binary feature components
Bonus-Malus Level, Density, Driver's Age, Vehicle Age and Vehicle Gas
of 5,000 randomly selected instances $\bx_t$ of ${\cal T}$; the $y$-scale is the same in all plots and the
magenta color gives a spline fit to the feature contributions.}
\label{attentions 1}
\end{figure}

Figure \ref{attentions 1} shows the feature contributions $\widehat{\beta}_j(\bx_t)x_{j,t}$ of the selected
continuous and binary feature components Bonus-Malus Level, Density, Driver's Age, Vehicle Age and Vehicle Gas
of 5,000 randomly selected instances $\bx_t$, and the magenta line shows a spline fit to these feature contributions;
the $y$-scale is the same in all plots. We observe that all these components contribute substantially to the regression
function, the Bonus-Malus variable being the most important one, and Vehicle Gas being the least important
one. Bonus-Malus and Density have in average an increasing trend, and Vehicle Age 
has in average a decreasing trend,
Density being close to a linear function, and the remaining continuous variables
are clearly non-linear. 
The explanation of Driver's Age is more difficult as can be seen from the spline fit (magenta color).
Figure \ref{box plots 1} (rhs) shows the boxplot of Bonus-Malus Level vs.~Driver's Age. We observe that
new (young) drivers enter the bonus-malus system at 100, and every year of driving without an accident
decreases the bonus-malus level. Therefore, the lowest bonus-malus level can only be reached after 
multiple years of accident-free driving. This can be seen from Figure \ref{box plots 1} (rhs), and
it implies that the Bonus-Malus Level and the Driver's Age variables interact. We are going to
verify this by studying the gradients $\nabla \widehat{\beta}_j(\bx)$ of the regression attributions.

\begin{figure}[htb!]
\begin{center}
\begin{minipage}[t]{0.4\textwidth}
\begin{center}
\includegraphics[width=.9\textwidth]{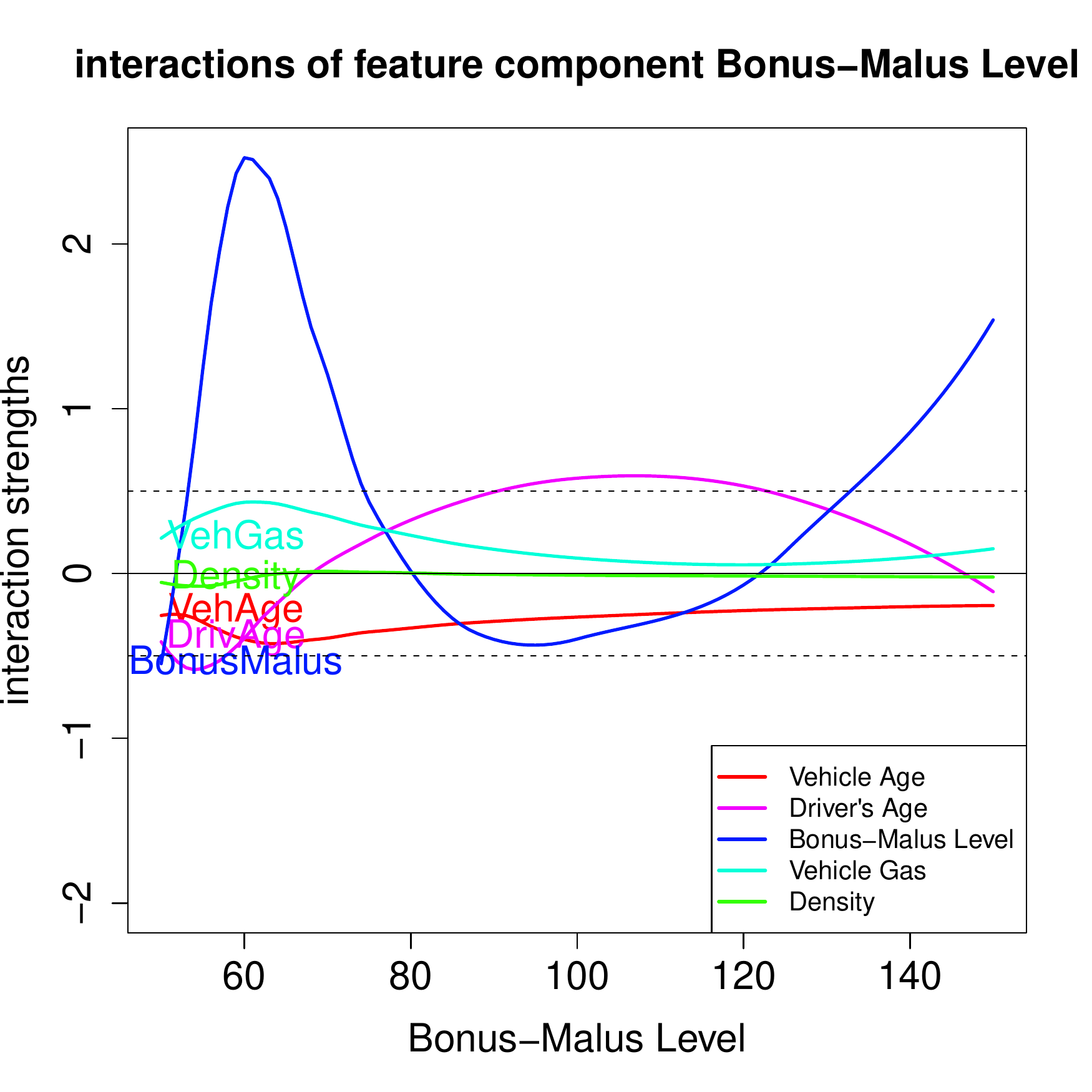}
\end{center}
\end{minipage}
\begin{minipage}[t]{0.4\textwidth}
\begin{center}
\includegraphics[width=.9\textwidth]{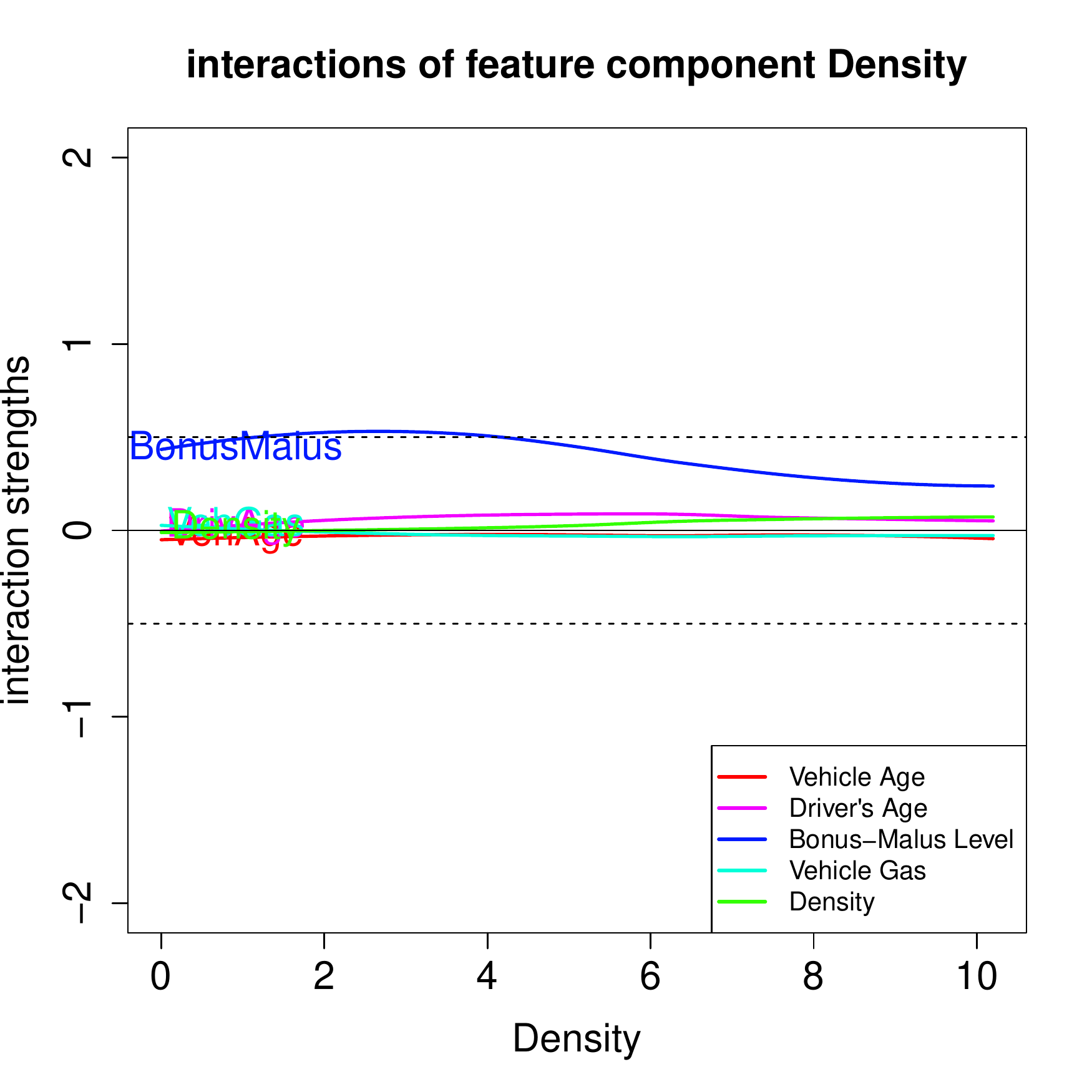}
\end{center}
\end{minipage}
\begin{minipage}[t]{0.4\textwidth}
\begin{center}
\includegraphics[width=.9\textwidth]{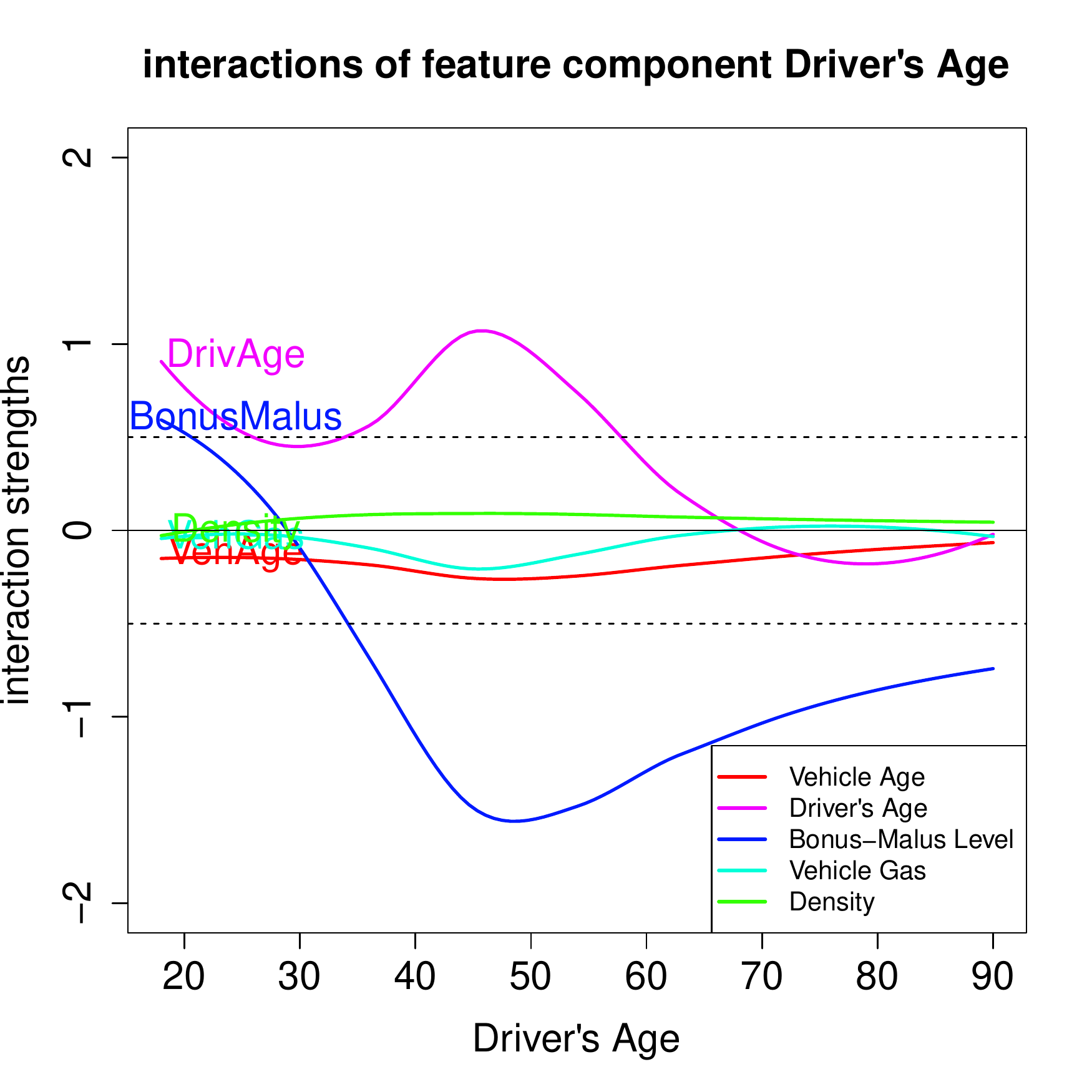}
\end{center}
\end{minipage}
\begin{minipage}[t]{0.4\textwidth}
\begin{center}
\includegraphics[width=.9\textwidth]{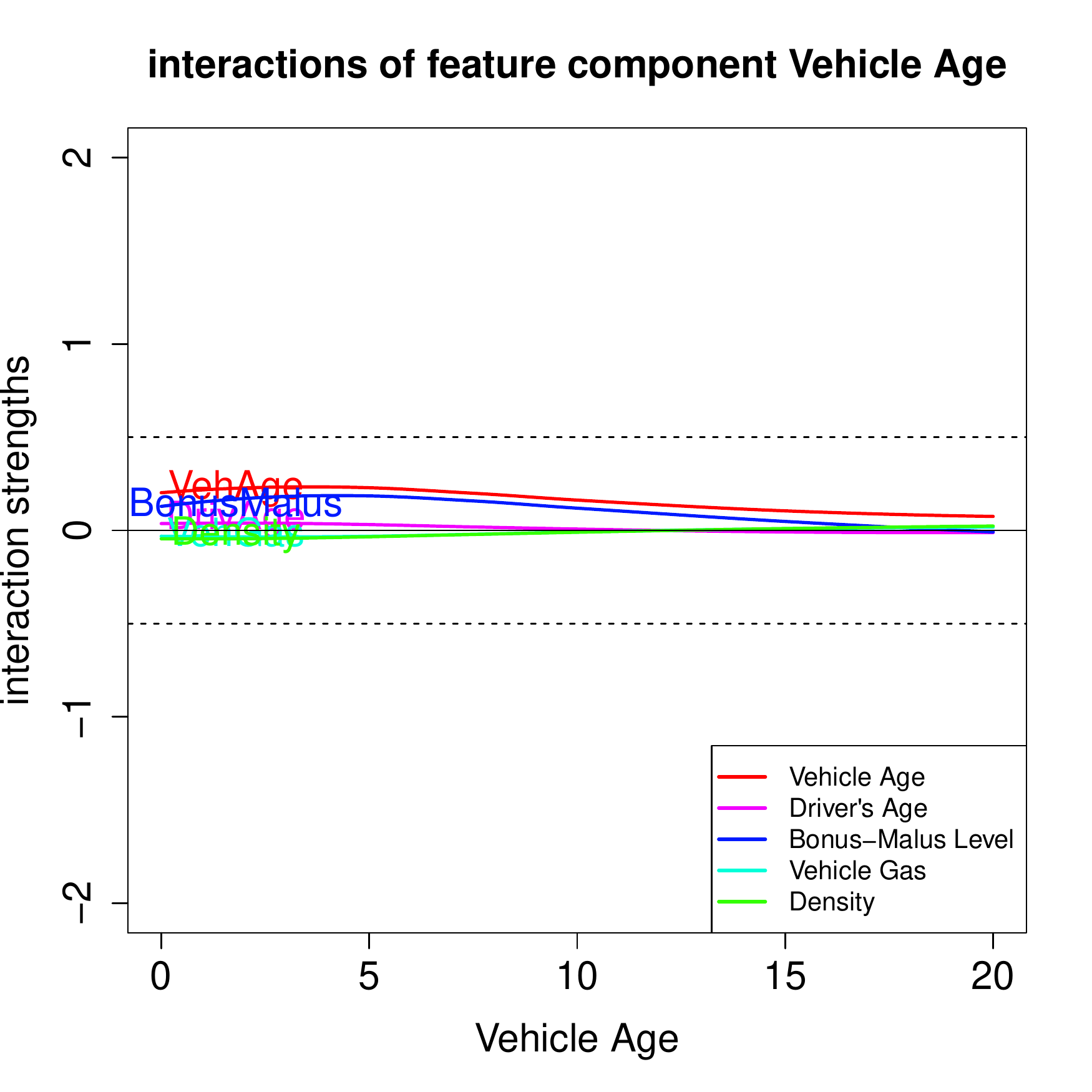}
\end{center}
\end{minipage}
\end{center}
\caption{Spline fits to the gradients $\partial_{x_k} \widehat{\beta}_j(\bx_i)$
of the continuous variables Bonus-Malus Level, Density, Driver's Age and Vehicle Age
over all instances $i=1,\ldots, n$.}
\label{interaction splines}
\end{figure}

Figure \ref{interaction splines} shows the spline fits to the gradients $\partial_{x_k} \widehat{\beta}_j(\bx)$
of the continuous variables Bonus-Malus Level, Density, Driver's Age and Vehicle Age.
Firstly, we observe that Bonus-Malus Level and Driver's Age have substantial non-linear
terms $\partial_{x_j} \widehat{\beta}_j(\bx) x_j$, Vehicle Age shows some non-linearity and
Density seems to be pretty linear since $\partial_{x_j} \widehat{\beta}_j(\bx) x_j \approx 0$.
This verifies the findings of Figure \ref{attentions 1} of the magenta spline fits.

Next we focus on interactions which requires the study of $\partial_{x_k} \widehat{\beta}_j(\bx)$
for $k\neq j$ in Figure \ref{interaction splines}. The most significant interactions can clearly
be observed between Bonus-Malus Level and Driver's Age, but also between the Bonus-Malus
Level and Density we encounter an interaction term saying that a higher Bonus-Malus Level
at a lower Density leads to a higher prediction, which intuitively makes sense as in less densely
populated areas we expect less claims. For Vehicle Age we do not find substantial interactions, it
only weakly interacts with Bonus-Malus Level and Driver's Age by entering the corresponding
regression attributions $\widehat{\beta}_j(\bx)$ of these two feature components.

There remains the discussion of the categorical feature components Vehicle Brand and Region.
This is going to be done in Section \ref{Categorical feature components}, below.

\subsection{Variable importance}
\label{Variable importance}
\begin{figure}[htb!]
\begin{center}
\begin{minipage}[t]{0.45\textwidth}
\begin{center}
\includegraphics[width=.9\textwidth]{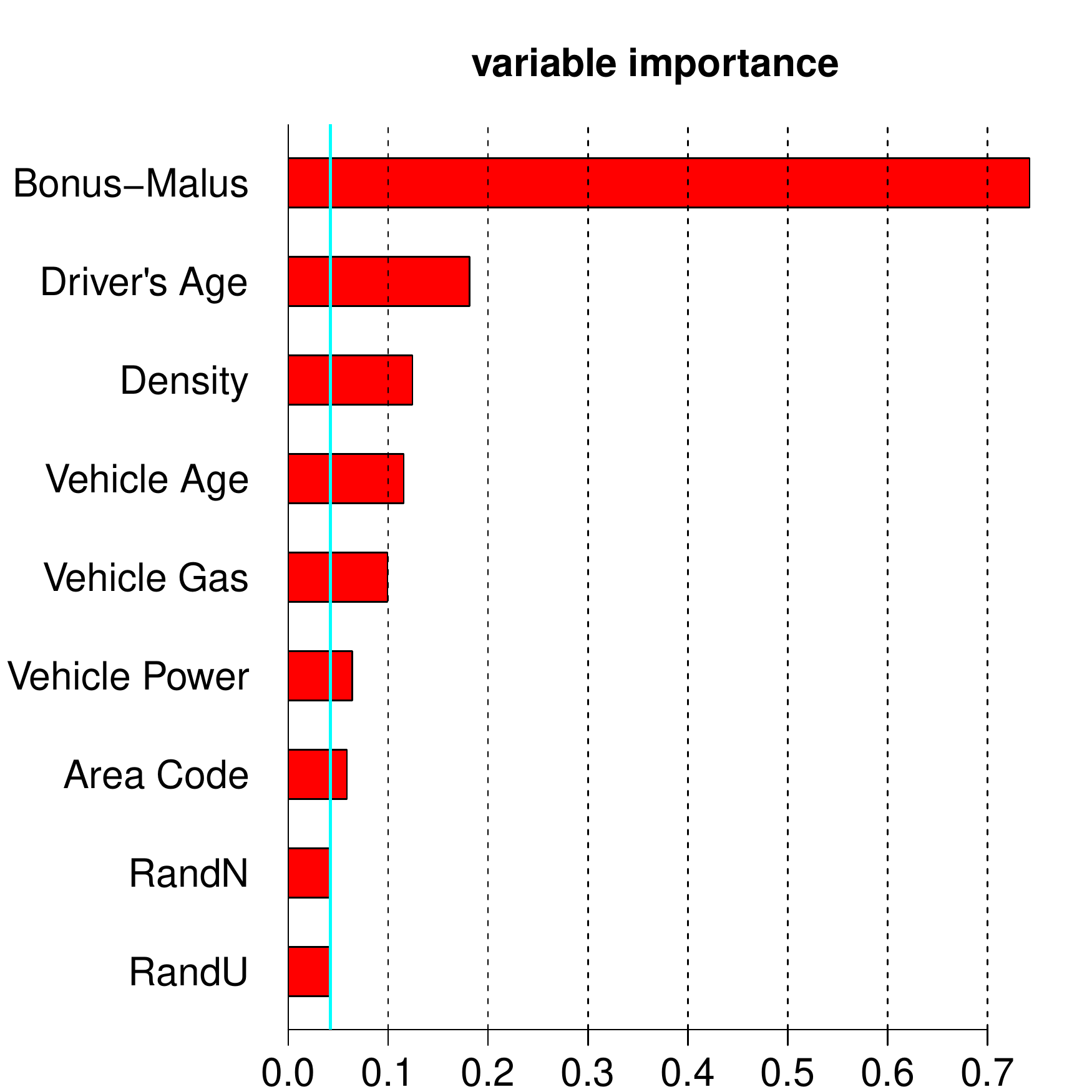}
\end{center}
\end{minipage}
\end{center}
\caption{Variable importance ${\rm VI}_j$, $1\le j \le q$.}
\label{variable importance plot}
\end{figure}

The estimated regression attentions $\widehat{\beta}_j(\bx)$, $1\le j \le q$, allow us to 
quantify variable importance. Coming back to the SHAP additive decomposition \eqref{SHAP}, 
a popular way of quantifying variable importance is obtained by aggregating the absolute values
of the attention weights. A simple measure of variable importance can thus be defined
by 
\begin{equation*}
{\rm VI}_j = \frac{1}{n} \sum_{i=1}^n \left|\widehat{\beta}_j(\bx_i)\right|,
\end{equation*}
for $1\le j \le q$ and where we aggregate over all instances $1\le i \le n$. Typically, the bigger these
values ${\rm VI}_j$ the more component $x_j$ influences the regression function. Note that all feature
components $x_j$ have been centered and normalized to unit variance, i.e., they live on the same
scale, otherwise such a comparison of ${\rm VI}_j$ across different $j$ would not make sense.
Figure \ref{variable importance plot} gives the variable importance results, which emphasizes that Area Code and
Vehicle Power are the least important variables which, in fact, have been dropped in a second step above.

\subsection{Categorical feature components}
\label{Categorical feature components}
In this section we discuss how categorical feature components can be considered.
Note that at the beginning of Section \ref{Real data example} we have emphasized that we
use one-hot encoding and not dummy coding for categorical variables. The reason for this
choice can be seen in Figure \ref{attentions 1}. Namely, if the feature value is $x_j=0$, then the corresponding
feature contribution gives ${\beta}_j(\bx)x_{j}=0$. This provides the calibration of the regression
model, i.e., it gives the reference level which is determined by the bias $\beta_0 \in \R$.
Since in GLMs we do not allow the components to interact in the linear predictor $\eta(\bx)$, all instances
that have the same level $x_j$ receive the same contribution $\beta_j x_j$, see \eqref{GLM}.
In the LocalGLMnet we allow the same level $x_j$ to have different contributions
${\beta}_j(\bx)x_{j}$ through interactions in the regression attention $\bbeta(\bx)$.
If we want to carry this forward to categorical feature components it requires that these components
receive an encoding that is not identically equal to zero. This is the case for one-hot encoding, but
not for dummy coding where the reference level is just identical to the bias $\beta_0$. For this
reason we recommend to use one-hot encoding for the LocalGLMnet.

\begin{figure}[htb!]
\begin{center}
\begin{minipage}[t]{0.45\textwidth}
\begin{center}
\includegraphics[width=.9\textwidth]{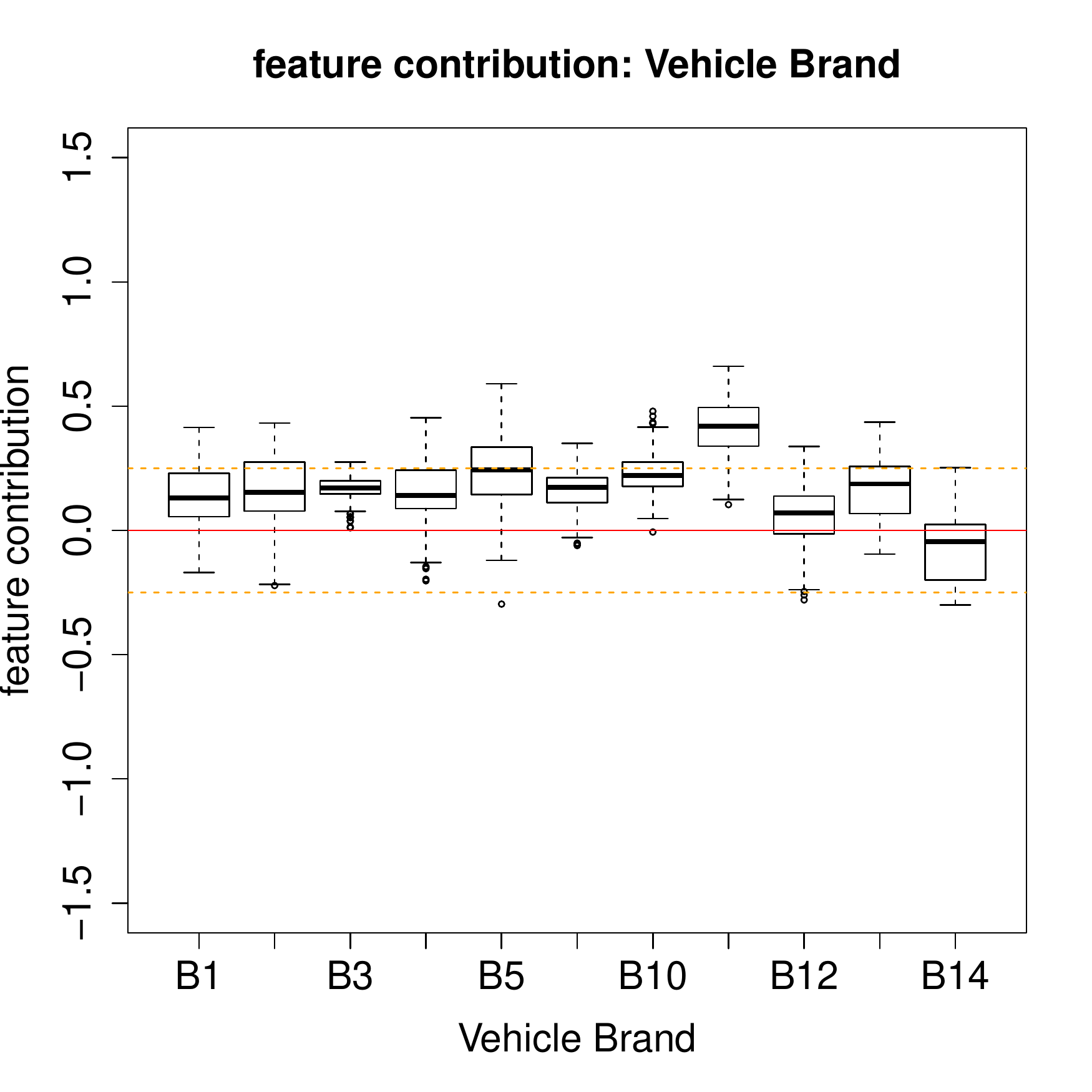}
\end{center}
\end{minipage}
\begin{minipage}[t]{0.45\textwidth}
\begin{center}
\includegraphics[width=.9\textwidth]{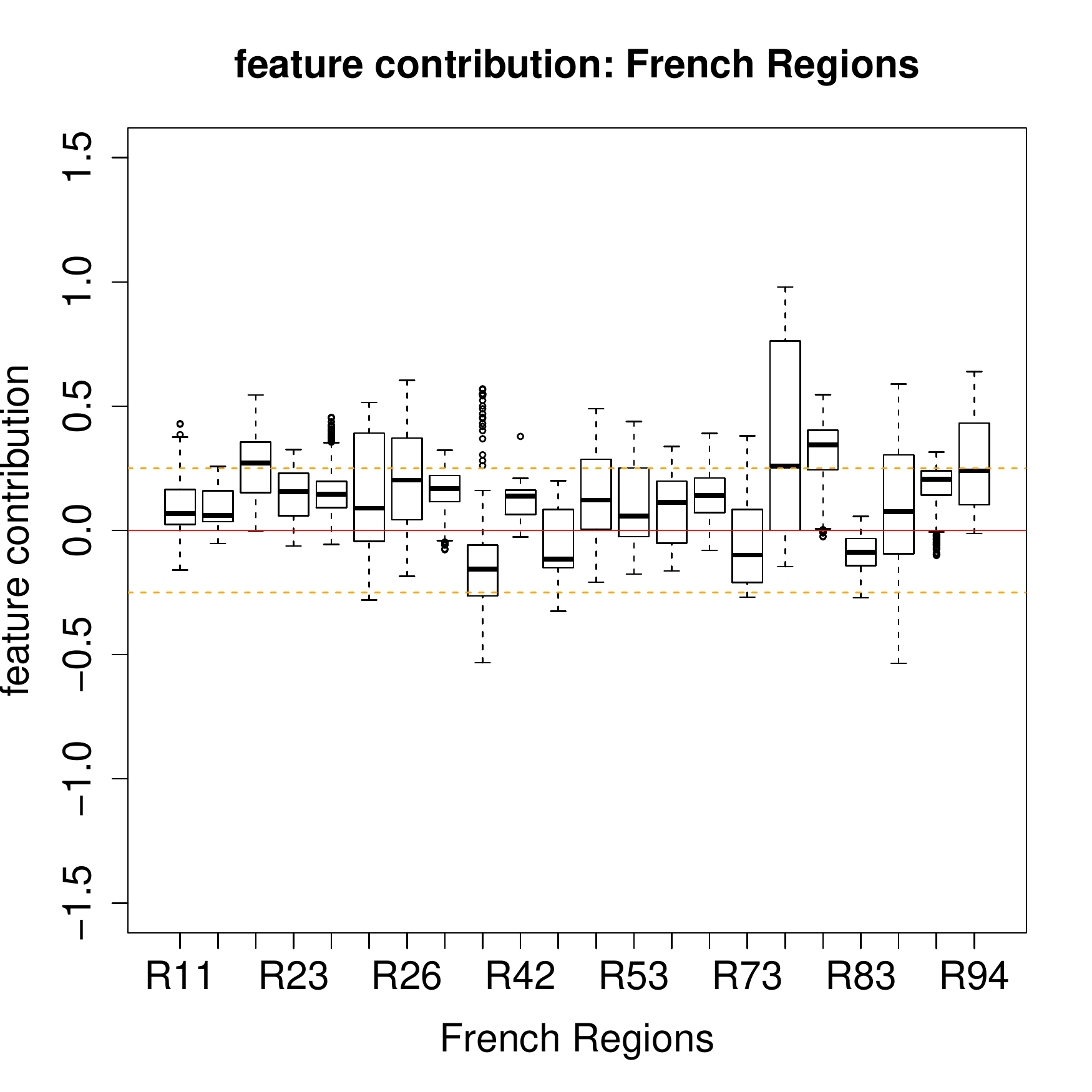}
\end{center}
\end{minipage}
\end{center}
\caption{Boxplot of the feature contributions $\widehat{\beta}_j(\bx)$ of the categorical feature components
Vehicle Brand and French Regions; the $y$-scale is the same as
in Figure \ref{attentions 1}.}
\label{attentions 3}
\end{figure}

Figure \ref{attentions 3} shows the feature contributions $\widehat{\beta}_j(\bx)$ of the categorical feature components,
note that each box corresponds to one level of the chosen categorical variable. Firstly, different medians between
the boxes indicate different parameter sizes $\widehat{\beta}_j(\bx)$ for different levels $j$; note that in one-hot
encoding $j$ describes the different levels of the categorical variable. Secondly, the larger the box the more
interaction this level has with other feature components, because $\widehat{\beta}_j(\bx_i)$ is more
volatile over different instances $i$. 

From Figure \ref{attentions 3} we observe that Vehicle Brands B11 and B14 are the two most extreme
Vehicle Brands w.r.t.~claims frequency, B11 having the highest expected frequency and B14 the lowest. From the
French Regions R74 (Limousin) and R82 (Rh{\^o}ne-Alpes) seem outstanding having a higher frequency
than elsewhere. This finishes our example.

In fact, Figure \ref{attentions 3} provides contextualized embeddings for the different levels of the categorical
features. Attention based embedding models exactly try to do such a contextualized embedding, we refer 
to Kuo--Richman \cite{KuoRichman} in an actuarial context and to Huang et al.~\cite{Huang}
for general tabular data.

We could add the learned categorical regression attentions $\widehat{\beta}_j(\bx)$ to the variable importance
plot of Figure \ref{variable importance plot}. This requires some care. Firstly, if categorical variables have many levels,
then showing individual levels will not result in clear plots. Secondly, one-hot encoding is not normalized and centered
to unit variance, thus, these one-hot encoded variables live on a different scale compared to the standardized ones, and a direct 
comparison is not sensible.

\section{Conclusions}
\label{Conclusion section}

We have introduced the LocalGLMnet which is inspired by classical generalized linear models. Making the regression
parameters of a generalized linear model feature dependent allows us to receive a flexible regression model that
shares representation learning and the predictive performance of classical fully-connected feed-forward neural networks, and at the same time
it remains interpretable. This appealing structure allows us to perform variable selection, it allows us to study variable
importance and it also allows us to determine interactions. Thus, it provides us with a fully transparent network
model that brings out the internal structure of the data. To the best of our knowledge this is rather unique in network regression modeling, since our proposal does not share the shortcomings of similar proposals like computational
burden or a loss of predictive power.

Above we have mentioned possible extensions, e.g., LocalGLM layers can be composed to receive deeper interpretable
networks, and the LocalGLMnet can serve as a surrogate model to shed more light into many other deep learning models. 
Whereas our architecture is most suitable for tabular input data, the question about optimal consideration of non-tabular data or
of categorical variables with many levels is one point that should be further explored.

\medskip

{\small 
\renewcommand{\baselinestretch}{.51}
}

\newpage

\appendix 
\section{{\sf R} code}

\begin{lstlisting}[float=h!,frame=tb,caption={Code to implement LocalGLMnet of depth $d=4$ for the
synthetic Gaussian case.},label=LocalGLMnetCode1]
library(keras)
#
Design = layer_input(shape = c(8), dtype = 'float32') 
#
Attention = Design %>% 
          layer_dense(units=20, activation='tanh') %>%
          layer_dense(units=15, activation='tanh') %>%
          layer_dense(units=10, activation='tanh') %>%
          layer_dense(units=40, activation='linear', name='Attention')
          
Response  = list(Design, Attention) %>% layer_dot(axes=1) %>% 
          layer_dense(units=1, activation='linear', name='Response')
#
model <- keras_model(inputs = c(Design), outputs = c(Response))
model %>% compile(loss = 'mse', optimizer = 'nadam')

\end{lstlisting}

\begin{lstlisting}[float=h!,frame=tb,caption={Extraction of the 
estimated weights $\widehat{\bbeta}(\bx)$.},label=LocalGLMnetWeights1]
zz <- keras_model(inputs=model$input, outputs=get_layer(model, 'Attention')$output)
#
beta.x <- data.frame(zz %>% predict(list(XX)))       # XX denotes design/feature matrix
#
# our architecture still requires a scaling coming from dense layer 'Response' which
# we add to have intercept beta_0 in our LocalGLMnet architecture
beta.x <- beta.x * as.numeric(get_weights(model)[[9]])  
\end{lstlisting}

\begin{lstlisting}[float=h!,frame=tb,caption={Extraction of the gradients $\nabla \widehat{\beta}_j(\bx)$ and code for spline fit.},label=GradientListing]
j <- 1  # select the feature component
#
beta.j  <- Attention %>% layer_lambda(function(x) x[,j])
model.j <- keras_model(inputs = c(Design), outputs = c(beta.j))
#
grad <- beta.j %>% layer_lambda(function(x) k_gradients(model.j$outputs, model.j$inputs))
model.grad <- keras_model(inputs = c(Design), outputs = c(grad))
#
grad.beta <- data.frame(model.grad %>% predict(as.matrix(XX)))
#
#
k <- 4  # select component for interaction (j,k)
#
library(locfit)
predict(locfit(grad.beta[,k]~ beta.x[,j], alpha=0.1, deg=2), newdata=c(-400:400)/100)
\end{lstlisting}

\end{document}